\documentclass{article}

\PassOptionsToPackage{numbers, compress}{natbib}

\usepackage[preprint]{neurips_2021}

\usepackage[utf8]{inputenc} %
\usepackage[T1]{fontenc}    %
\usepackage{hyperref}       %
\usepackage{url}            %
\usepackage{booktabs}       %
\usepackage{amsfonts}       %
\usepackage{nicefrac}       %
\usepackage{microtype}      %
\usepackage{xcolor}         %
\usepackage{changes}
\usepackage{caption}

\usepackage[T1]{fontenc}

\usepackage{xifthen}
\usepackage{physics}
\usepackage{hyperref}
\usepackage{cleveref}
\usepackage{amsthm,thmtools}
\usepackage{comment}
\usepackage{graphicx}
\usepackage{tcolorbox}
\usepackage[toc,page]{appendix}
\usepackage{algorithm}
\usepackage[noend]{algpseudocode}

\newtheorem{theorem}{Theorem}[section]
\newtheorem{lemma}[theorem]{Lemma}

\newtheorem{definition}{Definition}[section]

\newtheorem{assumption}{Assumption}[section]
\newtheorem{proposition}{Proposition}[section]
\newtheorem{corollary}[theorem]{Corollary}

\let\originalleft\left
\let\originalright\right
\renewcommand{\left}{\mathopen{}\mathclose\bgroup\originalleft}
\renewcommand{\right}{\aftergroup\egroup\originalright}

\newcommand{\bP}[2][]{\Pr\ifthenelse{\isempty{#1}}{}{_{#1}}\left[#2\right]}
\newcommand{\bE}[2][]{\mathop\mathbb{E}\ifthenelse{\isempty{#1}}{}{_{#1}}\left[#2\right]}
\newcommand{\bI}[2][]{\mathop\mathbb{I}\ifthenelse{\isempty{#1}}{}{_{#1}}\left[#2\right]}
\newcommand{\Var}[2][]{\mathbf{Var}\ifthenelse{\isempty{#1}}{}{_{#1}}\left[#2\right]}

\DeclareMathOperator*{\argmax}{arg\,max}
\DeclareMathOperator*{\argmin}{arg\,min}

\newboolean{showcomments}
\setboolean{showcomments}{true}
\newcommand{\yiheng}[1]{  \ifthenelse{\boolean{showcomments}}
	{ \textcolor{purple}{(Yiheng says:  #1)}} {}  }
\newcommand{\guannan}[1]{  \ifthenelse{\boolean{showcomments}}
	{ \textcolor{blue}{(Guannan says:  #1)}} {}  }
\newcommand{\longbo}[1]{  \ifthenelse{\boolean{showcomments}}
	{ \textcolor{red}{(Longbo says:  #1)}} {}  }
\newcommand{\adam}[1]{  \ifthenelse{\boolean{showcomments}}
	{ \textcolor{orange}{(Adam says:  #1)}} {}  }

\title{Multi-Agent Reinforcement Learning\\ in Stochastic Networked Systems}
\usepackage{times}

\author{%
  Yiheng Lin \\
  CMS, Caltech\\
  \texttt{yihengl@caltech.edu} \\
  \And
  Guannan Qu \\
  CMS, Caltech\\
  \texttt{gqu@caltech.edu} \\
  \AND
  Longbo Huang \\
  IIIS, Tsinghua University \\
  \texttt{longbohuang@tsinghua.edu.cn} \\
  \And
  Adam Wierman \\
  CMS, Caltech \\
  \texttt{adamw@caltech.edu} \\
}

\begin{document}

\renewcommand\footnotemark{}
\thanks{This work was supported by NSF grants CNS-2106403 and NGSDI-2105648, with additional support from Amazon AWS, PIMCO, and the Resnick Sustainability Insitute.  Yiheng Lin was supported by Kortschak Scholars program. The work of Longbo Huang was supported by the Technology and Innovation Major Project of the Ministry of Science and Technology of China under Grants 2020AAA0108400  and 2020AAA0108403. }

\maketitle

\begin{abstract}%
  We study multi-agent reinforcement learning (MARL) in a stochastic network of agents. The objective is to find localized policies that maximize the (discounted) global reward. In general, scalability is a challenge in this setting because the size of the global state/action space can be exponential in the number of agents. Scalable algorithms are only known in cases where dependencies are static, fixed and local, e.g., between neighbors in a fixed, time-invariant underlying graph. In this work, we propose a Scalable Actor Critic framework that applies in settings where the dependencies can be non-local and stochastic, and provide a finite-time error bound that shows how the convergence rate depends on the speed of information spread in the network.  Additionally, as a byproduct of our analysis, we obtain novel finite-time convergence results for a general stochastic approximation scheme and for temporal difference learning with state aggregation, which apply beyond the setting of MARL in networked systems.
\end{abstract}

\section{Introduction}
Multi-Agent Reinforcement Learning (MARL) has achieved impressive performance in a wide array of applications including multi-player game play \cite{silver2016mastering,mnih2015human}, multi-robot systems \cite{duan2016benchmarking}, and autonomous driving \cite{li2019reinforcement}. In comparison to single-agent reinforcement learning (RL), MARL poses many challenges, chief of which is scalability \cite{zhang2019multi}. Even if each agent's local state/action spaces are small, the size of the global state/action space can be large, potentially exponentially large in the number of agents, which renders many RL algorithms such as $Q$-learning not applicable.

A promising approach for addressing the scalability challenge that has received attention in recent years is to exploit application-specific structures, e.g., \cite{gu2020qlearning,qu2019exploiting,qu2019scalable}.  A particularly important example of such a structure is a networked structure, e.g., applications in multi-agent networked systems such as social networks \cite{application_chakrabarti2008epidemic,application_llas2003nonequilibrium}, communication networks \cite{zocca2019temporal,application_communication}, queueing networks \cite{complexity_papadimitriou1999complexity}, and smart transportation networks \cite{zhang2016control}. 
In these networked systems, it is often possible to exploit \emph{static, local} dependency structures \cite{gamarnik2013correlation,gamarnik2014correlation,bamieh2002distributed,motee2008optimal}, e.g., the fact that agents only interact with a fixed set of neighboring agents throughout the game. This sort of dependency structure often leads to scalable, distributed algorithms for optimization and control \cite{gamarnik2013correlation,bamieh2002distributed,motee2008optimal}, and has proven effective for designing scalable and distributed MARL algorithms, e.g. \cite{qu2019exploiting,qu2019scalable}.

However, many real-world networked systems have inherently \emph{time-varying, non-local} dependencies. For example, in the context of wireless networks, each node can send packets to other nodes within a fixed transmission range. However, the interference range, in which other nodes can interfere the transmission, can be larger than the transmission range \cite{hierarchyDependence}. As a result, due to potential collisions, the local reward of each node not only depends on its own local state/action, but also depends on the actions of other nodes within the interference range, which may be more than one-hop away. In addition, a node may be able to observe other nodes' local states before picking its local action \cite{Backpressure}. Things become even more complex when mobility and stochastic network conditions are considered. These lead to  dependencies that are both stochastic and non-local. Although one can always fix and localize the dependence model, this leads to considerably reduced performance. Beyond wireless networks, similar stochastic and non-local dependencies exists in epidemics \cite{epi_mei2017dynamics}, social networks \cite{application_chakrabarti2008epidemic,application_llas2003nonequilibrium}, and smart transportation networks \cite{zhang2016control}. %

A challenging open question in MARL is to understand how to obtain algorithms that are scalable in settings where the dependencies are stochastic and non-local.  Prior work considers exclusively static and local dependencies, e.g., \cite{qu2019exploiting,qu2019scalable}. It is clear that hardness results apply when the dependencies are too general \cite{tabular-lower-bound}. %
Further, results in the static, local setting to this point rely on the concept of exponential decay \cite{qu2019exploiting,gamarnik2013correlation}, meaning the agents' impact on each other decays exponentially in their graph distance. This property relies on the fact that the dependencies are purely local and static, and it is not clear whether it can still be exploited when the interactions are more general. This motivates an important open question: \emph{Is it possible to design scalable algorithms for stochastic, non-local networked MARL?}

\textbf{Contributions.} In this paper, we introduce a class of stochastic, non-local dependency structures where every agent is allowed to depend on a random subset of agents. In this context, we propose and analyze a Scalable Actor Critic (SAC) algorithm that provably learns a near-optimal local policy in a scalable manner (Theorem \ref{coro:sample-complexity}). This result represents the \emph{first} provably scalable method for stochastic networked MARL. Key to our approach is that the class of dependencies we consider leads to a $\mu$-decay property (Definition \ref{def:c_rho_expo_decay}).  This property generalizes the exponential decay property underlying recent results such as \cite{qu2019exploiting,gamarnik2013correlation}, which does not apply to stochastic non-local dependencies, and enables the design of an efficient and scalable algorithm for settings with stochastic, non-local dependencies.  Our analysis of the algorithm reveals an important trade-off: as deeper interactions appear more frequently, the ``information'' can spread more quickly from one part of the network to another, which leads to the efficiency of the proposed method to degrade.  This is to be expected, as when the agents are allowed to interact globally, the problem becomes a single-agent tabular $Q$-learning problem with an exponentially large state space, which is known to be intractable since the sample complexity is polynomial in the size of the state/action space \cite{dong2019q, tabular-lower-bound}. 

The key technical result underlying our analysis of the Scalable Actor Critic algorithm is a finite-time analysis of a general stochastic approximation scheme featuring infinite-norm contraction and state aggregation (Theorem \ref{thm:Stochastic-Approx-Main}).  We apply this result to networked MARL using the local neighborhood of each agent to provide state aggregation (SA). This result also applies 
beyond MARL. Specifically, we show that it yields finite-time bounds on Temporal Difference (TD)/$Q$ learning with state aggregation (Theorem \ref{thm:TD-indicator-feature-finite}). To the best of our knowledge the resulting bound is the first finite-time bound on asynchronous $Q$-learning with state aggregation.   Additionally, it yields a novel analysis for TD-learning with state aggregation (the first error bound in the infinity norm) that sheds new insight into how the error depends on the quality of state abstraction.  These two results are important contributions in their own right.  Due to space constraints, we discuss asynchronous $Q$-learning with state aggregation in Appendix \ref{appendix:asynchronous-Q}.

\textbf{Related literature. } The prior work that is most related to our paper is \cite{qu2019scalable}, which also studies MARL in a networked setting. The key difference is that we allow the dependency structure among agents to be non-local and stochastic, while \cite{qu2019scalable} requires the dependency structure to be local and static. The generality of setting means techniques from \cite{qu2019scalable} do not apply and adds considerable complexity to the proof in two aspects. First, instead of analyzing the algorithm directly like \cite{qu2019scalable}, we derive a finite-time error bound for TD learning with state aggregation (Section \ref{sec:stocApprox} and \ref{sec:TD_indicator}), and then establish its connection with the algorithm (Section \ref{subsec:NetworkedRL:convergence}). Second, we need a more general decay property (Definition \ref{def:c_rho_expo_decay}) than the exponential one used in \cite{qu2019scalable}. Defining and establishing this general decay property for the non-local and stochastic setting is highly non-trivial (Section \ref{subsec:NetworkedRL:exp-decay}).

More broadly, MARL has received considerable attention in recent years, see \cite{zhang2019multi} for a survey. The line of work most relevant to the current paper focuses on cooperative MARL. In the cooperative setting, each agent can decide its local actions but share a common global state with other agents.  The objective is to maximize a global reward by working cooperatively.  Notable examples of this approach include \cite{marl_bu2008comprehensive, pmlr-v97-doan19a} and the references therein. In contrast, we study a situation where each agent has its own state that it acts upon. Despite the differences, like our situation, cooperative MARL problems still face scalability issues since the joint-action space is exponentially large. A variety of methods have been proposed to deal with this, including independent learners \cite{marl_claus1998dynamics,matignon2012independent}, where each agent employs a single-agent RL policy. Function approximation is another approach that can significantly reduce the space/computational complexity. One can use linear functions \cite{zhang2018fully} or neural networks \cite{lowe2017multi} in the approximation. A limitation of these approaches is the lack of theoretical guarantees on the approximation error. In contrast, our technique not only reduces the space/computational complexity significantly, but also has theoretical guarantees on the performance loss in settings with stochastic and non-local dependencies. 

The mean-field approach \cite{subramanian2019reinforcement, yang2018mean, gu2020q} provides another way to address the scalability issue, but under very different settings compared to ours. Specifically, the mean-field approach typically assumes homogeneous agents with identical local state/action space and policies, and each agent depends on other agents through their population or ``mean'' behavior. In contrast, our approach considers a local-interaction model, where there is an underlying graph and each agent depends on neighboring agents in the graph. Further, our approach allows heterogeneous agents, which means that the local state/action spaces and policies can differ among the agents. %

Another related line of work uses centralized training with decentralized execution, e.g.,  \cite{lowe2017multi, foerster2018counterfactual}, where there is a centralized coordinator that can communicate with all the agents and keep track of their experiences and policies.  In contrast, our work only requires distributed training, where we constrain the scale of communication in training within the $\kappa$-hop neighborhood of each agent.%

More broadly, this paper contributes to a growing literature that uses exponential decay to derive scalable algorithms for learning in networked systems. The specific form of exponential decay that we generalize is related to the idea of ``correlation decay'' studied in \cite{gamarnik2013correlation,gamarnik2014correlation}, though their focus is on solving static combinatorial optimization problems whereas ours is on learning policies in dynamic environments. Most related to the current paper is \cite{qu2019scalable}, which shows an exponential decay property in a restricted networked MARL model with purely local dependencies. In contrast, we show a more general $\mu$-decay property holds for a general form of stochastic, non-local dependencies.

The technical work in this paper contributes to the analysis of stochastic approximation (SA), which has received considerable attention over the past decade \cite{wu2020finite, srikant2019finite, doan2019finitetime, tengyu2019twotimescale}.  Our work is most related to \cite{qu2020finite}, which uses an asynchronous nonlinear SA to study the finite-time convergence rate for asynchronous $Q$-learning on a single trajectory. Beyond \cite{qu2020finite}, there are many other works that use SA schemes to study TD learning and $Q$-learning, e.g. \cite{srikant2019finite, wainwright2019stochastic, Lee2019AUS}. The finite-time error bound for TD learning with state aggregation in our work is most related to the asymptotic convergence limit given in \cite{TDwithFuncApprox} and the application of SA scheme to asynchronous $Q$-learning in \cite{qu2020finite}. Beyond these papers, other related work in the broader area of RL with state aggregation includes \cite{Li2006TowardsAU, Jong2005StateAD, Jiang2015AbstractionSI, Dann2018OnOP, Singh1994ReinforcementLW}. We add to this literature with a novel finite-time convergence bound for a general SA with state aggregation.  This result, in turn, yields the first finite-time error bound in the infinity norm for both TD learning with state aggregation and Q-learning with state aggregation. 

\section{Networked MARL}\label{sec:NetworkedRL}

\label{subsec:NetworkedRL:model}

We consider a network of agents that are associated with an underlying undirected graph $\mathcal{G} = (\mathcal{N}, \mathcal{E})$, where $\mathcal{N} = \{1, 2, \cdots, n\}$ denotes the set of agents and $\mathcal{E} \subseteq \mathcal{N} \times \mathcal{N}$ denotes the set of edges. 
The distance $d_{\mathcal{G}}(i, j)$ between two agents $i$ and $j$ is defined as the number of edges on the shortest path that connects them on graph $\mathcal{G}$. Each agent is associated with its local state $s_i \in \mathcal{S}_i$ and local action $a_i \in \mathcal{A}_i$ where $\mathcal{S}_i$ and $\mathcal{A}_i$ are finite sets. 
The global state/action is defined as the combination of all local states/actions, i.e.,
$s = (s_1, \cdots, s_n) \in \mathcal{S} := \mathcal{S}_1 \times \cdots \times \mathcal{S}_n,$ and
$a = (a_1, \cdots, a_n) \in \mathcal{A} := \mathcal{A}_1 \times \cdots \times \mathcal{A}_n.$ 
We use $N_i^\kappa$ to denote the $\kappa$-hop neighborhood of agent $i$ on $\mathcal{G}$, i.e., $N_i^\kappa := \{j \in \mathcal{N}\mid d_{\mathcal{G}}(i, j) \leq \kappa\}$. Let $f(\kappa) := \sup_i\abs{N_i^\kappa}$. For a subset $M \subseteq \mathcal{N}$, we use $s_M/a_M$ to denote the tuple formed by the states/actions of agents in $M$.

Before we define the transitions and rewards, we first define the notion of active link sets, which are directed graphs on the agents $\mathcal{N}$ and they characterize the interaction structure among the agents.
More specifically, an active link set is a set of directed edges that contains all self-loops, i.e., a subset of $\mathcal{N} \times \mathcal{N}$ and a super set of $\{(i, i)\mid i \in \mathcal{N}\}$. Generally speaking, $(j, i) \in L$ means agent $j$ can affect agent $i$ in the active link set $L$. Given an active link set $L$, we also use $N_i(L) := \{j \in \mathcal{N}\mid (j, i) \in L\}$ to denote the set of all agents (include itself) who can affect agent $i$ in the active link set $L$. In this paper, we consider a pair of active link sets $(L^s_t, L^r_t)$ that is independently drawn from some joint distribution $\mathcal{D}$ at each time step $t$,\footnote{Here, correlations between $L^s_t$ and $L^r_t$ are possible} where the distribution $\mathcal{D}$ will be defined using the underlying graph $\mathcal{G}$ later in Section \ref{subsec:NetworkedRL:exp-decay}. 
The role of $L_t^s/L_t^r$ is that they define the dependence structure of state transition/reward at time $t$, which we detail below. 

\textit{Transitions.} At time $t$, given the current state, action $s(t), a(t)$ and the active link set $L_t^s$, the next individual state $s_i(t+1)$ is independently generated and only depends on the state/action of the agents in $N_i(L_t^s)$. In other words, we have,
\begin{align}
    P(s(t+1)|s(t),a(t),L_t^s) = \prod_{i\in\mathcal{N}}P_i(s_i(t+1)|s_{N_i(L_t^s)}(t), a_{N_i(L_t^s)}(t),L_t^s ).\label{eq:prob_factorization}
\end{align}
\textit{Rewards.} Each agent is associated with a local reward function $r_i$. At time $t$, it is a function of $L_t^r$ and the state/action of agents in $N_i(L_t^r)$: $r_i(L^r_t, s_{N_i(L^r_t)}(t), a_{N_i(L^r_t)}(t))$. The global reward $r(t)$ is defined to be the summation of the local rewards $r_i(t)$.

\textit{Policy.} Each agent follows a localized policy that depends on its $\beta$-hop neighborhood, where $\beta \geq 0$ is a fixed integer. Specifically, at time step $t$, given the global state $s(t)$, agent $i$ adopts a local policy $\zeta_i$ parameterized by $\theta_i$ to decide the distribution of $a_i(t)$ based on the the states of agents in $N_i^\beta$. %

Our objective is for all the agents to \emph{cooperatively} maximize the discounted global reward, i.e.,
$J(\theta) = \mathbb{E}_{s\sim \pi_0}\bigg[\sum_{t=0}^\infty \gamma^t r(s(t), a(t))\mid s(0) = s\bigg],$
where $\pi_0$ is a given distribution on the initial global state, and we recall $r(s(t), a(t))$ is the global stage reward defined as the sum of all local rewards at time $t$. 

\textit{Examples.} To highlight the applicability of the general model, we include two examples of networked systems that feature the dependence structure captured by our model in Appendix \ref{appendix:experiment}: a wireless communication example and an example of controlling a process that spreads over a network. 

Note that a limitation of our setting is that the dependence structure we consider is stationary, in the sense that dependencies are sampled i.i.d.\ from the distribution $\mathcal{D}$. It is important to consider more general time-varying forms (e.g. Markovian) in future research.

\textit{Background.} Before moving on, we review a few key concepts in RL which will be useful in the rest of the section. We use $\pi_t^\theta$ to denote the distribution of $s(t)$ under policy $\theta$ given that $s(0)\sim \pi_0$. A well-known result \cite{PolicyGradientTheorem} is that the gradient of the objective $\nabla J(\theta)$ can be computed by
$\frac{1}{1 - \gamma}\mathbb{E}_{s\sim \pi^\theta, a\sim \zeta^\theta(\cdot \mid s)}Q^\theta (s, a)\nabla \log \zeta^\theta (a\mid s),$
where distribution $\pi^\theta(s) = (1 - \gamma)\sum_{t=0}^\infty \gamma^t \pi_t^\theta (s)$
is the \textit{discounted state visitation distribution}. Evaluating the $Q$-function $Q^\theta(s, a)$ plays a key role in approximating $\nabla J(\theta)$. The local $Q$-function for agent $i$ is the discounted local reward, i.e. 
$Q_i^\theta(s, a) = \mathbb{E}_{\zeta^\theta}\bigg[\sum_{t=0}^\infty \gamma^t r_i(t)\mid s(0) = s, a(0) = a\bigg],$
where we use $r_i(t)$ to denote the local reward of agent $i$ at time step $t$. Using local $Q$-functions, we can decompose the global $Q$-function as
$Q^\theta(s, a) = \frac{1}{n}\sum_{i=1}^n Q_i^\theta(s, a),$
which allows each node to evaluate its local $Q$-function separately. 

A key challenge in our MARL setting is that directly estimating the $Q$-functions is not scalable since the size of the $Q$-functions is exponentially large in the number of agents. Therefore, in Section~\ref{subsec:NetworkedRL:exp-decay}, we study structural properties of the $Q$-functions resulting from the dependence structure in the transition~\eqref{eq:prob_factorization}, which enables us to design a scalable RL algorithm in Section~\ref{subsec:NetworkedRL:SAC}.

\subsection{\texorpdfstring{$\mu$}{Lg}-decay Property}\label{subsec:NetworkedRL:exp-decay}

One of the core challenges for MARL is that the size of the $Q$ function is exponentially large in the number of agents. The key to our algorithm and its analysis is the identification of a novel structural decay property for the $Q$-function, which says that the local $Q$-function of each agent $i$ is mainly decided by the states of the agents who are near $i$. This property is critical for the design of scalable algorithms because it enables the agents to reduce the dimension of the $Q$-function
by truncating its dependence of the states and actions of far away agents. 
Recently, exponential decay has been shown to hold in networked MARL when the network is static \cite{qu2019scalable, qu2020scalable}, which is exploited to design a scalable RL algorithm. 
However, in stochastic network settings it is too much to hope for exponential decay in general \cite{easley2012networks}, and so we introduce the more general notion of $\mu$-decay here, where $\mu$ is a function that converges to $0$ as $\kappa$ tends to infinity. The case of exponential decay that has been studied previously corresponds to $\mu(\kappa) = \gamma^\kappa/(1 - \gamma)$. The formal definition of $\mu$-decay is given below, where for simplicity, we use $i \xrightarrow{L} j$ to denote $(i, j) \in L$ and denote $N_{-i}^\kappa := \mathcal{N}\setminus N_i^\kappa$. %

\begin{definition}\label{def:c_rho_expo_decay}
For a function $\mu: \mathbb{N} \to \mathbb{R}^+$ that satisfies $\lim_{\kappa \to +\infty}\mu(\kappa) = 0$, the $\mu$-decay property holds if for any policy $\theta$ and any $i \in \mathcal{N}$, the local $Q$ function $Q_i^\theta$ satisfies $\abs{Q_i^\theta(s, a) - Q_i^\theta(s', a')} \leq \mu(\kappa)$
for any $(s, a), (s', a')$ that are identical within $N_i^\kappa$, i.e. $s_{N_i^\kappa} = s_{N_i^\kappa}', a_{N_i^\kappa} = a_{N_i^\kappa}'$.
\end{definition}

Intuitively, if the $\mu$-decay property holds and $\mu(\kappa)$ decays quickly as $\kappa$ increases, we can approximately decompose the global $Q$ function as
$Q^\theta(s, a) = \frac{1}{n} \sum_{i = 1}^n Q^\theta_i(s, a) \approx \frac{1}{n} \sum_{i = 1}^n \hat{Q}^\theta_i(s_{N_i^\kappa}, a_{N_i^\kappa}),$
where $\hat{Q}_i$ only depends on the states and actions within the $\kappa$-hop neighborhood of agent $i$. Before our work, \cite{sunehag2018value}  empirically showed that such a value decomposition allows efficient training of MARL. Under the assumption that such decomposition exists, \cite{sunehag2018value} propose an approach to learn this decomposition. 
In contrast, as we prove in this section, the $\mu$ decay property holds provably and therefore, the global $Q$ function can be directly decomposed in the networked MARL model and that the error of such decomposition is provably small.

Our first result is Theorem \ref{lemma:info-spread} which shows the relationship between the random active link sets and the $\mu$-decay property. The proof of Theorem \ref{lemma:info-spread} is deferred to Appendix \ref{appendix:info-spread}.

\begin{theorem}\label{lemma:info-spread}
Define $L^a$ as the static active link set that contains all pairs $(i, j)$ whose graph distance on $\mathcal{G}$ is less than or equal to $\beta$, which is the dependency of local policy. Let random variable $X_i(\kappa)$ denote the smallest $t \in \mathbb{N}$ such that there exists a chain of agents
{\small\[j_0^a \xrightarrow{L_0^s} j_1^s \xrightarrow{L^a} j_1^a \xrightarrow{L_1^s} \cdots \xrightarrow{L_{t-1}^s} j_t^s \xrightarrow{L^a} j_t^a,\]}
that satisfies $j_0^a \in N_{-i}^\kappa$ and $j_t^a \xrightarrow{L_{t}^r} i$. The $\mu$-decay property holds for
$\mu(\kappa) = \frac{1}{1 - \gamma}\mathbb{E}\left[\gamma^{X_i(\kappa)}\right].$
\end{theorem}

To make the $\mu$-decay result more concrete, we provide several scenarios that yield different upper bounds on the term $\mathbb{E}\left[\gamma^{X_i(\kappa)}\right]$. In the first scenario, we study the case where long range links do not exist in Corollary \ref{lemma:exponential-decay}. In this case, we obtain an exponential decay property that generalizes the result in \cite{qu2019scalable}. %
A proof is in Appendix \ref{appendix:exponential-decay}.

\begin{corollary}[Exponential Decay]\label{lemma:exponential-decay}
Consider a distribution $\mathcal{D}$ of active link sets that satisfies
{\small\begin{align*}
    &P_{(L^s, L^r) \sim \mathcal{D}}\{(i, j) \in L^s\} = 0, \text{ for all }i, j \in \mathcal{N} \text{ s.t. }d_{\mathcal{G}}(i, j) \geq \alpha_1,\\
    &P_{(L^s, L^r) \sim \mathcal{D}}\{(i, j) \in L^r\} = 0, \text{ for all }i, j \in \mathcal{N} \text{ s.t. }d_{\mathcal{G}}(i, j) \geq \alpha_2.
\end{align*}}
Then, $\mathbb{E}\left[\gamma^{X_i(\kappa)}\right] \leq C \rho^\kappa$, where $\rho = \gamma^{1/(\alpha_1 + \beta)}, C = \gamma^{- \alpha_2/(\alpha_1 + \beta)}$.
\end{corollary}

In the second scenario, long range active links can occur, but with exponentially small probability with respect to their distance. In this case, we can obtain a near-exponential decay property where $\mu(\kappa) = O( \rho^{\kappa/\log\kappa}))$ for some $\rho\in(0,1)$. A proof can be found in Appendix \ref{appendix:sub-exponential-decay}.

\begin{theorem}[Near-Exponential Decay]\label{lemma:sub-exponential-decay}
Suppose the distribution $\mathcal{D}$ of active link sets satisfies
{\small\[P_{(L^s, L^r) \sim \mathcal{D}}\{(i, j) \in L^s\cup L^r\} \leq c \lambda^{d_{\mathcal{G}}(i, j)}, \text{ for all } i, j \in \mathcal{N},\]}where $c \geq 1, 1 > \lambda > 0$ are constants. If the largest size of the $\kappa$ neighborhood in the underlying graph $\mathcal{G}$ can be bounded by a polynomial of $\kappa$, i.e., there exists some constants $c_0 \geq 1, n_0 \in \mathbb{N}$ such that $\abs{\{j \in \mathcal{N} \mid d_{\mathcal{G}}(i, j) = \kappa\}} \leq c_0 (\kappa + 1)^{n_0}$ holds for all $i$, then
$\mathbb{E}\left[\gamma^{X_i(\kappa - 1)}\right] \leq C \rho^{\kappa / (1 + \ln (\kappa + 1))}$ for some positive constant $C$ and decay rate $\rho < 1$. \footnote{The explicit expression of $C$ and $\rho$ can be found in Appendix \ref{appendix:sub-exponential-decay}.}
\end{theorem}

It is interesting to compare the result above with models of the so-called ``small world phenomena" in social networks, e.g., \cite{easley2012networks}.  In these models, a link $(i, j)$ occurs with probability $1/poly(d_\mathcal{G}(i, j))$, as opposed to the exponential dependence in Lemma \ref{lemma:sub-exponential-decay}. In this case, one can see function $\mu(\kappa)$ is lower bounded by $1/poly(\kappa)$, which leads us to conjecture that $\mu(\kappa)$ is also upper bounded by $O(1/poly(\kappa))$. %
Thus, when information spreads ``slowly'' it helps a localized algorithm to learn efficiently.

\subsection{A Scalable Actor Critic Algorithm}\label{subsec:NetworkedRL:SAC}
    \begin{algorithm}[t]
        \caption{Scalable Actor Critic}\label{alg:SAC_Outline}
        \begin{algorithmic}[1]
        \For{$m = 0, 1, 2, \cdots$}
            \State Sample initial global state $s(0) \sim \pi_0$.\label{alg:SAC_Outline:Critic_Start}
            \State Each node $i$ takes action $a_i(0) \sim \zeta_i^{\theta_i(m)}(\cdot \mid s_{N_i^{\beta}}(0))$ to obtain the global state $s(1)$.
            \State Each node $i$ records $s_{N_i^\kappa}(0), a_{N_i^\kappa}(0)$, $r_i(0)$ and initialize $\hat{Q}_i^0$ to be all zero vector.
            \For{$t = 1, \cdots, T$}
                \State Each node $i$ takes action $a_i(t) \sim \zeta_i^{\theta_i(m)}(\cdot \mid s_{N_i^{\beta}}(t))$ to obtain the global state $s(t+1)$.
                \State Each node $i$ update the local estimation $\hat{Q}_i$ with step size $\alpha_{t-1} = \frac{H}{t - 1 + t_0}$,
               {\small \begin{equation*}
                    \begin{aligned}
                    &\hat{Q}_i^t \left(s_{N_i^\kappa}(t-1), a_{N_i^\kappa}(t-1)\right) =\\
                    &(1 - \alpha_{t-1})\hat{Q}_i^{t-1} \left(s_{N_i^\kappa}(t-1), a_{N_i^\kappa}(t-1)\right) + \alpha_{t-1}\left(r_i(t) + \gamma \hat{Q}_i^{t-1} \left(s_{N_i^\kappa}(t), a_{N_i^\kappa}(t)\right)\right),\\
                    &\hat{Q}_i^t \left(s_{N_i^\kappa}, a_{N_i^\kappa}\right) = \hat{Q}_i^{t-1} \left(s_{N_i^\kappa}, a_{N_i^\kappa}\right)\text{ for }\left(s_{N_i^\kappa}, a_{N_i^\kappa}\right) \not = \left(s_{N_i^\kappa}(t-1), a_{N_i^\kappa}(t-1)\right).
                    \end{aligned}
                \end{equation*}}
                \label{alg:SAC_Outline:Critic_End}
            \EndFor
            \State Each node $i$ approximate $\nabla_{\theta_i} J(\theta)$ by
            
          {\small  $\hat{g}_i(m) = \sum_{t=0}^T \gamma^t \frac{1}{n}\sum_{j \in N_i^\kappa}\hat{Q}_j^T\big(s_{N_j^\kappa}(t), a_{N_j^\kappa}(t)\big)\nabla_{\theta_i}\log \zeta_i^{\theta_i(m)}\big(a_i(t)\mid s_{N_i^{\beta}}(t)\big).$}
            \label{alg:SAC_Outline:Actor_Start}
            \State Each node $i$ conducts gradient ascent by 
            $\theta_i(m+1) = \theta_i(m) + \eta_m \hat{g}_i(m).$\label{alg:SAC_Outline:Actor_End}
        \EndFor
        \end{algorithmic}
    \end{algorithm}

Motivated by the $\mu$-decay property of the $Q$-functions, we design a novel Scalable Actor Critic algorithm (Algorithm \ref{alg:SAC_Outline}) for networked MARL problem, which exploits the $\mu$-decay result in the previous section.  The Critic part (from line \ref{alg:SAC_Outline:Critic_Start} to line \ref{alg:SAC_Outline:Critic_End}) uses the local trajectory $\{(s_{N_i^\kappa}, a_{N_i^\kappa}, r_i)\}$ to evaluate the local $Q$-functions under parameter ${\theta(m)}$. Intuitively, the $\mu$-decay property guarantees that we can achieve good approximation error even when $\kappa$ is not large. The Actor part (from line \ref{alg:SAC_Outline:Actor_Start} to line \ref{alg:SAC_Outline:Actor_End}) computes the estimated partial derivative using the estimated local $Q$-functions, and uses the partial derivative to update local parameter $\theta_i$. The step size sequence $\{\eta_m\}$ will be defined in Theorem \ref{thm:actor-SAC}. Compared with the Scalable Actor Critic algorithm proposed in \cite{qu2019scalable}, Algorithm \ref{alg:SAC_Outline} extends the policy dependency structure considered.  No longer is the dependency completely local; it now extends to all agents within the $\beta$-hop neighborhood. Interestingly, the time-varying dependencies do not add complexity into the algorithm (though the analysis is more complex).

Algorithm \ref{alg:SAC_Outline} is highly scalable. Each agent $i$ needs only to query and store the information within its $\kappa$-hop neighborhood during the learning process. The parameter $\kappa$ can be set to balance accuracy and complexity. Specifically, as $\kappa$ increases, the error bound becomes tighter at the expense of increasing computation, communication, and space complexity.

\subsection{Convergence}\label{subsec:NetworkedRL:convergence}
We now present our main result, a finite-time error bound for the Scalable Actor Critic algorithm (Algorithm \ref{alg:SAC_Outline}) that holds under general (non-local) dependencies.  To that end, we first describe the assumption needed in our result. It focuses on the Markov chain formed by the global state-action pair $(s, a)$ under a fixed policy parameter $\theta$ and is standard for finite-time convergence results in RL, e.g., \cite{srikant2019finite, bremaud2013markov, qu2020finite}.

\begin{assumption}\label{network-assp:geometric-mixing}
Under any fixed policy $\theta$, $\{z(t) := \left(s(t), a(t)\right)\}$ is an aperiodic and irreducible Markov chain on state space $\mathcal{Z} := \mathcal{S} \times \mathcal{A}$ with a unique stationary distribution $d^\theta = (d_z^\theta, z \in \mathcal{Z})$, which satisfies $d_z^\theta > 0, \forall z \in \mathcal{Z}$. Define $d^\theta(z') = \sum_{z \in \mathcal{Z}: z_{N_i^\kappa} = z'} d^\theta(z)$ and $\sigma'(\kappa) := \inf_{z' \in \mathcal{Z}_{N_i^\kappa}} d^\theta(z')$. There exists positive constants $K_1, K_2$ such that $K_2 \geq 1$ and $\forall z' \in \mathcal{Z}, \forall t \geq 0, \sup_{\mathcal{K} \subseteq \mathcal{Z}}\abs{\sum_{z \in \mathcal{K}}d_z^\theta - \sum_{z \in \mathcal{K}}\mathbb{P}(z(t) = z \mid z(0) = z')} \leq K_1 e^{-t/K_2}$.
\end{assumption}

We next analyze the Critic part of Algorithm \ref{alg:SAC_Outline} within a given outer loop iteration $m$. Since the policy is fixed in the inner loop, the global state/action pair $(s, a)$ in the original MDP can be viewed as the state of a Markov chain. 
We observe that each local estimate $\hat{Q}_i^t \left(s_{N_i^\kappa}, a_{N_i^\kappa}\right) $ can be viewed as a form of state aggregation, where the global state $(s,a)$ is ``compressed'' to $h(s,a):= (s_{N_i^\kappa}, a_{N_i^\kappa})$. Broadly speaking, the technique of state aggregation is one of the easiest-to-deploy schemes for state space compression \cite{Jiang2018NotesOS,Singh1994ReinforcementLW}, while its final performance relies heavily on whether the state aggregation map $h$ only aggregates ``similar'' states. To have a good approximate equivalence, we need to find a good $h$, i.e., if two states are mapped to the same abstract state, their value functions are required to be close (to be discussed in Theorem \ref{thm:TD-indicator-feature-finite}). In the context of networked MARL, the $\mu$ decay property (Definition \ref{def:c_rho_expo_decay}) provides a natural mapping for state aggregation $h(s,a):= (s_{N_i^\kappa}, a_{N_i^\kappa})$ which we defined earlier. This mapping $h$ maps the global state/action to the local states/actions in agent $i$'s $\kappa$-hop neighborhood and the $\mu$-decay property guarantees that if $h(s, a) = h(s', a')$, the difference in their $Q$-functions is upper bounded by $\mu(\kappa)$, which is vanishing as $\kappa$ increases. This shows that the mapping $h$ we used is ``good'' in the sense it aggregates very similar global state-action pairs. This idea leads to the following theorem about the Critic part of Scalable Actor Critic (Algorithm \ref{alg:SAC_Outline}).

\begin{theorem}\label{coro:critic-SAC}
Suppose Assumption \ref{network-assp:geometric-mixing} and $\mu$-decay property (Definition \ref{def:c_rho_expo_decay}) hold. Let the step size be $\alpha_t = \frac{H}{t + t_0}$ with $t_0 = \max(4H, 2K_2 \log T)$, and $H \geq \frac{2}{(1 - \gamma) \sigma'(\kappa)}$. Define constant $C_b := 4 K_1(1 + 2 K_2 + 4H)$. Then, inside outer loop iteration $m$, for each $i \in \mathcal{N}$, with probability at least $1 - \delta$, we have $
    \sup_{(s, a)\in \mathcal{S}\times \mathcal{A}}\abs{Q_i^{\theta(m)}(s, a) - \hat{Q}_i^T (s_{N_i^\kappa}, a_{N_i^\kappa})}
    \leq \frac{C_a}{\sqrt{T + t_0}} + \frac{C_a'}{T + t_0} + \frac{\mu(\kappa)}{1 - \gamma},
$
where the constants are given by $C_a = \frac{40H}{(1 - \gamma)^2}\sqrt{K_2 \log T\left(\log \left(\frac{4f(\kappa) K_2 T}{\delta}\right) + \log \log T \right)}$ and $C_a' = \frac{8}{(1 - \gamma)^2}\max\{\frac{144 K_2 H \log T}{\sigma'(\kappa)} + C_b, 2K_2 \log T + t_0\}$.
\end{theorem}

The proof of Theorem \ref{coro:critic-SAC} can be found in Appendix \ref{appendix:critic-SAC}. The most related result in the literature to Theorem \ref{coro:critic-SAC} is Theorem 7 in \cite{qu2019scalable}.  In comparison, Theorem \ref{coro:critic-SAC} applies for more general, potentially non-local, dependencies and, also, improves the constant term by a factor of $1/(1 - \gamma)$.

To analyze the Actor part of Algorithm \ref{alg:SAC_Outline}, we make the following additional boundedness and Lipschitz continuity assumptions on the gradients. These are standard assumptions in the literature.

\begin{assumption}\label{network-assp:bounded-grad}
For any $i, a_i, s_{N_i^\beta}$ and $\theta_i$, we assume $\norm{\nabla_{\theta_i}\log \zeta_i^{\theta_i}(a_i \mid s_{N_i^\beta})} \leq W_i$. Then, for any $L_t^a$, $\norm{\nabla_\theta \log \zeta^\theta(a \mid s)} \leq W :=  \sqrt{\sum_{i=1}^n W_i^2}$. We further assume $\nabla J(\theta)$ is $W'$-Lipschitz in $\theta$.
\end{assumption}

Intuitively, since the quality of the estimated policy gradient depends on the quality of the estimation of $Q$-functions, if every agent $i$ has learned a good approximation of its local $Q$-function in the Critic part of Algorithm \ref{alg:SAC_Outline}, the policy gradient can be approximated well. Therefore, the Actor part can obtain a good approximation of a stationary point of the objective function. We state the sample complexity result in Theorem \ref{coro:sample-complexity} and defer the detailed bounds and a proof to Appendix \ref{appendix:sample-complexity}.

\begin{theorem}\label{coro:sample-complexity}
Under Assumption \ref{network-assp:bounded-grad}, to reach an $O(\epsilon)$-approximate stationary point with probability at least $1 - \delta$, we need to choose $\kappa$ such that $\mu(\kappa) = O\left(W^{-2} (1 - \gamma)^4 \epsilon \right)$.
The number of required iterations of the outer loop should satisfy $M = \tilde{\Omega}\big(\epsilon^{-2} poly(W, W', \frac{1}{1 - \gamma})\big)$ and the number of required iterations of the inner loop is
$T = \tilde{\Omega}\big( \epsilon^{-2} poly(W, \frac{1}{\sigma'(\kappa)}, K_2, \frac{1}{1 - \gamma}, \log f(\kappa), \log (1/\delta))\big)$.
\end{theorem}

Note that $W$ scales with the number of agents $n$. Thus, Theorem \ref{coro:sample-complexity} shows that the complexity of our algorithm scales with the largest state-action space size of any $\kappa$-hop neighborhood and the number of agents $n$, which avoids the exponential blowup in $n$ when the graph is sparse and achieves scalable RL for networked agents even under stochastic, non-local settings.

\section{Proof Idea: Stochastic Approximation and State Aggregation}\label{sec:stocAndTD}
In this section, we present the key technical innovation underlying our results on MARL in Theorem \ref{coro:critic-SAC}: a new finite-time analysis of a general asynchronous stochastic approximation (SA) scheme. As we mention in \Cref{sec:NetworkedRL}, the truncation enabled by $\mu$-decay provides a form of state aggregation, which we analyze via a general SA scheme in \Cref{sec:stocApprox}. Further, this SA scheme is of interest more broadly, e.g., to the settings of  TD learning with state aggregation (Section \ref{sec:TD_indicator}) and asynchronous $Q$-learning with state aggregation (Appendix \ref{appendix:asynchronous-Q}). %

\subsection{Stochastic Approximation}
\label{sec:stocApprox}

Consider a finite-state Markov chain whose state space is given by $\mathcal{N} = \{1, 2, \cdots, n\}$. Let $\{i_t\}_{t=0}^\infty$ be the sequence of states visited by this Markov chain. Our focus is generalizing the following asynchronous stochastic approximation (SA) scheme, which is studied in \cite{tsitsiklis1994asynchronous,shah2018q,wainwright2019stochastic}: Let parameter $x \in \mathbb{R}^\mathcal{N}$, and $F: \mathbb{R}^\mathcal{N} \to \mathbb{R}^\mathcal{N}$ be a $\gamma$-contraction in the infinity norm. The update rule of the SA scheme is given by
\begin{equation}\label{equ:initial-Asy-Stoc-Approx}
    \begin{aligned}
    x_{i_t}(t+1) &= x_{i_t}(t) + \alpha_t\left(F_{i_t}\left(x(t)\right) - x_{i_t}(t) + w(t)\right),\\
    x_j(t+1) &= x_j(t) \text{ for }j \not = i_t, j \in \mathcal{N},
    \end{aligned}
\end{equation}
where $w(t)$ is a noise sequence. It is shown in \cite{qu2020finite} that parameter $x(t)$ converges to the unique fixed point of $F$ at the rate of $O\left({1}/{\sqrt{t}}\right)$.

While general, in many cases, including networked MARL, we do not wish to calculate an entry for every state in $\mathcal{N}$ in parameter $x$, but instead, wish to calculate ``aggregated entries.'' Specifically, at each time step, after $i_t$ is generated, we use a surjection $h$ to decide which dimension of parameter $x$ should be updated. This technique, referred to as state aggregation, is one of the easiest-to-deploy schemes for state space compression in the RL literature \cite{Jiang2018NotesOS,Singh1994ReinforcementLW}.
In the generalized SA scheme, our objective is to specify the convergence point as well as obtain a finite-time error bound.

Formally, to define the generalization of \eqref{equ:initial-Asy-Stoc-Approx}, let $\mathcal{N} = \{1, \cdots, n\}$ be the state space of $\{i_t\}$ and $\mathcal{M} = \{1, \cdots, m\}, (m \leq n)$ be the \textit{abstract} state space. The surjection $h: \mathcal{N} \to \mathcal{M}$ is used to convert every state in $\mathcal{N}$ to its abstraction in $\mathcal{M}$. Given parameter $x \in \mathbb{R}^\mathcal{M}$ and function $F: \mathbb{R}^\mathcal{N} \to \mathbb{R}^\mathcal{N}$, we consider the generalized SA scheme that updates $x(t)\in \mathbb{R}^{\mathcal{M}}$ starting from $x(0) = \mathbf{0}$, 
\begin{equation}\label{equ:Generalized-Asy-Stoc-Approx}
    \begin{aligned}
    x_{h(i_t)}(t+1) ={}& x_{h(i_t)}(t) + \alpha_t \left(F_{i_t}\left( \Phi x(t)\right) - x_{h(i_t)}(t) + w(t)\right),\\
    x_j(t+1) ={}& x_j(t) \text{ for }j \not = h(i_t), j \in \mathcal{M},
    \end{aligned}
\end{equation}
where the feature matrix $\Phi \in \mathbb{R}^{\mathcal{N} \times \mathcal{M}}$ is defined as
\begin{equation}\label{equ:def:Phi}
    \Phi_{ij} = \begin{cases}
    1 & \text{ if }h(i) = j\\
    0 & \text{ otherwise}
    \end{cases}, \forall i \in \mathcal{N}, j \in \mathcal{M}.
\end{equation}

In order to state our main result characterizing the convergence of \eqref{equ:Generalized-Asy-Stoc-Approx}, we must first state a few definitions and assumptions.  To begin, we define the weighted infinity norm as in \cite{qu2020finite}, except that we extend its definition so as to define the contraction of function $F$. The reason we use the weighted infinity norm as opposed to the standard infinity norm is that its generality can be used in certain settings for undiscounted RL, as shown in \cite{tsitsiklis1994asynchronous,bertsekas2007dpbook}.

\begin{definition}[Weighted Infinity Norm]
Fix a positive vector $v \in \mathbb{R}^\mathcal{M}$. For $x \in \mathbb{R}^\mathcal{M}$, we define
$\norm{x}_v := \sup_{i\in \mathcal{M}}\frac{\abs{x_i}}{v_i}$. For $x \in \mathbb{R}^\mathcal{N}$, we define $\norm{x}_v := \sup_{i\in \mathcal{N}}\frac{\abs{x_i}}{v_{h(i)}}$.
\end{definition}

Next, we state our assumption on the mixing rate of the Markov chain $\{i_t\}$, which is common in the literature \cite{tsitsiklis1997analysis, srikant2019finite}. It holds for any finite-state Markov chain which is aperiodic and irreducible \cite{bremaud2013markov}.

\begin{assumption}[Stationary Distribution and Geometric Mixing Rate]\label{assump:geometric-mixing}
$\{i_t\}$ is an aperiodic and irreducible Markov chain on state space $\mathcal{N}$ with stationary distribution $d = (d_1, d_2, \cdots, d_n)$. Let
$d_j' = \sum_{i \in h^{-1}(j)}d_i$ and $\sigma' = \inf_{j \in \mathcal{M}}d_j'.$
There exists positive constants $K_1, K_2$ which satisfy that
$\sup_{\mathcal{S} \subseteq \mathcal{N}}\abs{\sum_{i \in \mathcal{S}}d_i - \sum_{i \in \mathcal{S}}\mathbb{P}(i_t = i \mid i_0 = j)} \leq K_1 \exp(-t/K_2), \forall j \in \mathcal{N}, \forall t \geq 0$ and $K_2 \geq 1$.
\end{assumption}

Our next assumption ensures contraction of $F$. It is also standard, e.g., \cite{tsitsiklis1994asynchronous, wainwright2019stochastic, qu2020finite}, and ensures that $F$ has a unique fixed point $y^*$.

\begin{assumption}[Contraction]\label{assump:contraction}
Operator $F$ is a $\gamma$ contraction in $\norm{\cdot}_v$, i.e., for any $x, y \in \mathbb{R}^\mathcal{N}$, we have $\norm{F(x) - F(y)}_v \leq \gamma \norm{x - y}_v.$ Further, there exists some constant $C > 0$ such that for any $x \in \mathbb{R}^\mathcal{N}$, we have $\norm{F(x)}_v \leq \gamma \norm{x}_v + C.$
\end{assumption}

In Assumption \ref{assump:contraction}, notice that the first sentence directly implies the second with $C = (1 + \gamma)\norm{y^*}_v$,
where $y^* \in \mathbb{R}^\mathcal{N}$ is the unique fixed point of $F$.  Further, while Assumption \ref{assump:contraction} implies that $F$ has a unique fixed point $y^*$, we do not expect our stochastic approximation scheme to converge to it. Instead, we show that the convergence is to the unique $x^*$ that solves
\begin{equation}\label{equ:def-Pi}
    \Pi F(\Phi x^*) = x^*, \text{ where } \Pi := \left(\Phi^\top D \Phi\right)^{-1}\Phi^\top D.
\end{equation}
Here $D = diag(d_1, d_2, \cdots, d_n)$ denotes the steady-state probabilities for the process $\{i_t\}$. Note that $x^*$ is well-defined because the operator $\Pi F(\Phi \cdot)$, which defines a mapping from $\mathbb{R}^\mathcal{M}$ to $\mathbb{R}^\mathcal{M}$, is also a contraction in $\norm{\cdot}_v$. We state and prove this as Proposition \ref{proposition:proj-contraction} in Appendix \ref{appendix:proposition:proj-contraction}.

Our last assumption is on the noise sequence $w(t)$. It is also standard, e.g., \cite{shah2018q,qu2020finite}.

\begin{assumption}[Martingale Difference Sequence]\label{assump:martingale}
$w_t$ is $\mathcal{F}_{t+1}$ measurable and satisfies $\mathbb{E} w(t)\mid \mathcal{F}_t = 0$. Further, $\abs{w(t)} \leq \bar{w}$ almost surely for constant $\bar{w}.$
\end{assumption}

We are now ready to state our finite-time convergence result for stochastic approximation.

\begin{theorem}\label{thm:Stochastic-Approx-Main}
Suppose Assumptions \ref{assump:geometric-mixing}, \ref{assump:contraction}, \ref{assump:martingale} hold. Further, assume there exists constant $\bar{x} \geq \norm{x^*}_v$ such that $\forall t, \norm{x(t)}_v \leq \bar{x}$ almost surely.\footnote{The assumption on $\bar{x}$ follows from Assumptions \ref{assump:contraction} and \ref{assump:martingale}. See Proposition \ref{proposition:x-bar} in Appendix \ref{appendix:proposition:x-bar}.} Let the step size be $\alpha_t = \frac{H}{t + t_0}$ with $t_0 = \max(4H, 2K_2 \log T)$, and $H \geq \frac{2}{\sigma' (1 - \gamma)}$. Let $x^*$ be the unique solution of equation $\Pi F(\Phi x^*) = x^*$, and define constants $C_1 := 2\bar{x} + C + \frac{\bar{w}}{\underline{v}}, C_2 := 4\bar{x} + 2C + \frac{\bar{w}}{\underline{v}}, C_3 := 2K_1(2\bar{x} + C)(1 + 2K_2 + 4H)$.
Then, with probability at least $1 - \delta$,
{\small\[\norm{x(T) - x^*}_v \leq \frac{C_a}{\sqrt{T + t_0}} + \frac{C_a'}{T + t_0} = \tilde{O}\left(\frac{1}{\sqrt{T}}\right),\]}where the constants are given by $C_a = \frac{4H C_2}{1 - \gamma}\sqrt{K_2 \log T \left(\log \left(\frac{4m K_2 T}{\delta}\right) + \log \log T\right)}$ and $C_a'=  4\max\{\frac{48K_2 C_1 H \log T + \sigma' C_3}{(1 - \gamma)\sigma'},
    \frac{2\bar{x}(2K_2 \log T + t_0)}{1 - \gamma}\}$.

\end{theorem}
A proof of Theorem \ref{thm:Stochastic-Approx-Main} can be found in Appendix \ref{appendix:thm:Stochastic-Approx-Main}. Compared with Theorem 4 in \cite{qu2020finite}, Theorem \ref{thm:Stochastic-Approx-Main} holds for a more general SA scheme where state aggregation is used to reduce the dimension of the parameter $x$. The proof technique used in \cite{qu2020finite} does not apply to our setting because our stationary point $x^*$ has a more complex form \eqref{equ:def:Phi}.
To do the generalization, we need to use a different error decomposition method compared to \cite{qu2020finite} that leverages the stationary distribution $D$ rather than the distribution of $i_t$ condition on $i_{t-\tau}$ (see Appendix \ref{appendix:thm:Stochastic-Approx-Main} for details).
Because of this generality, Theorem \ref{thm:Stochastic-Approx-Main} requires a stronger but standard assumption on the mixing rate of the Markov chain $\{i_t\}$. %

\subsection{State Aggregation}\label{sec:TD_indicator}

To illustrate the impact of our analysis of SA (Theorem \ref{thm:Stochastic-Approx-Main}) beyond the network setting, we study a simpler application to the cases of TD-learning and $Q$-learning with state aggregation in this section.  Understanding state aggregation methods is a foundational goal of analysis in the RL literature and it has been studied in many previous works, e.g.,  \cite{Li2006TowardsAU, Jong2005StateAD, Jiang2015AbstractionSI, Dann2018OnOP, Singh1994ReinforcementLW}.  %
Further, the result is extremely useful in the analysis in networked MARL that follows since the $\mu$-decay property we introduce  (Definition \ref{def:c_rho_expo_decay}) provides a natural state aggregation in the network setting (see Corollary \ref{coro:critic-SAC}). Due to space constraints, in this section we only introduce the results on TD-learning; the results on $Q$-learning are given in Appendix \ref{appendix:asynchronous-Q}. 

In TD learning with state aggregation \cite{Singh1994ReinforcementLW, TDwithFuncApprox}, given the sequence of states visited by the Markov chain is $\{i_t\}$, the update rule of TD$(0)$ is given by 
\begin{equation}\label{equ:TD0_update}
    \begin{aligned}
    \theta_{h(i_t)}(t + 1) &= \theta_{h(i_t)}(t) + \alpha_t \left(r_t + \gamma \theta_{h(i_{t+1})}(t) - \theta_{h(i_t)}(t)\right),\\
    \theta_j(t + 1) &= \theta_j(t) \text{ for }j \not = h(i_t), j \in \mathcal{M},
    \end{aligned}
\end{equation}
where $h: \mathcal{N} \to \mathcal{M}$ is a surjection that maps each state in $\mathcal{N}$ to an abstract state in $\mathcal{M}$ and $r_t$ is the reward at time step $t$ such that $\mathbb{E}[r_t] = r(i_t, i_{t+1})$.

Taking $F$ as the Bellman Policy Operator, i.e., the $i$'th dimension of function $F$ is given by 
\[F_i(V) = \mathbb{E}_{i' \sim \mathbb{P}(\cdot \mid i)}\left[r(i, i') + \gamma V_{i'}\right], \forall i \in \mathcal{N}, V \in \mathbb{R}^\mathcal{N}.\]
The value function (vector) $V^*$ is defined as $V_i^* = \mathbb{E}\left[\sum_{t=0}^\infty \gamma^t r(i_t, i_{t+1})\mid i_0 = i\right], i \in \mathcal{N}$ \cite{TDwithFuncApprox}. By defining the feature matrix $\Phi$ as \eqref{equ:def:Phi} and the noise sequence as
\[w(t) = r_t + \gamma \theta_{h(i_{t+1})}(t) - \mathbb{E}_{i'\sim \mathbb{P}(\cdot \mid i_t)}[r(i_t, i') + \gamma \theta_{h(i')}(t)],\]
we can rewrite the update rule of TD$(0)$ in \eqref{equ:TD0_update} in the form of an SA scheme \eqref{equ:Generalized-Asy-Stoc-Approx}. Therefore, we can apply Theorem \ref{thm:Stochastic-Approx-Main} to obtain a finite-time error bound for TD learning with state aggregation. A proof of Theorem \ref{thm:TD-indicator-feature-finite} can be found in Appendix \ref{appendix:thm:TD-indicator-feature-finite}. 

\begin{theorem}\label{thm:TD-indicator-feature-finite}
Let Assumption \ref{assump:geometric-mixing} hold for the Markov chain $\{i_t\}$ and let the stage reward $r_t$ be upper bounded by $\bar{r}$ almost surely. Assume that if $h(i) = h(i')$ for $i, i' \in \mathcal{N}$, we have $\abs{V_i^* - V_{i'}^*} \leq \zeta$ for a constant $\zeta$. Consider TD$(0)$ with the step size $\alpha_t = \frac{H}{t + t_0}$, where $t_0 = \max(4H, 2K_2 \log T)$ and $H \geq \frac{2}{\sigma' (1 - \gamma)}$. Define constant $C_4 := 4K_1(1 + 2K_2 + 4H)$. Then, with probability at least $1 - \delta$,
{\small
\[\norm{\Phi \cdot \theta(T) - V^*}_\infty \leq \frac{C_a}{\sqrt{T + t_0}} + \frac{C_a'}{T + t_0} + \frac{\zeta}{1 - \gamma}, \]
where the constants are given by $C_a = \frac{40H\bar{r}}{(1 - \gamma)^2}\sqrt{K_2 \log T \left(\log \left(\frac{4m K_2 T}{\delta}\right) + \log \log T\right)}$ and $C_a' = \frac{8\bar{r}}{(1 - \gamma)^2}\max\{\frac{144 K_2 H \log T}{\sigma'} + C_4, 2K_2 \log T + t_0\}$.
 }
\end{theorem}
The most related prior results to  Theorem \ref{thm:TD-indicator-feature-finite} are \cite{srikant2019finite, bhandari2018finite}. 
In contrast to these, Theorem \ref{thm:TD-indicator-feature-finite} considers the infinity norm, which is more natural for measuring error when using state aggregation. Further, our analysis is different and extends to the case of $Q$-learning with state aggregation (see Appendix \ref{appendix:asynchronous-Q}), where we obtain the first finite-time error bound. Moreover, unlike \cite{bhandari2018finite}, our TD-learning algorithm does not require a projection step. %

\section{Concluding Remarks}

In this paper, we propose and analyze the Scalable Actor Critic Algorithm that provably learns a near-optimal local policy in a setting where every agent is allowed to interact with a random subset of agents. The $\mu$-decay property, which enables the decentralized approximation of local $Q$ functions, is the key to our approach.

There are a number of future directions motivated by the results in this paper. For example, we allow the interaction structure among the agents to change in a stochastic way in this work. It is interesting to see if such structure can be time-varying in more general ways (e.g., Markovian or adversarial). Besides, although our Scalable Actor Critic algorithm consumes much less memory than a centralized tabular approach, the memory space required by each agent $i$ to store $\hat{Q}_i$ grows exponentially with respect to $f(\kappa)$, which denotes the size of the largest $\kappa$-hop neighborhood. Thus, memory problems may still arise if $f$ grows quickly as $\kappa$ increases. Therefore, an interesting open problem is whether we can apply additional function approximations on truncated state/action pair $(s_{N_i^\kappa}, a_{N_i^\kappa})$, and obtain similar finite-time convergence guarantees as Scalable Actor Critic.

\bibliographystyle{abbrv}
\bibliography{main.bib}

\clearpage

\appendix

\section{Examples}\label{appendix:experiment}

\subsection{Wireless Networks}\label{appendix:experiment-wireless}
\begin{figure}[h]
    \centering
    \includegraphics[width = .5\textwidth]{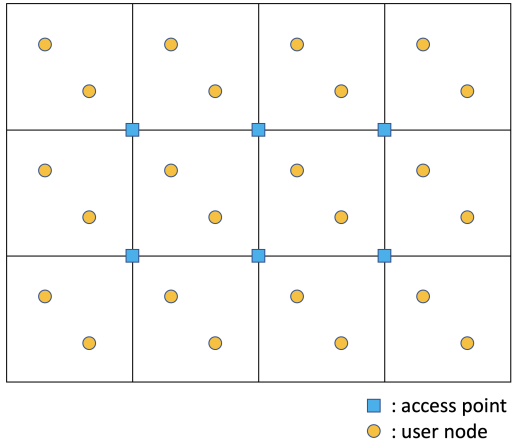}
    \caption{An example setup of wireless networks. Each user node can send packets to the access points at the corners of its grid.}
    \label{fig:grid-structure}
\end{figure}

We consider a wireless network with multiple access points setting shown in Fig.~\ref{fig:grid-structure}, where a set of user nodes in a wireless  network, denoted by $U = \{u_1, u_2, \cdots, u_n\},$ share a set of access points $Y = \{y_1, y_2, \cdots, y_m\}$ \cite{zocca2019temporal}. Each access point $y_i$ is associated with a probability $p_i$ of successful transmission. Each user node $u_i$ only has access to a subset $Y_i \subseteq Y$ of the access points. Typically, this available set is determined by each user node's physical connections to the access points. To apply the networked MARL model, we identify the set of user nodes $U$ as the set of agents $\mathcal{N}$ in Section \ref{sec:NetworkedRL}. The underlying graph $G = (\mathcal{N}, \mathcal{E})$ is defined as the conflict graph, i.e., edge $(u_i, u_j) \in \mathcal{E}$ if and only if $Y_i \cap Y_j \not = \emptyset$.%

At each time step $t$, each user $u_i$ receives a packet with initial life span $d$ with probability $q$. Each user maintains a queue to cache the packets it receives. At each time step, if the packet is successfully sent to an access point, it will be removed from the queue. Otherwise, its life span will decrease by $1$. A packet is discarded from the queue immediately if its remaining life span is $0$. At each time step $t$, a user node $u_i$ can choose to send one of the packets in its queue to one of the access point $y_{i,t} \in Y_i$. If no other user node sends packets to access point $y_{i,t}$ at time step $t$, the packet from user $i$ can be delivered successfully   with probability $p_i$.  Otherwise, the sending action will fail. A user $u_i$ receives a local reward of $r_{i,t} = 1$ immediately after successfully sending a packet at time step $t$, and receives $r_{i, t} = 0$ otherwise. Our objective is to find a policy that maximizes the global discounted reward under a discounted factor $0 \leq \gamma < 1$:
\[\mathbb{E}\left[\sum_{i=1}^n \sum_{t=0}^\infty \gamma^t r_{i,t}\right].\] %

To see how this setting fits into our model, we first define the local state/action and specify the parameters. Since each packet has a life span of $d$, and each user node receives at most one packet at a time step, we  use a $d$-tuple $s_i = (e_1, e_2, \cdots, e_d) \in \mathcal{S}_i := \{0, 1\}^d$ to denote the local state of user node $i$. Specifically, $e_j$ indicates whether user node $u_i$ has a packet with remaining life span $j$ in its queue. A local action of user node $u_i$ is $2$-tuple $(l, y)$, which means sending the packet with remaining life span $l \in \{1, 2, \cdots, d\}$ to an access point $y \in Y_i$. Note that we define an empty action that does nothing at all. If a user node performs an action $(l, y)$ when there is no packet with life span $l$ in its queue, we view this as an empty action. This setting falls into the category we studied in Corollary \ref{lemma:exponential-decay}, where long range links do not exist. Specifically, in this setting, the next local state of user node $u_i$ depends on the current local states/actions in its $1$-hop neighborhood ($\alpha_1 = 1$ in Corollary \ref{lemma:exponential-decay}). We assume each user node can choose its action only based on its current local state ($\beta = 0$). Due to potential collisions, the local reward of user $u_i$ also depends on the states/actions in its $1$-hop neighborhood ($\alpha_2 = 1$ in Corollary \ref{lemma:exponential-decay}). Though this is a static setting, note that the results of \cite{qu2019scalable} do not apply.

\begin{figure}
  \begin{minipage}[b]{0.5\textwidth}
    \centering
    \captionsetup{width=.8\linewidth}
    \includegraphics[width = .9\textwidth]{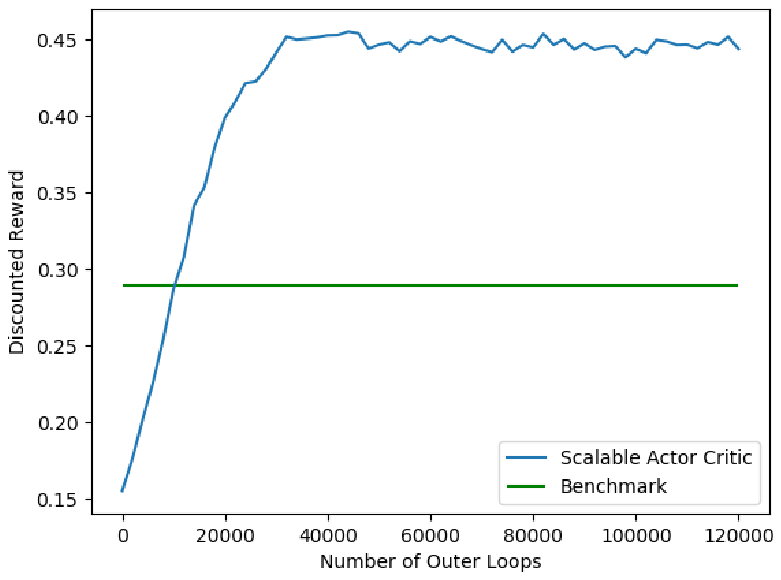}
    \caption{Discounted reward in the training process. $5\times 5$ grid, $1$ user per grid.}
    \label{fig:w5h5c1-120000-05}
  \end{minipage}~
  \begin{minipage}[b]{0.5\textwidth}
    \centering
    \captionsetup{width=.8\linewidth}
    \includegraphics[width = .9\textwidth]{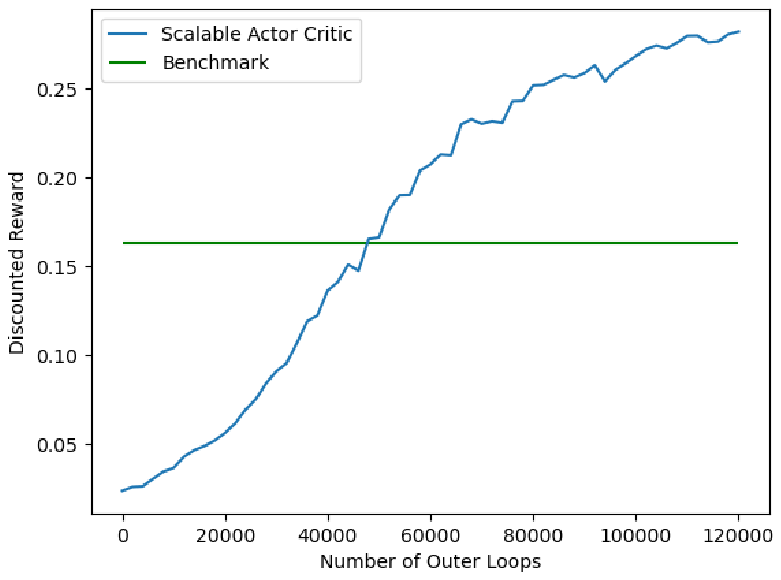}
    \caption{Discounted reward in the training process. $3\times 4$ grid, $2$ users per grid.}
    \label{fig:w4h3c2-120000-05}
  \end{minipage}
\end{figure}

The detailed setting we use is as follows. We consider the setting where the user nodes are located in $h \times w$ grids (see Fig.~\ref{fig:grid-structure}). There are $c$ user nodes in each grid, and each user can send packets to an access point on the corner of its grid. We set the initial life span $d = 2$, the arrival probability $q = 0.5$, and the discounted factor $\gamma = 0.7$. The successful transmission probability $p_i$ for each access point $y_i$ is sampled uniformly randomly from $[0, 1]$. We run the Scalable Actor Critic algorithm with parameter $\kappa = 1$ to learn a localized stochastic policy in two cases $(h, w, c) = (5, 5, 1)$ (see Fig.~\ref{fig:w5h5c1-120000-05}) and $(h, w, c) = (3, 4, 2)$ (see Fig.~\ref{fig:w4h3c2-120000-05}). For comparison, we use a benchmark based on the localized ALOHA protocol \cite{roberts1975aloha}. Specifically, the benchmark policy works as following: At time step $t$, each user node $u_i$ takes the empty action with a certain probability $p'$; otherwise, it sends the packet with the minimum remaining life span to a random access point in $Y_i$, with the probability proportional to the successful transmission probability of this access point and inverse proportional to the number of users sharing this access point. In Fig.~\ref{fig:w5h5c1-120000-05} and Fig.~\ref{fig:w4h3c2-120000-05}, we have tuned the parameter $p'$ to find the one with the highest discounted reward.

As shown in Fig.~\ref{fig:w5h5c1-120000-05} and Fig.~\ref{fig:w4h3c2-120000-05}, starting from the initial policy that chooses an local action uniformly at random, the Scalable Actor Critic algorithm with parameter $\kappa = 1$ can learn a policy that performs better than the benchmark. As a remark, the benchmark policy requires the set $\{p_i\}_{1\leq i \leq m}$, the probability of successful transmission, as input. Moreover, in the benchmark policy, the probability of performing an empty action also needs to be tuned manually. In contrast, the Scalable Actor Critic algorithm can learn a better policy without these specific inputs by interacting with the system.
\subsection{Spreading Networks}\label{appendix:experiment-spreading}
\begin{figure}
    \centering
    \includegraphics[width = .5\textwidth]{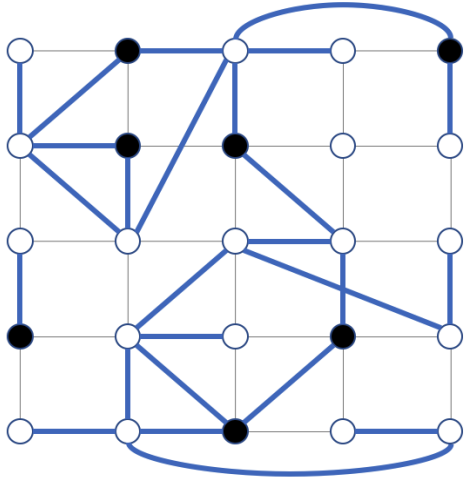}
    \caption{An illustration of the spreading network with $25$ agents on a $5\times 5$ grid network. The black nodes denote ``infected'' agents; The white nodes denote ``susceptible'' agents; The blue edges denote the active links at some time step.}
    \label{fig:spreadNetwork}
\end{figure}

We consider a spreading network with $n$ agents and an underlying graph $\mathcal{G}$. See Fig.~\ref{fig:spreadNetwork} for an illustration of $n = wh$ agents on a $w \times h$ grid network. For each agent $i$, the local state/action space is given by $\mathcal{S}_i = \{0, 1\}$ and $\mathcal{A}_i = \{0, 1\}$. To make the discussion more concrete, in the following we present the spreading network model in the context of SIS epidemic network. This version of the SIS model has been studied in, for example, \cite{ruhi2016improved}. Our setting is more general and can be generalized to other types of spreading networks like opinion networks, social networks, etc.
At time step $t$, the local state $s_i(t) = 0$ means agent $i$ is ``susceptible'', while the local state $s_i(t) = 1$ means the agent $i$ is ``infected''. By taking action $a_i(t) = 1$, agent $i$ can suppress its infection probability at the expense of incurring an action cost. In the meantime, agent $i$ will incur an infection cost if $s_i(t) = 1$. %
The interaction among agents is modeled by a set of undirected links, where two agents can affect each other if they are connected by a link. To model the influence of physical distance on the pattern of social contact, we assume the short range links occur more frequently than long range links. An illustration of the spreading network is shown in Fig.~\ref{fig:spreadNetwork} (a), where the black nodes denote the agents with state $1$; the white nodes denote the agents with state $0$; the blue edges denote the set of active links at some time step.

Mathematically, the model can be described as follows. At each time step $t$, each agent $i$ can decide her/his local action $a_i(t)$ based on the information of local states in the 1-hop neighborhood $N_i^1$, i.e., $\beta = 1$.
The local reward $r_i(t)$ is a function of the local state $s_i(t)$ and the local action $a_i(t)$, i.e., $L_t^r$ is static and only contains self loops.
Specifically, we define
\begin{equation*}
    r_i(t) = - c^{(a)}_i  \mathbf{1}(a_i(t) = 1) - c^{(s)}_i \mathbf{1}(s_i(t)= 1),
\end{equation*}
where $\left(c^{(s)}_i, c^{(a)}_i\right)$ are parameters associated with agent $i$ and can be different among agents. As mentioned earlier, $c^{(s)}_i$ penalizes the agent for being ``infected'', while $c^{(a)}_i$ is the cost of taking epidemic control measure. The stage reward is the sum of these two costs.

To describe the state transition rule, we first define the way the active link set $L_t^s$ is generated: independently for each pair of agents $(i, j) \in \mathcal{N} \times \mathcal{N}$ with $i\neq j$, with probability $2^{-d_\mathcal{G}(i, j)}$, we include edges $(i, j)$ and $(j, i)$ in the set $L_t^s $; otherwise, neither edge is included in the set, i.e. $(i, j),(j, i) \not\in L_t^s$. Given $L_t^s$, the next local state $s_i(t + 1)$ is sampled from a distribution that depends on the local states in $N_i(L_t^s)$. Specifically, define the quantities
\begin{align*}
    n_i(t) &= \abs{\{j \mid j \in N_i(L_t)\setminus \{i\}, s_j(t) = 1, a_j(t) = 0\}},\\
    m_i(t) &= \abs{\{j \mid j \in N_i(L_t)\setminus \{i\}, s_j(t) = 1, a_j(t) = 1\}}.
\end{align*}
Then, the probability that $s_i(t + 1) = 0$ is given by
\begin{equation*}
    P(s_i(t+1) = 0 \mid s_{N_i(L_t)}, a_{N_i(L_t)}) =\begin{cases}
    p_i^{(r)} & \text{ if } s_i(t) = 1;\\
    \left(1 - p^{(h)}_i\right)^{n_i(t)} \left(1 - p^{(m)}_i\right)^{m_i(t)}& \text{ if } s_i(t) = 0, a_i(t) = 1;\\
    \left(1 - p^{(m)}_i\right)^{n_i(t)} \left(1 - p^{(l)}_i\right)^{m_i(t)}& \text{ if } s_i(t) = 0, a_i(t) = 0,
    \end{cases}
\end{equation*}
where $\left(p_i^{(r)}, p_i^{(h)}, p_i^{(m)}, p_i^{(l)}\right)$ are parameters associated with agent $i$ and can be different among agents. Due to control actions, we assume $p_i^{(h)} > p_i^{(m)} > p_i^{(l)}$.
This provides the transition rule, and the underlying intuition is that the local state of agent $i$ turns from ``infected'' ($s_i(t)=1$) to ``susceptible'' ($s_i(t+1) = 0$) with a fixed recovering probability $p_i^{(r)}$; the probability that agent $i$ turns from ``susceptible'' ($s_i(t) = 0$) to ``infected'' ($s_i(t+1) = 1$) depends on the number of neighboring agents in the active link set that are already infected, and further, whether agent $i$ or the nearby agents $j$ take epidemic control measures ($a_i(t) = 1, a_j(t) = 1$) or not. Roughly speaking, the more nearby infected agents, the more likely agent $i$ will become infected; however, if epidemic control measures are taken by agent $i$ and nearby agents in $N_i(L_t^s)$, the probability of agent $i$ getting infected will be smaller.

\begin{figure}
    \centering
    \includegraphics[width = .5\textwidth]{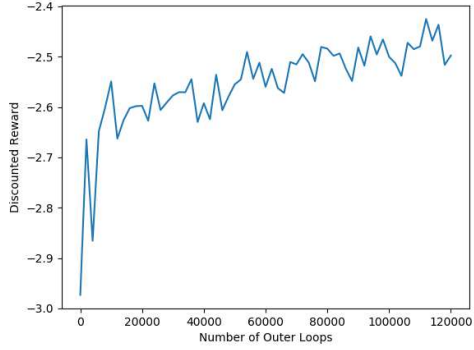}
    \caption{Discounted reward in the training process. $5\times 5$ grid.}
    \label{fig:w5h5-120000-spread}
\end{figure}

We run the Scalable Actor Critic algorithm with parameter $\kappa = 1$ to learn a localized stochastic policy in the case $(h, w) = (5, 5)$ (Fig.~\ref{fig:w5h5-120000-spread}). For each agent $i$, parameters $\left(c^{(s)}_i, c^{(a)}_i, p_i^{(r)}, p_i^{(h)}\right)$ are sampled independently from the distribution
\[c^{(s)}_i \sim U[1.0, 3.0], c^{(a)}_i \sim U[0.01, 0.20], p_i^{(r)} \sim U[0.1, 0.5], p_i^{(h)} \sim U[0.5, 0.9],\]
and we set $p_i^{(m)} = p_i^{(h)}/4, p_i^{(l)} = p_i^{(m)}/4$. At time step $0$, for each $i \in \mathcal{N}$, we initialize local state $s_i(0)$ to be $1$ with probability $0.3$.

\section{Stochastic Networked MARL}

\subsection{Proof of Theorem~\ref{lemma:info-spread}}\label{appendix:info-spread}
For ease of exposition, let $A, B$ be two subsets of the agent set $\mathcal{N}$ and we use $A \xrightarrow{\tau} B$ to denote the event that there exists a chain
\[j_0^a \xrightarrow{L_0^s} j_1^s \xrightarrow{L^a} j_1^a \xrightarrow{L_1^s} \cdots \xrightarrow{L_{\tau - 1}^s} j_\tau^s \xrightarrow{L^a} j_\tau^a,\]
whose head and tail satisfies $j_0^a \in A$ and $j_\tau^a \in B$.

Given a sequence of active link sets $\{L_t^s\}_{t=0}^\infty$ and under fixed global policy $\theta$, we say the information at set $A \subseteq \mathcal{N}$ spread to another set $B \subseteq \mathcal{N}$ in $\tau$ time steps (denoted by $I(A) \xrightarrow{\tau} I(B)$) if there exists $(s, a)$ and $(s', a')$ such that $(s_{\mathcal{N}\setminus A}, a_{\mathcal{N}\setminus A}) = (s'_{\mathcal{N}\setminus A}, a'_{\mathcal{N}\setminus A})$ and the distribution of $(s_B(\tau), a_B(\tau))$ given $(s(0), a(0)) = (s, a)$ is different with that given $(s(0), a(0)) = (s', a')$.

We show by induction that $I(A) \xrightarrow{\tau} I(B)$ happens only if $A \xrightarrow{\tau} B$ happens.

If $\tau = 0$, since $I(A) \xrightarrow{0} I(B)$, we see that $A \cap B \not = \emptyset$. Therefore, we can let $j_0^a$ be any agent in $A \cap B$. Hence we also have $A \xrightarrow{0} B$.

Suppose the statement holds for $\tau = t$. When $\tau = t + 1$, suppose that $I(A) \xrightarrow{t + 1} I(B)$. 
Define sets
\[B' := \{j \in \mathcal{N} \mid \exists k\in B, s.t. j \xrightarrow{L^a} k\}, B'' := \{j \in \mathcal{N} \mid \exists k\in B', s.t. j \xrightarrow{L_t^s} k\}.\]
Notice that $B \subseteq B' \subseteq B''$. By the definition of transition probability and policy dependence, we know that the distribution of $a_B(t + 1)$ is decided by $s_{B'}(t + 1)$, and the distribution of $s_{B'}(t + 1)$ is decided by $(s_{B''}(t), a_{B''}(t))$. Therefore, we must have $I(A) \xrightarrow{t} I(B'')$. By the induction hypothesis, we have $A \xrightarrow{t} B''$, which further implies $A \xrightarrow{t + 1} B$. This finishes the induction.

Given a sequence of active link sets $\{(L_t^s, L_t^r)\}$, we use $\pi_{t, i}$ to denote the distribution of\\ $\left(s_{N_i(L_t^r)}(t), a_{N_i(L_t^r)}(t)\right)$ given that $(s(0), a(0)) = (s, a)$; we use $\pi_{t, i}'$ to denote the distribution of $\left(s_{N_i(L_t^r)}(t), a_{N_i(L_t^r)}(t)\right)$ given that $(s(0), a(0)) = (s', a')$. We notice that $\pi_{t, i} \not = \pi_{t,i}'$ happens only if $I(N_{-i}^\kappa) \xrightarrow{t} I(N_i(L_t^r))$, which is true only if $N_{-i}^\kappa \xrightarrow{t} N_i(L_t^r)$. Recall that $X_i(\kappa)$ is defined as the smallest $t$ such that $N_{-i}^\kappa \xrightarrow{t} N_i(L_t^r)$ holds. Hence, we obtain that
\begin{align*}
    &\abs{Q_i^\theta(s, a) - Q_i^\theta(s', a')}\\
    \leq{}& \mathbb{E}_{\{(L_t^s, L_t^r)\}} \sum_{t=0}^\infty \abs{\gamma^t \mathbb{E}_{\pi_{t, i}}r_i(s_{N_i(L_t^r)}, a_{N_i(L_t^r)}) - \gamma^t \mathbb{E}_{\pi_{t, i}'}r_i(s_{N_i(L_t^r)}, a_{N_i(L_t^r)})}\\
    \leq{}& \mathbb{E}_{\{(L_t^s, L_t^r)\}} \sum_{t = X_i(\kappa)}^\infty \abs{\gamma^t \mathbb{E}_{\pi_{t, i}}r_i(s_{N_i(L_t^r)}, a_{N_i(L_t^r)}) - \gamma^t \mathbb{E}_{\pi_{t, i}'}r_i(s_{N_i(L_t^r)}, a_{N_i(L_t^r)})}\\
    \leq{}& \frac{1}{1 - \gamma} \mathbb{E}\left[\gamma^{X_i(\kappa)}\right],
\end{align*}
where we use the definition of $X_i(\kappa)$ in the second step.

\subsection{Proof of Corollary \ref{lemma:exponential-decay}}\label{appendix:exponential-decay}
Given a sequence of active link sets $\{(L_t^s, L_t^r)\}$, let $t = X_i(\kappa)$. By the definition of $X_i(\kappa)$, we assume that a chain of agents
\[j_0^a \xrightarrow{L_0^s} j_1^s \xrightarrow{L^a} j_1^a \xrightarrow{L_1^s} \cdots \xrightarrow{L_{t-1}^s} j_t^s \xrightarrow{L^a} j_t^a\]
satisfies $j_0^a \in N_{-i}^\kappa$ and $j_t^a \xrightarrow{L_{t}^r} i$.

By the triangle inequality and the assumptions of Lemma \ref{lemma:exponential-decay}, we obtain that
\begin{align*}
    d_\mathcal{G}(j_0^a, i) &\leq \sum_{\tau = 0}^{t - 1} \left(d_{\mathcal{G}}(j_\tau^a, j_{\tau + 1}^s) + d_{\mathcal{G}}(j_{\tau + 1}^s, j_{\tau + 1}^a)\right) + d_{\mathcal{G}}(j_t^a, i)\\
    &\leq t(\beta + \alpha_1) + \alpha_2.
\end{align*}
Therefore, we see that $t$ is lower bounded by $\frac{\kappa - \alpha_2}{\beta + \alpha_1}$, which also gives a lower bound of $X_i(\kappa)$.

\subsection{Proof of Theorem \ref{lemma:sub-exponential-decay}}\label{appendix:sub-exponential-decay}

To simplify notation, we adopt the same notations as in the proof of Theorem \ref{lemma:info-spread} (Appendix \ref{appendix:info-spread}). Specifically, recall that we use $A \xrightarrow{\tau} B$ to denote the event that there exists a chain
\[j_0^a \xrightarrow{L_0^s} j_1^s \xrightarrow{L^a} j_1^a \xrightarrow{L_1^s} \cdots \xrightarrow{L_{\tau - 1}^s} j_\tau^s \xrightarrow{L^a} j_\tau^a,\]
whose head and tail satisfies $j_0^a \in A$ and $j_\tau^a \in B$. We will use $\partial N_i^\kappa$ to denote the set of neighbors whose distance to $i$ is $\kappa$, i.e., $\partial N_i^\kappa := \{j \in \mathcal{N} \mid d_{\mathcal{G}}(i, j) = \kappa\} = N_i^\kappa \setminus N_i^{\kappa - 1}$. 
Define $a_\kappa := \mathbb{E}\left[\gamma^{X_i(\kappa - 1)}\right]$. Define function $cat$ (concatenation) such that for a pair of active link sets $(L^s, L^a)$, $(x, y) \in cat(L^s, L^a)$ if and only if $\exists z \in \mathcal{N}$ such that $x \xrightarrow{L^s} z \xrightarrow{L^a} y$.

Before proving Theorem~\ref{lemma:sub-exponential-decay}, we first give an upper bound for the sum of an infinite sequence $\{poly(k + i)\cdot \nu^i\}_{i \in \mathbb{N}}$, where $\nu < 1$ is a positive constant. This result is helpful for showing an upper bound of $P(N_{-i}^\kappa \to N_i^j)$.

\begin{lemma}\label{lemma:sum-poly-times-exp}
If $m \in \mathbb{N}^*$ and $0 < \nu < 1$ are constants, for all $k \geq \frac{2m}{\ln(1/\nu)}$, we have
\[\sum_{i = 0}^\infty (k + i)^m \nu^i \leq \frac{1}{1 - \sqrt{\nu}}\cdot k^m.\]
\end{lemma}
\begin{proof}[Proof of Lemma \ref{lemma:sum-poly-times-exp}]
Define function $f: \mathbb{R}^+\cup\{0\} \to \mathbb{R}^+$ as
\[f(t) = (k + t)^m \cdot \nu^{t/2}.\]
The derivative of function $f$ is given by
\[f'(t) = (k + t)^{m - 1}\cdot \nu^{t/2} \left(m + \frac{1}{2}\ln \nu \cdot (k + t)\right).\]
Since $k \geq \frac{2m}{\ln (1/\nu)}$, $f'(t) \leq 0$ holds for all $t \geq 0$, hence we have $f(t) \leq f(0) = k^m$.

Therefore, we obtain that
\begin{align*}
    \sum_{i = 0}^\infty (k + i)^m \nu^i &\leq \sum_{i = 0}^\infty f(i) \cdot \nu^{i/2}\\
    &\leq k^m \sum_{i = 0}^\infty \nu^{i/2}\\
    &\leq \frac{1}{1 - \sqrt{\nu}}\cdot k^m.
\end{align*}
\end{proof}

Now we come back to the proof of Theorem \ref{lemma:sub-exponential-decay}. 

By union bound, we derive an upper bound of the probability that a link $(x, y)$ is in $cat(L^s, L^a)$. Suppose $d \in \mathbb{N}$ is constant that satisfies $d_{\mathcal{G}}(x, y) \geq d$, and the probability $P$ is taken over $(L^s, L^r) \sim \mathcal{D}$:
\begin{align}
    P\left((x, y) \in cat(L^s, L^a)\right)
    ={}& P\left(\exists z \in \mathcal{N}, (x, z) \in L^s \land (z, y) \in L^a\right)\nonumber\\
    \leq{}& \sum_{z: d_\mathcal{G}(z, y) \leq \beta} P\left((x, z) \in L^s\right)\nonumber\\
    \leq{}& c_0 (\beta + 1)^{n_0 + 1}\cdot c \lambda^{d - \beta}\nonumber\\
    ={}& c_g \lambda^d, \label{lemma:sub-exponential-decay:e03}
\end{align}
where constant $c_g$ is defined as $c_0 c (\beta + 1)^{n_0 + 1} \lambda^{-\beta}$.  

By the assumption on the size of $\kappa$-hop neighborhood, we know that for some constant $c_0$ and $n_0 \in \mathbb{N}^*$, $\abs{\partial N_i^\kappa} \leq c_0 (\kappa + 1)^{n_0}$ holds for all $\kappa \geq 1$. Let $n_1 := 2 n_0$. 
With the help of Lemma \ref{lemma:sum-poly-times-exp}, we show that for some constant $c_2 > 0$, $P\left(N_{-i}^{\kappa - 1} \xrightarrow{1} \partial N_i^j\right)$ is upper bounded by $c_2 (\kappa + 1)^{n_1} \lambda^{\kappa - j}$ for all $j \leq \kappa - 1$ when $\kappa \geq \frac{2n_0}{\ln(1/\lambda)}$:
\begin{subequations}\label{lemma:sum-poly-times-exp:e1}
\begin{align}
    P\left(N_{-i}^{\kappa - 1} \xrightarrow{1} \partial N_i^j\right) &\leq P\left(\exists x \in N_{-i}^{\kappa - 1}, y \in \partial N_i^j \text{ s.t. } (x, y) \in cat(L^s, L^a)\right)\label{lemma:sum-poly-times-exp:e1:s0}\\
    &\leq \sum_{q = 0}^\infty P\left(\exists x \in \partial N_i^{\kappa + q}, y \in \partial N_i^j \text{ s.t. } (x, y) \in cat(L^s, L^a)\right)\label{lemma:sum-poly-times-exp:e1:s1}\\
    &\leq \sum_{q = 0}^\infty\sum_{x \in \partial N_i^{\kappa + q}, y \in \partial N_i^j} P\left((x, y) \in cat(L^s, L^a)\right)\label{lemma:sum-poly-times-exp:e1:s2}\\
    &\leq \sum_{q = 0}^\infty\sum_{x \in \partial N_i^{\kappa + q}, y \in \partial N_i^j} c_g \lambda^{(\kappa + q - j)}\label{lemma:sum-poly-times-exp:e1:s3}\\
    &\leq c_g \lambda^{\kappa - j}\sum_{q = 0}^\infty \abs{\partial N_i^{\kappa + q}} \cdot \abs{\partial N_i^j}\cdot \lambda^{q}\nonumber\\
    &\leq c_g c_0^2 (\kappa + 1)^{n_0} \lambda^{\kappa - j}\sum_{q = 0}^\infty (\kappa + q + 1)^{n_0} \lambda^{q}\label{lemma:sum-poly-times-exp:e1:s4}\\
    &\leq c_2 (\kappa + 1)^{n_1} \lambda^{\kappa - j},\label{lemma:sum-poly-times-exp:e1:s5}
\end{align}
\end{subequations}
where we use the definition of $N_{-i}^{\kappa - 1} \xrightarrow{1} \partial N_i^j$ in \eqref{lemma:sum-poly-times-exp:e1:s0}; we use union bound in \eqref{lemma:sum-poly-times-exp:e1:s1} and \eqref{lemma:sum-poly-times-exp:e1:s2}; we use the fact that $d_{\mathcal{G}}(x, y) \geq \kappa + q - j, \forall x \in \partial N_i^{\kappa + q}, y \in \partial N_i^j$ and \eqref{lemma:sub-exponential-decay:e03} in \eqref{lemma:sum-poly-times-exp:e1:s3}; we use the bounds $\abs{\partial N_i^j} \leq c_0 j^{n_0} \leq c_0 \kappa^{n_0}$ and $\abs{\partial N_i^{\kappa + q}} \leq c_0 (\kappa + q)^{n_0}$ in \eqref{lemma:sum-poly-times-exp:e1:s4}; we define $c_2 := \frac{c_g c_0^2}{1 - \sqrt{\lambda}}$ and use Lemma \ref{lemma:sum-poly-times-exp} in \eqref{lemma:sum-poly-times-exp:e1:s5}.

Let constants $c_3$ and $q$ be defined as
\begin{align*}
    c_3 &:= \frac{1}{2}\sqrt[4]{\lambda}(1 - \sqrt{\lambda})\left(\frac{1}{\sqrt{\gamma}} - 1\right),\\
    q &:= \frac{1}{\ln(1/\lambda)}\max\{(\ln c_2 - \ln c_3 - 2\ln (1 - \sqrt{\gamma})), (2 n_1 + 4)\},
\end{align*}
and define function $p(\kappa) := \left[q(1 + \ln (\kappa + 1))\right] + 1$. We can find $\kappa_0 \in \mathbb{Z}^+$ such that $p(\kappa) \geq \kappa$ for all $\kappa \leq \kappa_0$, and $p(\kappa) > \kappa$ for all $\kappa > \kappa_0$.

Let $\rho$ be a constant such that $1 > \rho > \max\{\gamma^{1/(2q)}, \sqrt[4]{\lambda}\}$. Let $C := \rho^{- \max\{q + 1, \frac{2n_0}{\ln(1/\lambda)}\}}$. Recall that we define $a_\kappa := \mathbb{E}\left[\gamma^{X_i(\kappa - 1)}\right]$, where $X_i(\kappa - 1)$ denotes the smallest $t$ such that $N_{-i}^{\kappa - 1} \xrightarrow{t} N_i(L_t^r)$ holds. Now we show by induction that
\begin{equation}\label{lemma:sub-exponential-decay:target}
    a_\kappa \leq C \rho^{\kappa /(1 + \ln (\kappa + 1))}, \forall \kappa \geq 1.
\end{equation}
Since $a_\kappa \leq 1$, \eqref{lemma:sub-exponential-decay:target} clearly holds when $\kappa \leq \kappa_0$. To see this, recall that we have $\kappa \leq p(\kappa)$ and $C \geq \rho^{-(q+1)}$ by definition, thus the right hand side of \eqref{lemma:sub-exponential-decay:target} can be lower bounded by
\[C \rho^{\kappa /(1 + \ln (\kappa + 1))} \geq \rho^{-(q + 1)}\cdot \rho^{p(\kappa) /(1 + \ln (\kappa + 1))} \geq \rho^{-(q + 1)}\cdot \rho^{q + 1} = 1.\]

When $\kappa > \kappa_0$, we have $\kappa > p(\kappa)$. Recall that $a_\kappa := \mathbb{E}\left[\gamma^{X_i(\kappa - 1)}\right]$. Notice that $X_i(\kappa - 1) = 0$ if and only if $N_{-i}^{\kappa - 1}\cap N_i(L_0^r) \not = \emptyset$. To simplify the notation, we denote the event $N_{-i}^{\kappa - 1}\cap N_i(L_0^r) \not = \emptyset$ by $E_0$. Using this and the idea of dynamic programming, we see that
\begin{align}\label{lemma:sub-exponential-decay:e1-0}
    a_\kappa \leq{}& \gamma \left(P\{\left(\neg N_{-i}^{\kappa - 1} \xrightarrow{1} N_i^{\kappa - 1}\right)\land \neg E_0\} a_\kappa + \sum_{j = 0}^{\kappa - 1} P\{\left(N_{-i}^\kappa \xrightarrow{1} \partial N_i^j\right)\land \left(\neg N_{-i}^\kappa \xrightarrow{1} N_i^{j - 1}\right)\land \neg E_0\} a_j\right)\nonumber\\
    &+ P(E_0)\nonumber\\
    \leq{}& \gamma \left(P\{\neg N_{-i}^{\kappa - 1} \xrightarrow{1} N_i^{\kappa - 1}\} a_\kappa + \sum_{j = 0}^{\kappa - 1} P\{\left(N_{-i}^\kappa \xrightarrow{1} \partial N_i^j\right)\land \left(\neg N_{-i}^\kappa \xrightarrow{1} N_i^{j - 1}\right)\} a_j\right) + P(E_0),
\end{align}
where the probability $P$ are taken over $(L_0^s, L_0^r) \sim D$.

Since $\kappa \geq p(\kappa) \geq q \geq \frac{2n_1}{\ln(1/\lambda)} \geq \frac{2n_0}{\ln(1/\lambda)}$, by Lemma \ref{lemma:sum-poly-times-exp}, we see that
\[P(E_0) = P\{\exists j \in N_{-i}^{\kappa - 1} \text{ s.t. } (j, i) \in L^r\} \leq \sum_{q = 0}^\infty c c_0 (\kappa + q + 1)^{n_0} \lambda^{\kappa + q} \leq \frac{c c_0}{1 - \sqrt{\lambda}} (\kappa + 1)^{n_0 + 1} \lambda^\kappa.\]
Substituting this into \eqref{lemma:sub-exponential-decay:e1-0} and rearranging the terms gives
\begin{align}\label{lemma:sub-exponential-decay:e1}
    \left(1 - \gamma P\{\neg N_{-i}^{\kappa - 1} \xrightarrow{1} N_i^{\kappa - 1}\}\right) a_\kappa \leq{}& \gamma \sum_{j = \kappa - p(\kappa) + 1}^{\kappa - 1} P\{\left(N_{-i}^{\kappa - 1} \xrightarrow{1} \partial N_i^j\right)\land \left(\neg N_{-i}^{\kappa - 1} \xrightarrow{1} N_i^{j - 1}\right)\} a_j\nonumber\\
    &+ \gamma \sum_{j = 0}^{\kappa - p(\kappa)} P\{\left(N_{-i}^{\kappa - 1} \xrightarrow{1} \partial N_i^j\right)\land \left(\neg N_{-i}^{\kappa - 1} \xrightarrow{1} N_i^{j - 1}\right)\} a_j\nonumber\\
    &+ \frac{c c_0}{1 - \sqrt{\lambda}} (\kappa + 1)^{n_0 + 1} \lambda^\kappa.
\end{align}
For simplicity, we define $\rho_\kappa := \rho^{1/(1 + \ln (\kappa + 1))}$. By the induction assumption, we have that
\[a_j \leq C \rho^{j/(\ln (j + 1) + 1)} \leq C \rho^{j/(\ln (\kappa + 1) + 1)} = C \rho_\kappa^j.\]
Substituting this into \eqref{lemma:sub-exponential-decay:e1} gives that
\begin{align}\label{lemma:sub-exponential-decay:e1-1}
    \left(1 - \gamma P\{\neg N_{-i}^{\kappa - 1} \xrightarrow{1} N_i^{\kappa - 1}\}\right) a_\kappa \leq{}& C \gamma \sum_{j = \kappa - p(\kappa) + 1}^{\kappa - 1} P\{\left(N_{-i}^{\kappa - 1} \xrightarrow{1} \partial N_i^j\right)\land \left(\neg N_{-i}^{\kappa - 1} \xrightarrow{1} N_i^{j - 1}\right)\} \rho_\kappa^j\nonumber\\
    &+ C \gamma \sum_{j = 0}^{\kappa - p(\kappa)} P\{\left(N_{-i}^{\kappa - 1} \xrightarrow{1} \partial N_i^j\right)\land \left(\neg N_{-i}^{\kappa - 1} \xrightarrow{1} N_i^{j - 1}\right)\} \rho_\kappa^j\nonumber\\
    &+ \frac{c_0}{1 - \sqrt{\lambda}} (\kappa + 1)^{n_0 + 1} \lambda^\kappa.
\end{align}

By the definition of $p(\kappa)$ and $q$, we see that
\begin{align*}
    \lambda^{-p(\kappa)} \geq \lambda^{-q(1 + \ln(\kappa + 1))} = \lambda^{-q} \cdot (\kappa + 1)^{q\ln(1/\lambda)} \geq \frac{c_2}{c_3 (1 - \sqrt{\gamma})^2}\cdot (\kappa + 1)^{n_1} \geq \frac{c_2}{c_3 (1 - \gamma)}\cdot (\kappa + 1)^{n_1}.
\end{align*}
Therefore, we obtain the upper bound
\begin{align*}
    P\{\left(N_{-i}^{\kappa - 1} \xrightarrow{1} \partial N_i^j\right)\land \left(\neg N_{-i}^\kappa \xrightarrow{1} N_i^{j - 1}\right)\} &\leq P\{N_{-i}^{\kappa - 1} \xrightarrow{1} \partial N_i^j\}\\
    &\leq c_2 (\kappa + 1)^{n_1}\lambda^{(\kappa - j)}\\
    &\leq (1 - \gamma) c_3 \lambda^{(\kappa - p(\kappa) - j)}.
\end{align*}
Using this and divide both sides of \eqref{lemma:sub-exponential-decay:e1-1} by $\left(1 - \gamma P\{\neg N_{-i}^\kappa \xrightarrow{1} N_i^{\kappa - 1}\}\right)$, we see that
\begin{align}\label{lemma:sub-exponential-decay:e2}
    a_\kappa \leq{}& \gamma \left(C \rho_\kappa^{\kappa - p(\kappa) + 1} + C c_3 (\rho_\kappa^{\kappa - p(\kappa)} + \lambda^{1} \cdot \rho_\kappa^{\kappa - p(\kappa) - 1} + \lambda^2 \cdot \rho_\kappa^{\kappa - p(\kappa) - 2} + \cdots )\right)\nonumber\\
    &+ \frac{c_0}{(1 - \gamma)(1 - \sqrt{\lambda})} (\kappa + 1)^{n_0 + 1} \lambda^\kappa,
\end{align}
where we also use the fact that
\[\sum_{j = \kappa - p + 1}^{\kappa - 1} P\{\left(N_{-i}^{\kappa - 1} \xrightarrow{1} \partial N_i^j\right)\land \left(\neg N_{-i}^{\kappa - 1} \xrightarrow{1} N_i^{j - 1}\right)\} \leq 1 - \gamma P\{\neg N_{-i}^{\kappa - 1} \xrightarrow{1} N_i^{\kappa - 1}\}.\]

By the definition of $p(\kappa), q$ and $c_2$, we have that
\[\lambda^{\frac{\kappa}{4}} \leq \lambda^{\frac{p(\kappa)}{4}} \leq (\kappa + 1)^{- \frac{q \ln(1/\lambda)}{4}} \leq (\kappa + 1)^{- n_0 - 1}\]
and
\[\lambda^{\frac{\kappa}{2}} \leq \lambda^{\frac{p(\kappa)}{2}} \leq \lambda^{\frac{q}{2}} \leq \frac{(1 - \sqrt{\gamma})(1 - \gamma)(1 - \sqrt{\lambda})}{2 c_0},\]
which implies
\begin{equation}\label{lemma:sub-exponential-decay:e3-0}
    \lambda^{\frac{3\kappa}{4}} \leq \frac{(1 - \sqrt{\gamma})(1 - \gamma)(1 - \sqrt{\lambda})}{2 c_0 (\kappa + 1)^{n_0 + 1}}.
\end{equation}

Dividing both sides of \eqref{lemma:sub-exponential-decay:e2} by $C \rho_\kappa^{\kappa}$ gives that
\begin{subequations}\label{lemma:sub-exponential-decay:e3}
\begin{align}
    \frac{a_\kappa}{C \rho_\kappa^\kappa} &\leq \gamma \left(\frac{1}{\rho_\kappa^{p(\kappa) - 1}} + \frac{c_3}{\rho_\kappa^{p(\kappa)}} \cdot \frac{1}{1 - (\lambda/\rho_\kappa)}\right) + \frac{c_0}{(1 - \gamma)(1 - \sqrt{\lambda})} (\kappa + 1)^{n_0 + 1} \lambda^{\frac{3\kappa}{4}} \label{lemma:sub-exponential-decay:e3:s0}\\
    &\leq \gamma \left(\frac{1}{\rho^q} + \frac{1}{\rho^{q + 1}}\cdot \frac{c_3}{1 - \sqrt{\lambda}}\right) + \frac{1}{2}(1 - \sqrt{\gamma})\label{lemma:sub-exponential-decay:e3:s1}\\
    &= \frac{\gamma}{\rho^q}\left(1 + \frac{c_3}{\rho (1 - \sqrt{\lambda})}\right) + \frac{1}{2}(1 - \sqrt{\gamma})\nonumber\\
    &\leq \sqrt{\gamma}\cdot \frac{1}{2}\left(1 + \frac{1}{\sqrt{\gamma}}\right) + \frac{1}{2}(1 - \sqrt{\gamma})\label{lemma:sub-exponential-decay:e3:s2}\\
    &= 1,\nonumber
\end{align}
\end{subequations}
where we use $\rho_\kappa = \rho^{1/(1 + \ln \kappa)} \geq \rho \geq \sqrt[4]{\lambda}$ in \eqref{lemma:sub-exponential-decay:e3:s0}; we use $\rho_\kappa \geq \sqrt[4]{\lambda}$, $p = \left[q(1 + \ln \kappa)\right] + 1$, and \eqref{lemma:sub-exponential-decay:e3-0} in \eqref{lemma:sub-exponential-decay:e3:s1}; we use $c_3 = \sqrt{\lambda}(1 - \sqrt{\lambda})(\sqrt{\gamma} - 1) \leq \rho(1 - \sqrt{\lambda})(\sqrt{\gamma} - 1)$ and $\rho \geq \gamma^{1/(2q)}$ in \eqref{lemma:sub-exponential-decay:e3:s2}.

\subsection{Proof of Theorem \ref{coro:critic-SAC}}\label{appendix:critic-SAC}
In the Critic part of Algorithm \ref{alg:SAC_Outline}, since the policy is fixed to be $\theta(m)$, the pair $(s, a)$ can be viewed as the state of a Markov chain $\mathcal{C}$, and $Q^{\theta(m)}(s, a)$ in the original MDP corresponds to the value function $V^*((s, a))$ on $\mathcal{C}$. Define the state aggregation map $h$ such that $h((s, a)) = (s_{N_i^\kappa}, a_{N_i^\kappa})$. By the $\mu$-decay property, we see that if $h((s, a)) = h((s', a'))$, then
\[\abs{V^*((s, a)) - V^*((s', a'))} = \abs{Q^{\theta(m)}(s, a) - Q^{\theta(m)}(s', a')} \leq \mu(\kappa).\]
Note that Assumption \ref{network-assp:geometric-mixing} implies that Assumption \ref{assump:geometric-mixing} holds for $\mathcal{C}$. Thus, we can apply Theorem \ref{thm:TD-indicator-feature-finite} to finish the proof of Theorem \ref{coro:critic-SAC}.

\subsection{Proof of Theorem \ref{coro:sample-complexity}}\label{appendix:sample-complexity}
 Before showing Theorem \ref{coro:sample-complexity}, we first state a theorem concerning the actor part of Algorithm \ref{alg:SAC_Outline}. The proof is deferred to Appendix \ref{appendix:thm:actor-SAC}.

\begin{theorem}\label{thm:actor-SAC}
Under the same assumption as Theorem \ref{coro:sample-complexity}, suppose inner loop length $T$ is sufficiently large such that $T + 1 \geq \log_\gamma ((1 - \gamma)\mu(\kappa))$ and with probability at least $1 - \frac{\delta}{2}$, the following inequality holds for all agents $i \in \mathcal{N}$:
{\small\[\sup_{m \leq M - 1}\sup_{(s, a)\in \mathcal{S}\times \mathcal{A}}\abs{Q_i^{\theta(m)}(s, a) - \hat{Q}^T (s_{N_i^\kappa}, a_{N_i^\kappa})} \leq \frac{\iota \mu(\kappa)}{1 - \gamma},\]}
where $\iota$ is a positive constant. Suppose the actor step size satisfies $\eta_m = \frac{\eta}{\sqrt{m + 1}}$ with $\eta \leq \frac{1}{4W'}$. Define $C_M := \frac{2}{\eta (1 - \gamma)} + \frac{8W^2 \sqrt{\log M \log \frac{4}{\delta}} + 96W' W^2 \eta \log M}{(1 - \gamma)^4}.$ Then, with probability at least $1 - \delta$,
{\small\begin{equation}\label{equ:thm:actor-SAC}
    \frac{\sum_{m=0}^{M-1}\eta_m \norm{\nabla J(\theta(m))}^2}{\sum_{m=0}^{M-1}\eta_m} \leq \frac{C_M}{\sqrt{M + 1}} + \frac{2(2 + \iota)W^2 \mu(\kappa)}{(1 - \gamma)^4}.
\end{equation}}
\end{theorem}

As a remark, note that the left hand side of \eqref{equ:thm:actor-SAC} is a weighted average of the squared norm of the gradients $\nabla J(\theta(m))$. We say the algorithm has reached an $O(\epsilon)$-approximate stationary point if the left hand side of \eqref{equ:thm:actor-SAC} is in the order of $O(\epsilon)$.

Now we come back to the proof of Theorem \ref{coro:sample-complexity}. Let constant $\iota = 2$ in Theorem \ref{thm:actor-SAC}. By Theorem \ref{thm:actor-SAC}, to satisfy
\[\frac{\sum_{m=0}^{M-1}\eta_m \norm{\nabla J(\theta(m))}^2}{\sum_{m=0}^{M-1}\eta_m} \leq O(\epsilon),\]
it suffices to guarantee that
\[\frac{C_M}{\sqrt{M + 1}} = O(\epsilon), \text{ and } \frac{2(2 + \iota)W^2 \mu(\kappa)}{(1 - \gamma)^4} = O(\epsilon).\]
These can be satisfied by letting
\[M = \tilde{\Omega}\left(\epsilon^{-2} \left(\frac{(W')^2}{(1 - \gamma)^2} + \frac{W^4 (1 + \log(1/\delta))}{(1 - \gamma)^8}\right)\right), \mu(\kappa) = O\left(W^{-2} (1 - \gamma)^4 \epsilon \right).\]

To satisfy
\[\sup_{m \leq M - 1}\sup_{(s, a)\in \mathcal{S}\times \mathcal{A}}\abs{Q_i^{\theta(m)}(s, a) - \hat{Q}^T (s_{N_i^\kappa}, a_{N_i^\kappa})} \leq \frac{\iota \mu(\kappa)}{1 - \gamma},\]
with probability at least $1 - \frac{\delta}{2}$, by Corollary \ref{coro:critic-SAC}, it suffices to select $T$ such that
\begin{align*}
    &\frac{1}{\sqrt{T + t_0}}\cdot \frac{40H}{(1 - \gamma)^2}\sqrt{K_2 \log T\left(\log \left(\frac{4f(\kappa) K_2 T}{\delta}\right) + \log \log T \right)}\\
    &\quad + \frac{1}{T + t_0}\cdot \frac{8}{(1 - \gamma)^2}\max\{\frac{144 K_2 H \log T}{\sigma'(\kappa)} + C_3, 2K_2 \log T + t_0\}\\
    &\leq{} \frac{\mu(\kappa)}{1 - \gamma}.
\end{align*}
Recall that
\[\mu(\kappa) = O\left(W^{-2} (1 - \gamma)^4 \epsilon \right), H \geq \frac{2}{(1 - \gamma)\sigma'(\kappa)}, t_0 = \max(4H, 2K_2 \log{T}).\]
Hence the required number of inner loop is
\[T = \tilde{\Omega}\left(\frac{W^4 \left(K_2 (\log f(\kappa) + \log(1/\delta) + 1) + K_1\right)}{\epsilon^2 (1 - \gamma)^{12}\sigma'(\kappa)^2}\right).\]

\subsection{Proof of Theorem \ref{thm:actor-SAC}}\label{appendix:thm:actor-SAC}
While Theorem 5 in \cite{qu2019scalable} studies the error bound of Scalable Actor Critic as a whole, we want to decouple the effect of the inner loop and the outer loop in Theorem \ref{thm:actor-SAC}. Our proof of Theorem \ref{thm:actor-SAC} uses similar techniques with the proof in \cite{qu2019scalable}, but we extend the analysis to a  more general dependence model. 

According to Algorithm \ref{alg:SAC_Outline}, at iteration $m$, agent $i$ performs gradient ascent by
\[\theta_i(m + 1) = \theta_i(m) + \eta_m \hat{g}_i(m),\]
with step size $\eta_m = \frac{\eta}{\sqrt{m + 1}}$. The approximate local gradient $\hat{g}_i(m)$ is given by
\begin{equation*}
    \hat{g}_i(m) = \sum_{t=0}^T \gamma^t \frac{1}{n}\sum_{j \in N_i^\kappa}\hat{Q}_j^{m, T}\left(s_{N_j^\kappa}(t), a_{N_j^\kappa}(t)\right)\nabla_{\theta_i}\log \zeta_i^{\theta_i(m)}\left(a_i(t)\mid s_{N_i^{\beta}}(t)\right).
\end{equation*}
Recall that the true local gradient is given by
\begin{equation*}
    \nabla_{\theta_i} J(\theta(m)) = \sum_{t=0}^\infty \mathbb{E}_{s\sim \pi_t^{\theta(m)}, a\sim \zeta_i^{\theta(m)}(\cdot \mid s)}\gamma^t Q^{\theta(m)} (s, a)\nabla_{\theta_i} \log \zeta^{\theta_i(m)} \left(a_i(t)\mid s_{N_i^{\beta}}(t)\right),
\end{equation*}
where we use $\pi_t^\theta$ to denote the distribution of global state $s(t)$ under fixed policy $\theta$.

To bound $\norm{\hat{g}(m) - \nabla_{\theta} J(\theta(m))}$, we define intermediate quantities $g(m)$ and $h(m)$ whose $i$'th component is given by
\begin{align}
    g_i(m) &= \sum_{t=0}^T \gamma^t \frac{1}{n}\sum_{j \in N_i^\kappa}Q_j^{\theta(m)}\left(s(t), a(t)\right)\nabla_{\theta_i} \log \zeta_i^{\theta_i(m)} \left(a_i(t)\mid s_{N_i^{\beta}}(t)\right),\nonumber\\
    h_i(m) &= \sum_{t=0}^T \mathbb{E}_{s\sim \pi_t^{\theta(m)}, a\sim \zeta^{\theta(m)}(\cdot \mid s)} \gamma^t \frac{1}{n}\sum_{j \in N_i^\kappa}Q_j^{\theta(m)}\left(s, a\right)\nabla_{\theta_i} \log \zeta_i^{\theta_i(m)} \left(a_i(t)\mid s_{N_i^{\beta}}(t)\right).\nonumber
\end{align}

\begin{lemma}\label{lemma:actor-grad-bound}
We have almost surely, $\forall m \leq M$,
\[\max\left(\norm{\hat{g}(m)}, \norm{g(m)}, \norm{h(m)}, \norm{\nabla J(\theta(m))}\right) \leq \frac{W}{(1 - \gamma)^2}.\]
\end{lemma}

To show Lemma \ref{lemma:actor-grad-bound}, we only need to replace $\zeta_i^{\theta_i(m)} \left(a_i(t)\mid s_i(t)\right)$ by $\zeta_i^{\theta_i(m)} \left(a_i(t)\mid s_{N_i^{\beta}}(t)\right)$ in the proof of Lemma 17 in \cite{qu2019scalable}.

Notice that
\begin{equation*}
    \hat{g}(m) - \nabla J(\theta(m)) = e^1(m) + e^2(m) + e^3(m),
\end{equation*}
where
\[e^1(m) := \hat{g}(m) - g(m), e^2(m) := g(m) - h(m), e^3(m) := h(m) - \nabla J(\theta(m)).\]
To bound $\norm{\hat{g}(m) - \nabla J(\theta(m))}$, we only need to bound $e_1(m), e_2(m), e_3(m)$ separately.

\begin{lemma}\label{lemma:actor-e1-bound}
With probability at least $1 - \frac{\delta}{2}$, we have
\[\sup_{0\leq m \leq M - 1} \norm{e^1(m)} \leq \frac{\iota  W \mu(\kappa)}{(1 - \gamma)^2}.\]
\end{lemma}
\begin{proof}[Proof of Lemma \ref{lemma:actor-e1-bound}]
By the assumption that
\[\sup_{m \leq M - 1}\sup_{i \in \mathcal{N}}\sup_{(s, a)\in \mathcal{S}\times \mathcal{A}}\abs{Q_i^{\theta(m)}(s, a) - \hat{Q}^T (s_{N_i^\kappa}, a_{N_i^\kappa})} \leq \frac{\iota \cdot \mu(\kappa)}{1 - \gamma},\]
we have for all $m \leq M - 1$ and $i \in \mathcal{N}$,
\begin{equation*}
    \begin{aligned}
    &\norm{\hat{g}_i(m) - g_i(m)}\\
    \leq{}&\norm{\sum_{t=0}^T \gamma^t \frac{1}{n}\sum_{j \in N_i^\kappa}\left[\hat{Q}_j^{m, T}\left(s_{N_j^\kappa}(t), a_{N_j^\kappa}(t)\right) - Q_j^{\theta(m)}(s(t), a(t))\right]\nabla_{\theta_i}\log \zeta_i^{\theta_i(m)}\left(a_i(t)\mid s_{N_i^{\beta}}(t)\right)}\\
    \leq{}&\sum_{t=0}^T \gamma^t \frac{1}{n}\sum_{j \in N_i^\kappa}\abs{\hat{Q}_j^{m, T}\left(s_{N_j^\kappa}(t), a_{N_j^\kappa}(t)\right) - Q_j^{\theta(m)}(s(t), a(t))}\norm{\nabla_{\theta_i}\log \zeta_i^{\theta_i(m)}\left(a_i(t)\mid s_{N_i^{\beta}}(t)\right)}\\
    \leq{}& \sum_{t=0}^T \gamma^t \frac{\iota \cdot \mu(\kappa)}{1 - \gamma} W_i\\
    <{}& \frac{2\iota W_i \cdot \mu(\kappa)}{(1 - \gamma)^2}.
    \end{aligned}
\end{equation*}
Combining all $n$ dimensions finishes the proof.
\end{proof}

\begin{lemma}\label{lemma:actor-e2-bound}
With probability at least $1 - \frac{\delta}{2}$, we have
\[\abs{\sum_{m=0}^{M - 1}\eta_m \langle \nabla J(\theta(m)), e^2(m)\rangle} \leq \frac{2W^2}{(1 - \gamma)^4}\sqrt{2\sum_{m=0}^{M - 1}\eta_m^2 \log \frac{4}{\delta}}.\]
\end{lemma}
To show Lemma \ref{lemma:actor-e2-bound}, we only need to replace $\zeta_i^{\theta_i(m)} \left(a_i(t)\mid s_i(t)\right)$ by $\zeta_i^{\theta_i(m)} \left(a_i(t)\mid s_{N_i^{\beta}}(t)\right)$ in the proof of Lemma 19 in \cite{qu2019scalable}.

\begin{lemma}\label{lemma:actor-e3-bound}
When $T + 1 \geq \log_\gamma ((1 - \gamma) \mu(\kappa))$, we have almost surely
\[\norm{e^3(m)} \leq \frac{2 W \mu(\kappa)}{1 - \gamma}.\]
\end{lemma}
To show Lemma \ref{lemma:actor-e3-bound}, we only need to replace $\zeta_i^{\theta_i(m)} \left(a_i(t)\mid s_i(t)\right)$ with $\zeta_i^{\theta_i(m)} \left(a_i(t)\mid s_{N_i^{\beta}}(t)\right)$ and replace $c\rho^{\kappa + 1}$ with $\mu(\kappa)$ in the proof of Lemma 20 in \cite{qu2019scalable}.

Now we come back to the proof of Theorem \ref{thm:actor-SAC}. Using the identical steps with the proof of Theorem 5 in \cite{qu2019scalable}, we can obtain that (equation (44) in \cite{qu2019scalable})
\begin{equation}\label{thm:actor-SAC:e0}
    \sum_{m=0}^{M-1}\frac{1}{2}\eta_m \norm{\nabla J(\theta(m))}^2 \leq J(\theta(m)) - J(\theta(0)) - \sum_{m=0}^{M - 1}\eta_m \epsilon_{m, 0} + \sum_{m=0}^{M - 1}\eta_m \epsilon_{m, 1} + \sum_{m=0}^{M - 1}\eta_m^2 \epsilon_{m, 2},
\end{equation}
where
\begin{equation*}
    \begin{aligned}
    \epsilon_{m, 0} &= \langle \nabla J(\theta(m)), e^2(m)\rangle,\\
    \epsilon_{m, 1} &= \norm{\nabla J(\theta(m))}(\norm{e^1(m)} + \norm{e^3(m)}),\\
    \epsilon_{m, 2} &= 2W'(\norm{e^1(m)}^2 + \norm{e^2(m)}^2 + \norm{e^3(m)}^2).
    \end{aligned}
\end{equation*}

By Lemma \ref{lemma:actor-e2-bound}, we have with probability at least $1 - \frac{\delta}{2}$,
\begin{equation}\label{thm:actor-SAC:e1}
    \abs{\sum_{m=0}^{M - 1}\eta_m \epsilon_{m, 0}} \leq \frac{2W^2}{(1 - \gamma)^4}\sqrt{2\sum_{m=0}^{M - 1}\eta_m^2 \log \frac{4}{\delta}}.
\end{equation}

By Lemma \ref{lemma:actor-e1-bound} and Lemma \ref{lemma:actor-e3-bound}, we have with probability at least $1 - \frac{\delta}{2}$,
\begin{align}
    \sup_{m \leq M - 1}\epsilon_{m, 1} &\leq \frac{W}{(1 - \gamma)^2}\left(\sup_{m\leq M - 1}\norm{e^1(m)} + \sup_{m \leq M - 1}\norm{e^3(m)}\right)\nonumber\\
    &\leq \frac{(2 + \iota)W^2 \mu(\kappa)}{(1 - \gamma)^4}.\label{thm:actor-SAC:e2}
\end{align}

By Lemma \ref{lemma:actor-grad-bound}, we have almost surely $\max(\norm{e^1(m)}, \norm{e^2(m)}, \norm{e^3(m)}) \leq 2\frac{W}{(1 - \gamma)^2}$, and hence almost surely
\begin{align}
    \sup_{m \leq M - 1} \epsilon_{m, 2} &= 2W'\left(\norm{e^1(m)}^2 + \norm{e^2(m)}^2 + \norm{e^3(m)}^2\right)\nonumber\\
    &\leq \frac{24 W' W^2}{(1 - \gamma)^4}.\label{thm:actor-SAC:e3}
\end{align}

By union bound, \eqref{thm:actor-SAC:e1}, \eqref{thm:actor-SAC:e2}, and \eqref{thm:actor-SAC:e3} hold simultaneously with probability $1 - \delta$. Combining them with \eqref{thm:actor-SAC:e0} gives
\begin{align}
    &\frac{\sum_{m=0}^{M-1}\eta_m \norm{\nabla J(\theta(m))}^2}{2\sum_{m=0}^{M-1}\eta_m}\nonumber\\
    \leq{}& \frac{(J(\theta(M)) - J(\theta(0))) + \abs{\sum_{m=0}^{M - 1}\eta_m \epsilon_{m, 0}} + \sup_{m\leq M - 1}\epsilon_{m, 2}\sum_{m=0}^{M-1}\eta_m^2}{\sum_{m=0}^{M-1}\eta_m} + 2\sup_{m \leq M - 1}\epsilon_{m, 1}.\label{thm:actor-SAC:e4}
\end{align}
We can use identical steps with the proof of Theorem 5 in \cite{qu2019scalable} to bound the first term in \eqref{thm:actor-SAC:e4}, and use \eqref{thm:actor-SAC:e2} to bound the second term in \eqref{thm:actor-SAC:e4}. This completes the proof.

\section{Stochastic Approximation Scheme}
\subsection{Contraction of the Update Operator}\label{appendix:proposition:proj-contraction}
To show that the equation $\Pi F(\Phi x) = x$ has a unique solution $x^*$, by the Banach–Caccioppoli fixed-point theorem, it suffices to show that operator $\Pi F(\Phi \cdot)$ is a $\gamma$-contraction in $\norm{\cdot}_v$.
\begin{proposition}\label{proposition:proj-contraction}
If Assumption \ref{assump:contraction} holds, operator $\Pi F(\Phi \cdot)$ is a contraction in $\norm{\cdot}_v$, i.e., for any $x, y \in \mathbb{R}^\mathcal{M}$, $\norm{\Pi F(\Phi x) - \Pi F(\Phi y)}_v \leq \gamma \norm{x - y}_v.$
\end{proposition}
To prove this proposition, we first show both operator $\Pi$ and operator $\Phi$ are non-expansive in $\norm{\cdot}_v$ before combining them with $F$.
\begin{proof}[Proof of Proposition \ref{proposition:proj-contraction}]
We first show that operator $\Pi$ is non-expansive in $\norm{\cdot}_v$, i.e. for any $x, y \in \mathbb{R}^\mathcal{N}$, we have
\begin{equation}\label{proposition:proj-contraction:e0}
    \norm{\Pi x - \Pi y}_v \leq \norm{x - y}_v.
\end{equation}
Since $\Pi$ is a linear operator, it suffices to show that for any $x \in \mathbb{R}^\mathcal{N}$, $\norm{\Pi x}_v \leq \norm{x}_v$.

Recall that $\forall j \in \mathcal{M}, h^{-1}(j) := \{i\in \mathcal{N} \mid h(i) = j\}.$ Using this notation, the $j$ th element of vector $\Pi x$ is given by
\[\left(\Pi x\right)_j = \frac{1}{\sum_{i \in h^{-1}(j)}d_i}\left(\Phi^\top D x\right)_j = \frac{1}{\sum_{i \in h^{-1}(j)}d_i} \cdot \sum_{i \in h^{-1}(j)}d_i x_i.\]
Hence we see that
\begin{equation}\label{proposition:proj-contraction:e1}
    \frac{\abs{\left(\Pi x\right)_j}}{v_j} \leq \frac{1}{\sum_{i \in h^{-1}(j)}d_i} \cdot \sum_{i \in h^{-1}(j)}d_i \frac{\abs{x_i}}{v_j} \leq \sup_{i \in h^{-1}(j)}\frac{\abs{x_i}}{v_j}.
\end{equation}
By taking $\sup_j$ on both sides of \eqref{proposition:proj-contraction:e1}, we see that
\begin{equation}\label{proposition:proj-contraction:e2}
    \norm{\Pi x}_v = \sup_{j \in \mathcal{M}}  \frac{\abs{\left(\Pi x\right)_j}}{v_j} \leq \sup_{j \in \mathcal{M}} \sup_{i \in h^{-1}(j)}\frac{\abs{x_i}}{v_j} = \sup_{i \in \mathcal{N}}\frac{\abs{x_i}}{v_{h(i)}} = \norm{x}_v,
\end{equation}
where we use the definition of $\norm{\cdot}_v$ on $\mathbb{R}^\mathcal{N}$ in the last equation. Hence we have shown that $\Pi$ is non-expansive in $\norm{\cdot}_v$ (inequality \eqref{proposition:proj-contraction:e0}).

We can also show that for any $x, y \in \mathbb{R}^\mathcal{M}$, we have
\begin{equation}\label{proposition:proj-contraction:e3}
    \norm{\Phi x - \Phi y}_v = \norm{x - y}_v.
\end{equation}
Since $\Phi$ is a linear operator, we only need to show that for any $x \in \mathbb{R}^\mathcal{M}$, $\norm{\Phi x}_v = \norm{x}_v$.

Since $(\Phi x)_i = x_{h(i)}, \forall i \in \mathcal{N}$, by the definition of $\norm{\cdot}_v$ on $\mathbb{R}^\mathcal{N}$, we see that
$$\norm{\Phi x}_v = \sup_{i \in \mathcal{N}}\frac{\abs{(\Phi x)_i}}{v_{h(i)}} = \sup_{i \in \mathcal{N}}\frac{\abs{x_{h(i)}}}{v_{h(i)}} = \sup_{j \in \mathcal{M}}\frac{\abs{x_j}}{v_j} = \norm{x}_v.$$
Hence we have shown that $\Phi$ is non-expansive in $\norm{\cdot}_v$ (equation \eqref{proposition:proj-contraction:e3}).

Therefore, for any $x, y \in \mathbb{R}^\mathcal{M}$, we have
\begin{subequations}\label{proposition:proj-contraction:e4}
\begin{align}
    \norm{\Pi F(\Phi x) - \Pi F(\Phi y)}_v \leq{}& \norm{F(\Phi x) - F(\Phi y)}_v\label{proposition:proj-contraction:e4:s1}\\
    \leq{}& \gamma \norm{\Phi x - \Phi y}_v\label{proposition:proj-contraction:e4:s2}\\
    ={}& \gamma \norm{x - y}_v,\label{proposition:proj-contraction:e4:s3}
\end{align}
\end{subequations}
where we use \eqref{proposition:proj-contraction:e0} in \eqref{proposition:proj-contraction:e4:s1}; Assumption \ref{assump:contraction} in \eqref{proposition:proj-contraction:e4:s2}; \eqref{proposition:proj-contraction:e3} in \eqref{proposition:proj-contraction:e4:s3}.
\end{proof}

\subsection{Proof of Theorem \ref{thm:Stochastic-Approx-Main}}\label{appendix:thm:Stochastic-Approx-Main}

The proof approach of Theorem \ref{thm:Stochastic-Approx-Main} is similar to the proof of Theorem 4 in \cite{qu2020finite}. Specifically, we show an upper bound for $\norm{x(t) - x^*}_v$ by induction on time step $t$. To do so, we divide the whole proof into three steps: In Step 1, we manipulate the update rule \eqref{equ:Generalized-Asy-Stoc-Approx} so that it can be written in a recursive form of sequence $\norm{x(t) - x^*}_v$ (see Lemma \ref{lemma:error-decomposition}); In Step 2, we bound the effect of noise terms in the recursive form we obtained in Step 1; In Step 3, we combine the first two steps to finish the induction.

For simplicity of notation, we use $e_i$ to denote the indicator vector in $\mathbb{R}^n$, i.e. the $i$ th entry is $1$ and all other entries are $0$. We also use $\xi_i$ to denote the indicator vector in $\mathbb{R}^m$.

One of the main proof techniques used in \cite{qu2020finite} is to consider $D_t = \mathbb{E}e_{i_t} e_{i_t}^\top \mid \mathcal{F}_{t-\tau}$, which is the distribution of $i_t$ condition on $\mathcal{F}_{t-\tau}$, in the coefficients of the recursive relationship of sequence $\norm{x(t) - x^*}_v$. However, this approach does not work in the more general setting we consider because $x^*$ may not be the stationary point of operator $(\Phi^\top D_t \Phi)^{-1}\phi^\top D_t F(\Phi \cdot)$. As a result, we cannot decompose $\norm{x(t) - x^*}_v$ recursively if we use $D_t$ in the coefficients. To overcome this difficulty, we use $D = diag(d_1, \cdots, d_n)$, which is the stationary distribution of $i_t$, in the coefficients of the recursive relationship (Lemma \ref{lemma:error-decomposition}). %

Now we begin the technical part of our proof.

\textbf{Step 1: Decomposition of Error.} Let $D_t = \mathbb{E}e_{i_t} e_{i_t}^\top \mid \mathcal{F}_{t-\tau},$ where $\tau$ is a parameter that we will tune later. Then $D_t$ is a $\mathcal{F}_{t-\tau}$-measurable $n$-by-$n$ diagonal random matrix, with its $i$'th entry being $d_{t, i} = \mathbb{P}(i_t = i \mid \mathcal{F}_{t-\tau}).$ 
Recall that $D = diag(d_1, \cdots, d_n)$, where $d$ is the stationary distribution of the Markov Chain $\{i_t\}$.

Notice that for all $i \in \mathcal{N}$, we have $\xi_{h(i)} = \Phi^\top e_{i}.$ We can rewrite the update rule as
\begin{subequations}\label{error-decomposition:e1}
    \begin{align}
    x(t+1) ={}& x(t) + \alpha_t[e_{i_t}^\top F(\Phi x(t)) - \xi_{h(i_t)}^\top x(t) + w(t)]\xi_{h(i_t)}\nonumber\\
    ={}& x(t) + \alpha_t[\xi_{h(i_t)} e_{i_t}^\top F(\Phi x(t)) - \xi_{h(i_t)} \xi_{h(i_t)}^\top x(t) + w(t) \xi_{h(i_t)}]\nonumber\\
    ={}& x(t) + \alpha_t \Phi^\top \left[e_{i_t}e_{i_t}^\top \left(F(\Phi x(t)) - \Phi x(t)\right) + w(t)e_{i_t}\right]\label{error-decomposition:e1:s1}\\
    ={}& x(t) + \alpha_t \left[\Phi^\top D F(\Phi x(t)) - \Phi^\top D \Phi x(t)\right]\nonumber\\
    &+ \alpha_t \Phi^\top \left[(e_{i_t}e_{i_t}^\top - D)\left(F(\Phi x(t)) - \Phi x(t)\right) + w(t)e_{i_t}\right]\nonumber\\
    ={}& x(t) + \alpha_t \left[\Phi^\top D F(\Phi x(t)) - \Phi^\top D \Phi x(t)\right]\nonumber\\
    &+ \alpha_t \Phi^\top \left[(e_{i_t}e_{i_t}^\top - D)\left(F(\Phi x(t - \tau)) - \Phi x(t - \tau)\right) + w(t)e_{i_t}\right]\nonumber\\
    &+ \alpha_t \Phi^\top (e_{i_t}e_{i_t}^\top - D)\left[F(\Phi x(t)) - F(\Phi x(t - \tau)) - \Phi (x(t) - x(t - \tau))\right]\nonumber\\
    ={}& (I - \alpha_t \Phi^\top D \Phi)x(t) + \alpha_t \Phi^\top D F(\Phi x(t)) + \alpha_t(\epsilon(t) + \psi(t)), \label{error-decomposition:e1:s2}
    \end{align}
\end{subequations}
where in \eqref{error-decomposition:e1:s1}, we use $\xi_{h(i_t)} = \Phi^\top e_{i_t}$. Additionally, in \eqref{error-decomposition:e1:s2}, we define
\[\epsilon(t) = \Phi^\top \left[(e_{i_t}e_{i_t}^\top - D)\left(F(\Phi x(t - \tau)) - \Phi x(t - \tau)\right) + w(t)e_{i_t}\right]\]
and
\[\psi(t) = \Phi^\top (e_{i_t}e_{i_t}^\top - D)\left[F(\Phi x(t)) - F(\Phi x(t - \tau)) - \Phi (x(t) - x(t - \tau))\right].\]

We further decompose $\epsilon(t)$ as $\epsilon(t) = \epsilon_1(t) + \epsilon_2(t)$, where $\epsilon_1(t)$ and $\epsilon_2(t)$ are defined as
\[\epsilon_1(t) = \Phi^\top \left[(e_{i_t}e_{i_t}^\top - D_t)\left(F(\Phi x(t - \tau)) - \Phi x(t - \tau)\right) + w(t)e_{i_t}\right]\]
and
\[\epsilon_2(t) = \Phi^\top (D_t - D)\left(F(\Phi x(t - \tau)) - \Phi x(t - \tau)\right).\]
We see that condition on $\mathcal{F}_{t-\tau}$, the expected value of $\epsilon_1(t)$ is zero, i.e.
\begin{equation*}
\begin{aligned}
    &\mathbb{E}\epsilon_1(t)\mid \mathcal{F}_{t - \tau}\\
    ={}& \Phi^\top \mathbb{E}\left[(e_{i_t}e_{i_t}^\top - D_t) \mid \mathcal{F}_{t - \tau}\right][F(\Phi x(t - \tau)) - \Phi x(t - \tau)] + \Phi^\top \mathbb{E}\left[\mathbb{E}[w(t)\mid \mathcal{F}_t]e_{i_t}\mid \mathcal{F}_{t - \tau}\right]\\
    ={}& 0.
\end{aligned}
\end{equation*}
Recall that matrix $\Pi$ is defined as
\[\Pi = \left(\Phi^\top D \Phi\right)^{-1}\Phi^\top D.\]
By expanding \eqref{error-decomposition:e1} recursively, we obtain that
\begin{align}
    x(t + 1) ={}& \prod_{k=\tau}^t \left(I - \alpha_k \Phi^\top D \Phi\right)x(\tau) + \sum_{k = \tau}^t \alpha_k \left(\prod_{l = k+1}^t (I - \alpha_l \Phi^\top D \Phi)\right)\Phi^\top D F(\Phi x(k))\nonumber\\
    &+ \sum_{k=\tau}^t \alpha_k \left(\prod_{l = k+1}^t (I - \alpha_l \Phi^\top D \Phi)\right)(\epsilon(k) + \psi(k))\nonumber\\
    ={}& \tilde{B}_{\tau-1,t}x(\tau) + \sum_{k=\tau}^t B_{k, t}\Pi F(\Phi x(k)) + \sum_{k=\tau}^t\alpha_k \tilde{B}_{k,t}(\epsilon(k) + \psi(k)),\label{error-decomposition:e2}
\end{align}
where $B_{k, t} = \alpha_k \left(\Phi^\top D \Phi\right) \prod_{l = k+1}^t (I - \alpha_l \Phi^\top D \Phi)$ and $\tilde{B}_{k, t} = \prod_{l = k+1}^t \left(I - \alpha_l \Phi^\top D \Phi\right).$

For simplicity of notation, we define $D' = \Phi^\top D \Phi \in \mathbb{R}^{\mathcal{M} \times \mathcal{M}}.$ Notice that $D'$ is a diagonal matrix in $\mathbb{R}^{\mathcal{M}\times \mathcal{M}}$ with the $j$'th entry $d_{j}' = \sum_{j \in h^{-1}(i)}d_i.$ Clearly, $B_{k, t}$ and $\tilde{B}_{k, t}$ are $m$-by-$m$ diagonal matrices, with the $i$'th diagonal entry given by $b_{k, t, i}$ and $\tilde{b}_{k, t, i}$, where $b_{k, t, i} = \alpha_k d_{i}'\prod_{l=k+1}^t(1 - \alpha_l d_{i}')$ and $\tilde{b}_{k, t, i} = \prod_{l=k+1}^t(1 - \alpha_l d_{i}')$. Therefore, for any $i \in \mathcal{M}$, we have
\begin{equation}\label{error-decomposition:e3}
    \tilde{b}_{\tau-1, t, i} + \sum_{k=\tau}^t b_{k, t, i} = 1.
\end{equation}
Also, by the definition of $\sigma'$, we have that for any $i$, almost surely
\begin{equation*}
    b_{k, t, i} \leq \beta_{k, t} := \alpha_k \prod_{l = k+1}^t (1 - \alpha_l \sigma'), \tilde{b}_{k, t, i} \leq \tilde{\beta}_{k, t} = \prod_{l = k+1}^t (1 - \alpha_l \sigma'),
\end{equation*}
where $\sigma' = \min \{d_1', \cdots, d_m'\}.$

Recall that $x^*$ is the unique solution of the equation $\Pi F(\Phi x^*) = x^*$. Lemma \ref{lemma:error-decomposition} shows that we can expand the error term $\norm{x(t) - x^*}_v$ recursively.
\begin{lemma}\label{lemma:error-decomposition}
Let $\Upsilon_t = \norm{x(t) - x^*}_v$, we have almost surely,
\[\Upsilon_{t+1} \leq \tilde{\beta}_{\tau - 1, t}\Upsilon_\tau + \gamma \sup_{i \in \mathcal{M}}\sum_{k = \tau}^t b_{k, t, i}\Upsilon_k + \norm{\sum_{k=\tau}^t\alpha_k \tilde{B}_{k,t}\epsilon(k)}_v + \norm{\sum_{k=\tau}^t\alpha_k \tilde{B}_{k,t}\psi(k)}_v.\]
\end{lemma}
\begin{proof}[Proof of Lemma \ref{lemma:error-decomposition}]
By \eqref{error-decomposition:e2} and the triangle inequality of $\norm{\cdot}_v$, we have
\begin{align}
    &\norm{x(t+1) - x^*}_v\nonumber\\
    \leq{}& \sup_{i \in \mathcal{M}}\frac{1}{v_i}\abs{\tilde{b}_{\tau - 1, t, i} x_i(\tau) + \sum_{k = \tau}^t b_{k, t, i}\left(\Pi F(\Phi x(k))\right)_i - x_i^*}\nonumber\\
    &+ \norm{\sum_{k=\tau}^t \alpha_k \tilde{B}_{k, t}\epsilon(k)}_v + \norm{\sum_{k=\tau}^t \alpha_k \tilde{B}_{k, t}\psi(k)}_v.\label{lemma:error-decomposition:e0}
\end{align}
We also see that for each $i \in \mathcal{M}$,
\begin{subequations}\label{lemma:error-decomposition:e1}
    \begin{align}
    &\frac{1}{v_i}\abs{\tilde{b}_{\tau - 1, t, i} x_i(\tau) + \sum_{k = \tau}^t b_{k, t, i}\left(\Pi F(\Phi x(k))\right)_i - x_i^*}\nonumber\\
    \leq{}& \tilde{b}_{\tau - 1, t, i}\frac{1}{v_i}\abs{x_i(\tau) - x_i^*} + \sum_{k=\tau}^t b_{k, t, i}\frac{1}{v_i}\abs{\left(\Pi F(\Phi x(k))\right)_i - x_i^*}\label{lemma:error-decomposition:e1:s1}\\
    \leq{}& \tilde{b}_{\tau - 1, t, i}\norm{x(\tau) - x^*}_v + \sum_{k=\tau}^t b_{k, t, i}\norm{\left(\Pi F(\Phi x(k))\right) - x^*}_v\nonumber\\
    \leq{}& \tilde{b}_{\tau - 1, t, i}\norm{x(\tau) - x^*}_v + \gamma \sum_{k=\tau}^t b_{k, t, i}\norm{x(k) - x^*}_v,\label{lemma:error-decomposition:e1:s2}
    \end{align}
\end{subequations}
where in \eqref{lemma:error-decomposition:e1:s1}, we use \eqref{error-decomposition:e3} which says $\tilde{b}_{\tau -1, t, i} + \sum_{k=\tau}^t b_{k, t, i} = 1$ holds for all $i \in \mathcal{M}$; in \eqref{lemma:error-decomposition:e1:s2}, we use Proposition \ref{proposition:proj-contraction}, which says $\Pi F(\Phi \cdot )$ is $\gamma$-contraction in $\norm{\cdot}_v$ with fixed point $x^*.$

Therefore, by substituting \eqref{lemma:error-decomposition:e1} into \eqref{lemma:error-decomposition:e0}, we obtain that
\[\Upsilon_{t+1} \leq \tilde{\beta}_{\tau - 1, t}\Upsilon_\tau + \gamma \sup_{i \in \mathcal{M}}\sum_{k = \tau}^t b_{k, t, i}\Upsilon_k + \norm{\sum_{k=\tau}^t\alpha_k \tilde{B}_{k,t}\epsilon(k)}_v + \norm{\sum_{k=\tau}^t\alpha_k \tilde{B}_{k,t}\psi(k)}_v.\]
\end{proof}

\textbf{Step 2: Bounding $\norm{\sum_{k=\tau}^t\alpha_k \tilde{B}_{k,t}\epsilon(k)}_v$ and $\norm{\sum_{k=\tau}^t\alpha_k \tilde{B}_{k,t}\psi(k)}_v$.}

We start with a bound on each individual $\epsilon_1(k), \epsilon_2(k)$, and $\psi(k)$ in Lemma \ref{lemma:Bound-Epsilon-Phi}. For simplicity of notation, we define $\underline{v} := \inf_{j \in \mathcal{M}} v_j$.
\begin{lemma}\label{lemma:Bound-Epsilon-Phi}
The following bounds hold almost surely.
\begin{enumerate}
    \item $\norm{\epsilon_1(t)}_v \leq 4\bar{x} + 2C + \frac{\bar{w}}{\underline{v}} := \bar{\epsilon}.$
    \item $\norm{\epsilon_2(t)}_v \leq (2\bar{x} + C)\cdot 2 K_1 \exp(-\tau/K_2).$
    \item $\norm{\psi(t)}_v \leq 3\left(2\bar{x} + C + \frac{\bar{w}}{\underline{v}}\right)\sum_{k = t - \tau + 1}^t \alpha_{k-1}.$
\end{enumerate}
\end{lemma}

\begin{proof}[Proof of Lemma \ref{lemma:Bound-Epsilon-Phi}]
By the definition of $\norm{\cdot}_v$ in $\mathbb{R}^\mathcal{M}$ and its extension to $\mathbb{R}^\mathcal{N}$, the induced matrix norm of $\norm{\cdot}$ for a matrix $A = [a_{ij}]_{i \in \mathcal{M}, j \in \mathcal{N}}$ is given by $\norm{A}_v = \sup_{i\in \mathcal{M}}\sum_{j\in \mathcal{N}} \frac{v_{h(j)}}{v_i}\abs{a_{ij}}.$ Recall that the $i$'th entry of the diagonal matrix $D_t$ is given by $d_{t,i} = \mathbb{P}\left(i_t = i \mid \mathcal{F}_{t - \tau}\right)$. Hence we have that
\begin{equation}\label{lemma:Bound-Epsilon-Phi:e0-1}
    \norm{\Phi^\top (e_{i_t}e_{i_t}^\top - D_t)}_v = \sup_{j \in \mathcal{M}}\sum_{i \in \mathcal{N}} 1(h(i) = j) \cdot \abs{1(i = i_t) - d_{t, i}} \leq 2.
\end{equation}

Therefore, we can upper bound $\norm{\epsilon_1(t)}_v$ by
\begin{subequations}\label{lemma:Bound-Epsilon-Phi:e0-2}
    \begin{align}
    \norm{\epsilon_1(t)}_v ={}& \norm{\Phi^\top \left[(e_{i_t}e_{i_t}^\top - D_t)\left(F(\Phi x(t - \tau)) - \Phi x(t - \tau)\right) + w(t)e_{i_t}\right]}_v\nonumber\\
    \leq{}& \norm{\Phi^\top (e_{i_t}e_{i_t}^\top - D_t)}_v \norm{F(\Phi x(t - \tau)) - \Phi x(t - \tau)}_v + \abs{w(t)}\norm{\Phi^\top e_{i_t}}_v\nonumber\\
    \leq{}& 2\norm{F(\Phi x(t - \tau)) - \Phi x(t - \tau)}_v + \abs{w(t)}\norm{\Phi^\top e_{i_t}}_v\label{lemma:Bound-Epsilon-Phi:e0-2:s1}\\
    \leq{}& 2\norm{F(\Phi x(t - \tau))}_v + 2\norm{x(t - \tau)}_v + \frac{\bar{w}}{\underline{v}}\label{lemma:Bound-Epsilon-Phi:e0-2:s2}\\
    \leq{}& 4\bar{x} + 2C + \frac{\bar{w}}{\underline{v}},\label{lemma:Bound-Epsilon-Phi:e0-2:s3}
    \end{align}
\end{subequations}
where we use \eqref{lemma:Bound-Epsilon-Phi:e0-1} in \eqref{lemma:Bound-Epsilon-Phi:e0-2:s1}; the triangle inequality, the definition of $\bar{v}$, and Assumption \ref{assump:martingale} in \eqref{lemma:Bound-Epsilon-Phi:e0-2:s2}; Assumption \ref{assump:contraction} in \eqref{lemma:Bound-Epsilon-Phi:e0-2:s3}.

For $\norm{\epsilon_2(t)}_v$, recall that
\begin{align}
    \norm{\epsilon_2(t)}_v ={}& \norm{\Phi^\top  (D_t - D)\left(F(\Phi x(t - \tau)) - \Phi x(t - \tau)\right)}_v\nonumber\\
    ={}& \sup_{j \in \mathcal{M}}\frac{1}{v_j}\abs{\sum_{i\in \mathcal{N}}1(h(i) = j)(d_{t, i} - d_i)\left(F(\Phi x(t - \tau)) - \Phi x(t - \tau)\right)_i}\nonumber\\
    ={}& \sup_{j \in \mathcal{M}}\frac{1}{v_j}\abs{\sum_{i\in h^{-1}(j)}(d_{t, i} - d_i)\left(F(\Phi x(t - \tau)) - \Phi x(t - \tau)\right)_i}.\label{lemma:Bound-Epsilon-Phi:e1}
\end{align}
By Assumption \ref{assump:geometric-mixing}, we have that
\begin{equation}\label{lemma:Bound-Epsilon-Phi:e2-0}
    \sup_{\mathcal{S} \subseteq \mathcal{N}}\abs{\sum_{i \in \mathcal{S}}d_i - \sum_{i \in \mathcal{S}}d_{t, i}} \leq K_1 \exp(-\tau/K_2).
\end{equation}

Our objective is to bound the following term in \eqref{lemma:Bound-Epsilon-Phi:e1} for all $j \in \mathcal{M}$:
\[\abs{\sum_{i\in h^{-1}(j)}(d_{t, i} - d_i)\left(F(\Phi x(t - \tau)) - \Phi x(t - \tau)\right)_i}.\]

Let $M_j := \sup_{i \in h^{-1}(j)} \abs{\left(F(\Phi x(t - \tau)) - \Phi x(t - \tau)\right)_i}$. Define function $g: [-M_j, M_j]^\mathcal{N} \to \mathbb{R}$ as
\[g(y) = \abs{\sum_{i \in h^{-1}(j)}(d_{t,i} - d_i)y_i}.\]
Suppose $y_{max} \in \argmax_y g(y).$ We know that for $i \in h^{-1}(j)$, $(y_{max})_i$ is either $M_j$ or $-M_j$ if $d_{t,i} - d_i \not = 0$. Let $S_j := \{i \in h^{-1}(j) \mid (y_{max})_i = M_j\}$ and $S_j' := \{i \in h^{-1}(j) \mid (y_{max})_i = - M_j\}.$

Therefore, we see that
\begin{subequations}\label{lemma:Bound-Epsilon-Phi:e2}
    \begin{align}
    &\abs{\sum_{i\in h^{-1}(j)}(d_{t, i} - d_i)\left(F(\Phi x(t - \tau)) - \Phi x(t - \tau)\right)_i}\nonumber\\
    \leq{}& \max_{y \in [-M_j, M_j]^\mathcal{N}}g(y)\label{lemma:Bound-Epsilon-Phi:e2:s1}\\
    ={}& \abs{\sum_{i \in S_j}(d_{t,i} - d_i)}M_j + \abs{\sum_{i \in S_j'}(d_{t, i} - d_i)}M_j\nonumber\\
    \leq{}& 2K_1 \exp(-\tau/K_2) M_j.\label{lemma:Bound-Epsilon-Phi:e2:s2}
    \end{align}
\end{subequations}
where we use the definition of function $g$ in \eqref{lemma:Bound-Epsilon-Phi:e2:s1}; we use \eqref{lemma:Bound-Epsilon-Phi:e2-0} in \eqref{lemma:Bound-Epsilon-Phi:e2:s2}.

Substituting \eqref{lemma:Bound-Epsilon-Phi:e2} into \eqref{lemma:Bound-Epsilon-Phi:e1} gives that
\begin{subequations}\label{lemma:Bound-Epsilon-Phi:e2-1}
\begin{align}
    \norm{\epsilon_2(t)}_v \leq{}& \norm{F(\Phi x(t - \tau)) - \Phi x(t - \tau)}_v\cdot 2 K_1 \exp(-\tau/K_2)\nonumber\\
    \leq{}& \left(\norm{F(\Phi x(t - \tau))}_v + \norm{\Phi x(t - \tau)}_v\right)\cdot 2 K_1 \exp(-\tau/K_2)\label{lemma:Bound-Epsilon-Phi:e2-1:s1}\\
    \leq{}& (2\bar{x} + C)\cdot 2 K_1 \exp(-\tau/K_2),\label{lemma:Bound-Epsilon-Phi:e2-1:s2}
\end{align}
\end{subequations}
where we use the triangle inequality in \eqref{lemma:Bound-Epsilon-Phi:e2-1:s1}; we use Assumption \ref{assump:contraction} in \eqref{lemma:Bound-Epsilon-Phi:e2-1:s2}.

As for $\norm{\psi(t)}_v$, we have the following bound
\begin{align}
    &\norm{\psi(t)}_v\nonumber\\
    ={}& \norm{\Phi^\top(e_{i_t}e_{i_t}^\top - D)\left(F(\Phi x(t)) - F(\Phi x(t - \tau))\right) - \Phi^\top (e_{i_t}e_{i_t}^\top - D) \Phi \left(x(t) - x(t - \tau)\right)}_v\nonumber\\
    \leq{}& \norm{\Phi^\top(e_{i_t}e_{i_t}^\top - D)\left(F(\Phi x(t)) - F(\Phi x(t - \tau))\right)}_v + \norm{\Phi^\top (e_{i_t}e_{i_t}^\top - D) \Phi \left(x(t) - x(t - \tau)\right)}_v\nonumber\\
    \leq{}& \norm{\Phi^\top(e_{i_t}e_{i_t}^\top - D)}_v \cdot \norm{\left(F(\Phi x(t)) - F(\Phi x(t - \tau))\right)}_v\nonumber\\
    &+ \norm{\Phi^\top (e_{i_t}e_{i_t}^\top - D) \Phi}_v \cdot \norm{ \left(x(t) - x(t - \tau)\right)}_v. \label{lemma:Bound-Epsilon-Phi:e3}
\end{align}
Notice that
\[\norm{\Phi^\top (e_{i_t}e_{i_t}^\top - D) \Phi}_v = \norm{\xi_{h(i_t)}\xi_{h(i_t)}^\top - D'}_v = \sup_{j\in \mathcal{M}}\abs{1(h(i_t) = j) - d_j'} \leq 1.\]
Substituting this into \eqref{lemma:Bound-Epsilon-Phi:e3} and use \eqref{lemma:Bound-Epsilon-Phi:e0-1}, we obtain that
\begin{align}
    \norm{\psi(t)}_v\nonumber \leq{}& 2\norm{F(\Phi x(t)) - F(\Phi x(t - \tau))}_v + \norm{x(t) - x(t - \tau)}_v\nonumber\\
    \leq{}& 3\norm{x(t) - x(t - \tau)}_v\nonumber\\
    \leq{}& 3\sum_{k = t - \tau + 1}^t \norm{x(k) - x(k - 1)}_v. \label{lemma:Bound-Epsilon-Phi:e3-1}
\end{align}

By the update rule of $x$ and Assumption \ref{assump:contraction}, we have that
\begin{align}
    \norm{x(t) - x(t-1)}_v \leq{}& \alpha_{t-1}\left(\norm{F(\Phi x(t-1))}_v + \norm{x(t-1)}_v + \frac{\bar{w}}{\underline{v}}\right)\nonumber\\
    \leq{}& \alpha_{t-1}\left(2\bar{x} + C + \frac{\bar{w}}{\underline{v}}\right).\label{lemma:Bound-Epsilon-Phi:e3-2}
\end{align}
Substituting \eqref{lemma:Bound-Epsilon-Phi:e3-2} into \eqref{lemma:Bound-Epsilon-Phi:e3-1}, we obtain that
\[\norm{\psi(t)}_v \leq 3\left(2\bar{x} + C + \frac{\bar{w}}{\underline{v}}\right)\sum_{k = t - \tau + 1}^t \alpha_{k-1}.\]
\end{proof}

\begin{lemma}\label{lemma:Step2-Auxiliary1}
If $\alpha_t = \frac{H}{t + t_0}$, where $H > \frac{2}{\sigma'}$ and $t_0 \geq \max(4H, \tau)$, then $\beta_{k, t}, \tilde{\beta}_{k, t}$ satisfies the following
\begin{enumerate}
    \item $\beta_{k, t} \leq \frac{H}{k + t_0}\left(\frac{k + 1 + t_0}{t + 1 + t_0}\right)^{\sigma' H}, \tilde{\beta}_{k, t} \leq \left(\frac{k + 1 + t_0}{t + 1 + t_0}\right)^{\sigma' H}.$
    \item $\sum_{k=1}^t \beta_{k, t}^2 \leq \frac{2H}{\sigma'}\frac{1}{t + 1 + t_0}.$
    \item $\sum_{k=\tau}^t \beta_{k, t}\sum_{l = k - \tau + 1}^k \alpha_{l-1} \leq \frac{8H\tau}{\sigma'}\frac{1}{t + 1 + t_0}.$
\end{enumerate}
\end{lemma}
\begin{proof}[Proof of Lemma \ref{lemma:Step2-Auxiliary1}]
To show Lemma \ref{lemma:Step2-Auxiliary1}, we only need to substitute $\sigma'$ for $\sigma$ in the proof of \cite{qu2020finite}[Lemma 10].
\end{proof}

\begin{lemma}\label{lemma:psi-bound}
The following inequality holds almost surely
\[\norm{\sum_{k=\tau}^t\alpha_k \tilde{B}_{k,t}\psi(k)}_v \leq \frac{24\left(2\bar{x} + C + \frac{\bar{w}}{\underline{v}}\right)H\tau}{\sigma'}\frac{1}{t + 1 + t_0} := C_\psi \frac{1}{t + 1 + t_0}.\]
\end{lemma}
\begin{proof}[Proof of Lemma \ref{lemma:psi-bound}]
We have that
\begin{subequations}\label{lemma:psi-bound:e1}
    \begin{align}
    \norm{\sum_{k=\tau}^t\alpha_k \tilde{B}_{k,t}\psi(k)}_v \leq{}& \sum_{k=\tau}^t \alpha_k \norm{\tilde{B}_{k, t}}_v \norm{\psi(k)}_v\nonumber\\
    \leq{}& 3\left(2\bar{x} + C + \frac{\bar{w}}{\underline{v}}\right)\sum_{k=\tau}^t \beta_{k, t}\sum_{l=k-\tau+1}^k \alpha_{l-1}\label{lemma:psi-bound:e1:s1}\\
    \leq{}& \frac{24\left(2\bar{x} + C + \frac{\bar{w}}{\underline{v}}\right)H\tau}{\sigma'}\frac{1}{t + 1 + t_0}, \label{lemma:psi-bound:e1:s2}
    \end{align}
\end{subequations}
where we use Lemma \ref{lemma:Bound-Epsilon-Phi} in \eqref{lemma:psi-bound:e1:s1}; Lemma \ref{lemma:Step2-Auxiliary1} in \eqref{lemma:psi-bound:e1:s2}.
\end{proof}

\begin{lemma}\label{lemma:epsilon1-bound}
For each $t$, with probability at least $1 - \delta$, we have
\[\norm{\sum_{k=\tau}^t \alpha_k \tilde{B}_{k, t} \epsilon_1(k)}_v \leq \frac{H\bar{\epsilon}}{t + t_0}\sqrt{2 \tau t \log\left(\frac{2\tau m}{\delta}\right)}.\]
\end{lemma}

To show Lemma \ref{lemma:epsilon1-bound}, we need to use Lemma \ref{lemma:shifted-Martingale}, which is Lemma 13 in \cite{qu2020finite}.

\begin{lemma}\label{lemma:shifted-Martingale}
Let $X_t$ be a $\mathcal{F}_t$-adapted stochastic process which satisfies $\mathbb{E}X_t \mid \mathcal{F}_{t - \tau} = 0.$ Further, $\abs{X_t} \leq \bar{X}_t$ almost surely. Then with probability $1 - \delta$, we have, $\abs{\sum_{k=0}^t X_t} \leq \sqrt{2\tau \sum_{k=0}^t \bar{X}_k^2 \log \left(\frac{2\tau}{\delta}\right)}.$
\end{lemma}

\begin{proof}[Proof of Lemma \ref{lemma:epsilon1-bound}]
Recall that $\sum_{k=\tau}\alpha_k \tilde{B}_{k, t}\epsilon_1(k)$ is a random vector in $\mathbb{R}^\mathcal{M}$, with its $i$'th entry
\[\sum_{k=\tau}^t \alpha_k (\epsilon_1)_i(k)\prod_{l = k+1}^t (1 - \alpha_l d_i').\]
Since step sizes $\{\alpha_l\}$ are deterministic, we see that
\[\mathbb{E}\left[\alpha_k (\epsilon_1)_i(k)\prod_{l = k+1}^t (1 - \alpha_l d_i')\mid \mathcal{F}_{k-\tau}\right] = \alpha_k \prod_{l = k+1}^t (1 - \alpha_l d_i') \mathbb{E}\left[(\epsilon_1)_i(k)\mid \mathcal{F}_{k-\tau}\right] = 0.\]
Notice that
\begin{subequations}\label{lemma:epsilon1-bound:e1}
    \begin{align}
    \alpha_k \prod_{l = k+1}^t (1 - \alpha_l d_i') ={}& \frac{H}{k + t_0}\prod_{l = k+1}^t \left(1 - \frac{H d_i'}{l + t_0}\right)\label{lemma:epsilon1-bound:e1:s1}\\
    \leq{}& \frac{H}{k + t_0}\prod_{l = k+1}^t \left(1 - \frac{2}{l + t_0}\right)\label{lemma:epsilon1-bound:e1:s2}\\
    \leq{}& \frac{H}{k + t_0}\prod_{l = k+1}^t \left(1 - \frac{1}{l + t_0}\right)\nonumber\\
    \leq{}& \frac{H}{t + t_0},\nonumber
    \end{align}
\end{subequations}
where we use $\alpha_l = \frac{H}{l + t_0}$ in \eqref{lemma:epsilon1-bound:e1:s1}; we use $H > \frac{2}{\sigma'}$ in \eqref{lemma:epsilon1-bound:e1:s2}.

By the definition of $\bar{\epsilon}$, we also see that
$\abs{(\epsilon_1)_i(k)} \leq v_i \bar{\epsilon}.$
Therefore, by Lemma \ref{lemma:shifted-Martingale}, we obtain that
\begin{equation*}
    \abs{\sum_{k=\tau}^t \alpha_k (\epsilon_1)_i(k)\prod_{l = k+1}^t (1 - \alpha_l d_i')} \leq \frac{H v_i \bar{\epsilon}}{t + t_0}\sqrt{2 \tau t \log\left(\frac{2\tau}{\delta}\right)}
\end{equation*}
holds with probability at least $1 - \delta$. By union bound, we see that with probability at least $1 - \delta$,
\[\norm{\sum_{k=\tau}^t \alpha_k \tilde{B}_{k, t} \epsilon_1(k)}_v \leq \frac{H\bar{\epsilon}}{t + t_0}\sqrt{2 \tau t \log\left(\frac{2\tau m}{\delta}\right)}.\]
\end{proof}

\begin{lemma}\label{lemma:epsilon2-bound}
If we set $\tau$ to be an integer such that 
\[\tau \geq 2K_2 \max \left(\log t, 1\right),\]
we have that
\[\norm{\sum_{k=\tau}^t \alpha_k \tilde{B}_{k, t} \epsilon_2(k)}_v \leq \frac{C_{\epsilon_2}}{t + t_0 + 1},\]
where $t_0 = \max (\tau, 4H)$ and $C_{\epsilon_2} = (2\bar{x} + C) \cdot 2K_1 (1 + 2K_2 + 4H).$ 
\end{lemma}
\begin{proof}[Proof of Lemma \ref{lemma:epsilon2-bound}]
Since $K_2 \geq 1$, the bound is trivial when $t = 1$. We consider the case when $t \geq 2$ below.

Since $\alpha_k \tilde{B}_{k, t}$ is a diagonal matrix and its entries are positive and less than $1$, we have that
\begin{subequations}\label{lemma:epsilon2-bound:e0}
    \begin{align}
    \norm{\sum_{k=\tau}^t \alpha_k \tilde{B}_{k, t} \epsilon_2(k)}_v &\leq \sum_{k=\tau}^t \norm{\alpha_k \tilde{B}_{k, t}}_v \cdot \norm{\epsilon_2(k)}_v\nonumber\\
    &\leq t \norm{\epsilon_2(k)}_v\label{lemma:epsilon2-bound:e0:s1}\\
    &\leq t (2\bar{x} + C)\cdot 2 K_1 \exp(-\tau/K_2).\label{lemma:epsilon2-bound:e0:s2}
    \end{align}
\end{subequations}
where we use $\norm{\alpha_k \tilde{B}_{k, t}}_v \leq 1$ in \eqref{lemma:epsilon2-bound:e0:s1}; Lemma \ref{lemma:Bound-Epsilon-Phi} in \eqref{lemma:epsilon2-bound:e0:s2}.

To show Lemma \ref{lemma:epsilon2-bound}, we only need to show
\begin{equation}\label{lemma:epsilon2-bound:e1}
    t(2\bar{x} + C) \cdot 2K_1 (t + \tau + 4H)\exp(-\tau/K_2) \leq C_{\epsilon_2}
\end{equation}
holds for all $\tau \geq 2K_2 \log t$ because $t + t_0 + 1 \leq t + \tau + 4H.$

To study how the left hand side of \eqref{lemma:epsilon2-bound:e1} changes with $\tau$, we define function
\[g(\tau) = (\tau + t + 4H)\exp(-\tau/K_2).\]
Notice that we view $\tau$ as real number in function $g$, so we can get the derivative of $g$:
\[g'(\tau) = \frac{\exp(-\tau/K_2)}{K_2}(K_2 - t - 4H - \tau).\]
Therefore, when $\tau \geq 2K_2 \log t$, we always have $g'(\tau) < 0$. Hence we obtain that
\begin{equation}\label{lemma:epsilon2-bound:e2}
    g(\tau) \leq g(2K_2 \log t) = \frac{2K_2 \log t + t + 4H}{t^2} \leq \frac{1 + 2K_2 + 4H}{t}
\end{equation}
holds for all $\tau \geq 2K_2 \log t.$

Substituting \eqref{lemma:epsilon2-bound:e2} into \eqref{lemma:epsilon2-bound:e1} finishes the proof.
\end{proof}

\textbf{Step 3: Bounding the error sequence.}
Based on the recursive relationship we derived in Lemma \ref{lemma:error-decomposition} and the bounds we obtained in Step 2, we want to show that, with probability $1 - \delta$,
\begin{equation}\label{Step3:main-obj}
    \Upsilon_t \leq \frac{C_a}{\sqrt{t + t_0}} + \frac{C_a'}{t + t_0},
\end{equation}
holds for all $\tau \leq t \leq T$, where
\[C_a = \frac{2H\bar{\epsilon}}{1 - \gamma}\sqrt{2 \tau \log\left(\frac{2\tau m T}{\delta}\right)}, C_a' = \frac{4}{1 - \gamma}\max\left(C_\psi + C_{\epsilon_2}, 2\bar{x}(\tau + t_0)\right).\]
Notice that $C_a$ and $C_a'$ are independent of $t$ but may dependent on $T$. We set $\tau = 2K_2 \log T.$

By applying union bound to Lemma \ref{lemma:epsilon1-bound}, we see that with probability at least $1 - \delta$, for any $t \leq T$,
\[\norm{\sum_{k=\tau}^t \alpha_k \tilde{B}_{k, t} \epsilon_1(k)}_v \leq \frac{C_{\epsilon_1}}{\sqrt{t + 1 + t_0}},\]
where $C_{\epsilon_1} = H\bar{\epsilon}\sqrt{2 \tau \log\left(\frac{2\tau m T}{\delta}\right)}$.

Therefore, we get with probability $1 - \delta$, \eqref{Step3:Induction} holds for all $\tau \leq t \leq T$:
\begin{equation}\label{Step3:Induction}
    \Upsilon_{t+1} \leq \tilde{\beta}_{\tau - 1, t}\Upsilon_\tau + \gamma \sup_{i \in \mathcal{M}}\sum_{k = \tau}^t b_{k, t, i}\Upsilon_k + \frac{C_{\epsilon_1}}{\sqrt{t + 1 + t_0}} + \frac{C_\psi + C_{\epsilon_2}}{t + 1 + t_0}.
\end{equation}
We now condition on \eqref{Step3:Induction} to show \eqref{Step3:main-obj} by induction. \eqref{Step3:main-obj} is true for $t = \tau$, as $\frac{C_a'}{\tau + t_0} \geq \frac{8}{1 - \gamma}\bar{x} \geq \Upsilon_\tau$, where we have used $\Upsilon_\tau = \norm{x(\tau) - x^*}_v \leq \norm{x(\tau)}_v + \norm{x^*}_v \leq 2\bar{x}.$ Then, assuming \eqref{Step3:main-obj} is true for up to $k \leq t$. By \eqref{Step3:Induction}, we have that
\begin{align}
    \Upsilon_{t+1} \leq{}& \tilde{\beta}_{\tau - 1, t}\Upsilon_\tau + \gamma \sup_{i \in \mathcal{M}}\sum_{k = \tau}^t b_{k, t, i}\left[\frac{C_a}{\sqrt{k + t_0}} + \frac{C_a'}{k + t_0}\right] + \frac{C_{\epsilon_1}}{\sqrt{t + 1 + t_0}} + \frac{C_\psi + C_{\epsilon_2}}{t + 1 + t_0}\nonumber\\
    \leq{}& \tilde{\beta}_{\tau - 1, t}\Upsilon_\tau + \gamma C_a \sup_{i \in \mathcal{M}}\sum_{k = \tau}^t b_{k, t, i}\frac{1}{\sqrt{k + t_0}} + \gamma C_a' \sup_{i \in \mathcal{M}}\sum_{k = \tau}^t\frac{1}{k + t_0}b_{k, t, i}\nonumber\\
    &+ \frac{C_{\epsilon_1}}{\sqrt{t + 1 + t_0}} + \frac{C_\psi + C_{\epsilon_2}}{t + 1 + t_0}.\label{Step3:Induction-e2}
\end{align}

We use the following auxiliary lemma to handle the second and the third term in \eqref{Step3:Induction-e2}.
\begin{lemma}\label{lemma:Step3-Auxiliary1}
If $\sigma' H (1 - \sqrt{\gamma}) \geq 1, t_0 \geq 1$, and $\alpha_0 \leq \frac{1}{2}$, then, for any $i \in \mathcal{N}$, and any $0 < \omega \leq 1$, we have
\[\sum_{k=\tau}^t b_{k, t, i}\frac{1}{(k + t_0)^\omega} \leq \frac{1}{\sqrt{\gamma}(t + 1 + t_0)^\omega}.\]
\end{lemma}
\begin{proof}[Proof of Lemma \ref{lemma:Step3-Auxiliary1}]
Recall that $\alpha_k = \frac{H}{k + t_0}$, and $b_{k, t, i} = \alpha_k d_i' \prod_{l = k+1}^t (1 - \alpha_l d_i')$, where $d_i' \geq \sigma'$. 

Define $e_t = \sum_{k=\tau}^t b_{k, t, i}\frac{1}{(k + t_0)^\omega}.$ We use induction on $t$ to show that $e_t \leq \frac{1}{\sqrt{\gamma}(t + 1 + t_0)^\omega}.$

The statement is clearly true for $t = \tau$. Assume it is true for $t-1$. Notice that
\begin{subequations}\label{lemma:Step3-Auxiliary1:e0}
    \begin{align}
    e_t ={}& \sum_{k=\tau}^{t-1}b_{k, t, i}\frac{1}{(k + t_0)^\omega} + b_{t, t, i}\frac{1}{(t + t_0)^\omega}\nonumber\\
    ={}& (1 - \alpha_t d_i')\sum_{k=\tau}^{t-1}b_{k, t-1, i}\frac{1}{(k + t_0)^\omega} + \alpha_t d_i'\frac{1}{(t + t_0)^\omega}\label{lemma:Step3-Auxiliary1:e0:s1}\\
    ={}& (1 - \alpha_t d_i')e_{t-1} + \alpha_t d_i'\frac{1}{(t + t_0)^\omega}\nonumber\\
    \leq{}& (1 - \alpha_t d_i')\frac{1}{\sqrt{\gamma}(t + t_0)^\omega} + \alpha_t d_i'\frac{1}{(t + t_0)^\omega}\label{lemma:Step3-Auxiliary1:e0:s2}\\
    ={}& \left[1 - \alpha_t d_i' (1 - \sqrt{\gamma})\right]\frac{1}{\sqrt{\gamma}(t + t_0)^\omega},\nonumber
    \end{align}
\end{subequations}
where we use $b_{t,t,i} = \alpha_t d_i'$ in \eqref{lemma:Step3-Auxiliary1:e0:s1}; we use the induction assumption in \eqref{lemma:Step3-Auxiliary1:e0:s2}.

Plugging in $\alpha_t = \frac{H}{t + t_0}$, we see that
\begin{subequations}\label{lemma:Step3-Auxiliary1:e1}
    \begin{align}
    e_t \leq{}& \left[1 - \frac{\sigma' H}{t + t_0} (1 - \sqrt{\gamma})\right]\frac{1}{\sqrt{\gamma}(t + t_0)^\omega}\label{lemma:Step3-Auxiliary1:e1:s1}\\
    ={}& \left[1 - \frac{\sigma' H}{t + t_0} (1 - \sqrt{\gamma})\right]\left(1 + \frac{1}{t + t_0}\right)^\omega \frac{1}{\sqrt{\gamma}(t + 1 + t_0)^\omega}\nonumber\\
    \leq{}& \left(1 - \frac{1}{t + t_0}\right)\left(1 + \frac{1}{t + t_0}\right)^\omega \frac{1}{\sqrt{\gamma}(t + 1 + t_0)^\omega}\label{lemma:Step3-Auxiliary1:e1:s2}\\
    \leq{}& \left(1 - \frac{1}{t + t_0}\right)\left(1 + \frac{1}{t + t_0}\right) \frac{1}{\sqrt{\gamma}(t + 1 + t_0)^\omega}\label{lemma:Step3-Auxiliary1:e1:s3}\\
    \leq{}& \frac{1}{\sqrt{\gamma}(t + 1 + t_0)^\omega},\nonumber
    \end{align}
\end{subequations}
where we use $d_i' \geq \sigma'$ in \eqref{lemma:Step3-Auxiliary1:e1:s1}; we use the assumption that $\sigma' H (1 - \sqrt{\gamma}) \geq 1$ in \eqref{lemma:Step3-Auxiliary1:e1:s2}; we use $0 < \omega \leq 1$ in \eqref{lemma:Step3-Auxiliary1:e1:s3}.
\end{proof}
Applying Lemma \ref{lemma:Step3-Auxiliary1} to \eqref{Step3:Induction-e2}, we see that
\begin{subequations}\label{Step3:Induction-e3}
    \begin{align}
    \Upsilon_{t+1} \leq{}& \tilde{\beta}_{\tau - 1, t}\Upsilon_\tau + \sqrt{\gamma}C_a\frac{1}{\sqrt{t + 1 + t_0}} + \sqrt{\gamma}C_a'\frac{1}{t + 1 + t_0}\nonumber\\
    &+ C_{\epsilon_1}\frac{1}{\sqrt{t + 1 + t_0}} + (C_\psi + C_{\epsilon_2})\frac{1}{t + 1 + t_0}\label{Step3:Induction-e3:s1}\\
    \leq{}& \left(\sqrt{\gamma}C_a\frac{1}{\sqrt{t + 1 + t_0}} + C_{\epsilon_1}\frac{1}{\sqrt{t + 1 + t_0}}\right)\nonumber\\
    &+ \left(\sqrt{\gamma}C_a'\frac{1}{t + 1 + t_0} + (C_\psi + C_{\epsilon_2})\frac{1}{t + 1 + t_0}+ \left(\frac{\tau + t_0}{t + 1 + t_0}\right)^{\sigma' H}\Upsilon_\tau\right),\label{Step3:Induction-e3:s2}
    \end{align}
\end{subequations}
where we use Lemma \ref{lemma:Step3-Auxiliary1} in \eqref{Step3:Induction-e3:s1}; we use the bound on $\tilde{\beta}_{\tau-1, t}$ in Lemma \ref{lemma:Step2-Auxiliary1} in \eqref{Step3:Induction-e3:s2}.

To bound the two terms in \eqref{Step3:Induction-e3:s2}, we define
\[\chi_t := \sqrt{\gamma}C_a\frac{1}{\sqrt{t + 1 + t_0}} + C_{\epsilon_1}\frac{1}{\sqrt{t + 1 + t_0}}\]
and
\[\chi_t' = \sqrt{\gamma}C_a'\frac{1}{t + 1 + t_0} + (C_\psi + C_{\epsilon_2})\frac{1}{t + 1 + t_0} + \left(\frac{\tau + t_0}{t + 1 + t_0}\right)^{\sigma' H}a_\tau.\]
To finish the induction, it suffices to show that $\chi_t \leq \frac{C_a}{\sqrt{t + 1 + t_0}}$ and $\chi_t' \leq \frac{C_a'}{t + 1 + t_0}$. To see this
\[\chi_t \frac{\sqrt{t + 1 + t_0}}{C_a} = \sqrt{\gamma} + \frac{C_{\epsilon_1}}{C_a}, \chi_t' \frac{t + 1 + t_0}{C_a'} = \sqrt{\gamma} + \frac{C_\psi + C_{\epsilon_2}}{C_a'} + \frac{\Upsilon_\tau (\tau + t_0)}{C_a'}\left(\frac{\tau + t_0}{t + 1 + t_0}\right)^{\sigma' H - 1}.\]
It suffices to show that $\frac{C_{\epsilon_1}}{C_a} \leq 1 - \sqrt{\gamma}$, $\frac{C_\psi + C_{\epsilon_2}}{C_a'} \leq \frac{1 - \sqrt{\gamma}}{2}$, and $\frac{\Upsilon_\tau (\tau + t_0)}{C_a'} \leq \frac{1 - \sqrt{\gamma}}{2}$. Recall that
\[C_a = \frac{2H\bar{\epsilon}}{1 - \gamma}\sqrt{2 \tau \log\left(\frac{2\tau m T}{\delta}\right)}, C_a' = \frac{4}{1 - \gamma}\max\left(C_\psi + C_{\epsilon_2}, 2\bar{x}(\tau + t_0)\right),\]
and
\[C_{\epsilon_1} = H\bar{\epsilon}\sqrt{2 \tau \log\left(\frac{2\tau m T}{\delta}\right)}.\]
Using that $\Upsilon_\tau \leq 2\bar{x}$, one can check that $C_a$ and $C_a'$ satisfy the above three inequalities.

\subsection{Parameter Upper Bound}\label{appendix:proposition:x-bar}

\begin{proposition}\label{proposition:x-bar}
Suppose Assumptions \ref{assump:contraction} and \ref{assump:martingale} hold. Then for all $t$,
\[\norm{x(t)}_v \leq \frac{1}{1 - \gamma}\left((1 + \gamma)\norm{y^*}_v + \frac{\bar{w}}{\underline{v}}\right)\]
holds almost surely, where $y^*\in \mathbb{R}^\mathcal{N}$ is the stationary point of $F$.
\end{proposition}
\begin{proof}[Proof of Proposition \ref{proposition:x-bar}]
By Assumption \ref{assump:contraction}, we have that for all $x \in \mathbb{R}^\mathcal{M}$,
\begin{subequations}\label{proposition:x-bar:e1}
\begin{align}
    \norm{F(\Phi x)}_v &\leq \norm{F(\Phi x) - F(y^*)}_v + \norm{F(y^*)}_v\label{proposition:x-bar:e1:s1}\\
    &\leq \gamma \norm{\Phi x - y^*}_v + \norm{y^*}_v\label{proposition:x-bar:e1:s2}\\
    &\leq \gamma \norm{x}_v + (1 + \gamma)\norm{y^*}_v,\label{proposition:x-bar:e1:s3}
\end{align}
\end{subequations}
where we use the triangle inequality in \eqref{proposition:x-bar:e1:s1} and \eqref{proposition:x-bar:e1:s3}; we use Assumption \ref{assump:contraction} in \eqref{proposition:x-bar:e1:s2}.

Let $\bar{x} = \frac{1}{1 - \gamma}\left((1 + \gamma)\norm{y^*}_v + \frac{\bar{w}}{\underline{v}}\right)$. We prove $\norm{x(t)}_v \leq \bar{x}$ by induction on $t$. Since we initialize $x(0)$ to be $\mathbf{0}$, the statement is true for $t = 0$.

Suppose the statement is true for $t$. By the update rule of $x$, we see that
\begin{subequations}\label{proposition:x-bar:e2}
    \begin{align}
        \frac{1}{v_{h(i_t)}}\abs{x_{h(i_t)}(t + 1)} \leq{}& (1 - \alpha_t)\frac{1}{v_{h(i_t)}}\abs{x_{h(i_t)}(t)} + \alpha_t \left(\frac{1}{v_{h(i_t)}}\abs{F_{i_t}(\Phi x(t))} + \frac{1}{v_{h(i_t)}}\abs{w(t)}\right)\nonumber\\
        \leq{}& (1 - \alpha_t)\norm{x(t)}_v + \alpha_t \left(\norm{F(\Phi x(t))}_v + \frac{\bar{w}}{\underline{v}}\right)\label{proposition:x-bar:e2:s1}\\
        \leq{}& (1 - \alpha_t)\norm{x(t)}_v + \alpha_t \left(\gamma \norm{x(t)}_v + (1 + \gamma)\norm{y^*}_v +  \frac{\bar{w}}{\underline{v}}\right)\label{proposition:x-bar:e2:s2}\\
        \leq{}& (1 - \alpha_t)\bar{x} + \alpha_t \left(\gamma \bar{x} + (1 + \gamma)\norm{y^*}_v +  \frac{\bar{w}}{\underline{v}}\right)\label{proposition:x-bar:e2:s3}\\
        ={}& \bar{x},\nonumber
    \end{align}
\end{subequations}
where we use Assumption \ref{assump:martingale} in \eqref{proposition:x-bar:e2:s1}; \eqref{proposition:x-bar:e1} in \eqref{proposition:x-bar:e2:s2}; the induction assumption in \eqref{proposition:x-bar:e2:s3}.

For $j \not = h(i_t), j \in \mathcal{M}$, we have that
\begin{equation}\label{proposition:x-bar:e3}
    \frac{1}{v_j}\abs{x_j(t+1)} = \frac{1}{v_j}\abs{x_j(t)} \leq \norm{x(t)}_v \leq \bar{x}.
\end{equation}

Combining \eqref{proposition:x-bar:e2} and \eqref{proposition:x-bar:e3}, we see that the statement also holds for $t+1$. Hence we have showed $\norm{x(t)}_v \leq \bar{x}$ by induction.
\end{proof}

\section{TD/Q-Learning with State Aggregation}
\subsection{Asymptotic Convergence of TD Learning with State Aggregation}\label{appendix:TD_Indicator_Asymptotic}
Our asymptotic convergence result for TD learning with state aggregation builds upon the asymptotic convergence result for TD learning with linear function approximation shown in \cite{TDwithFuncApprox}. For completeness, we first present the main result of \cite{TDwithFuncApprox} in Theorem \ref{thm:asy-D-bound}. In order to do this, we must first state a few definitions and assumptions made in \cite{TDwithFuncApprox}.

We use $\phi(i)\in \mathbb{R}^m$ to denote the feature vector associated with state $i \in \mathcal{N}$. Feature matrix $\Phi$ is a $n$-by-$m$ matrix whose $i$'th row is $\phi(i)^\top$. Starting from $\theta(0) = \mathbf{0}$, the $TD(\lambda)$ algorithm keeps updating $\theta, \psi$ by the following update rule,
\begin{equation*}
    \begin{aligned}
    \theta(t+1) &= \theta(t) + \alpha_t d_t \psi_t,\\
    \psi_{t+1} &= \gamma \lambda \psi_t + \phi(i_{t+1}),
    \end{aligned}
\end{equation*}
where $\psi_t$ is named \textit{eligible vector} in \cite{TDwithFuncApprox} and satisfies $\psi_0 = \phi(i_0)$.

Recall that $D = diag(d_1, d_2, \cdots, d_n)$ denotes the stationary distribution of Markov chain $\{i_t\}$. For vectors $x, y \in \mathbb{R}^n$, we define inner product $\langle x, y\rangle = x^\top D y$. The induced norm of this inner product is $\norm{\cdot}_D = \sqrt{\langle \cdot, \cdot \rangle_D}$. Let $L_2(\mathcal{N}, D)$ denote the set of vectors $V \in \mathbb{R}^n$ such that $\norm{V}_D$ is finite.

Recall that we define $\Pi = (\Phi^\top D \Phi)^{-1}\Phi^\top D$. As shown in \cite{TDwithFuncApprox}, the projection matrix that projects an arbitrary vector in $\mathbb{R}^n$ to the set $\{\Phi \theta \mid \theta \in \mathbb{R}^m\}$ is given by $\Phi \Pi$, i.e. for any $V \in L_2(\mathcal{N}, D)$, we have
\begin{equation*}
    \Phi \Pi V = \argmin_{\bar{V} \in \{\Phi \theta \mid \theta \in \mathbb{R}^m\}} \norm{V - \bar{V}}_D.
\end{equation*}
Notice that our definition of matrix $\Pi$ is slightly different with \cite{TDwithFuncApprox} because we want to be consistent with Section \ref{sec:stocApprox}.

To characterize the TD$(\lambda)$ algorithm's dynamics, \cite{TDwithFuncApprox} defines $T^{(\lambda)}: L_2(\mathcal{N}, D) \to L_2(\mathcal{N}, D)$ operator as following: for all $V \in \mathbb{R}^n$, let the $i$'th dimension of $\left(T^{(\lambda)} V\right)$ be defined as
\begin{equation*}
    \left(T^{(\lambda)} V\right)_i =
    \begin{cases}
    (1 - \lambda)\sum_{m=0}^\infty \lambda^m \mathbb{E}\left[\sum_{t=0}^m \gamma^t r(i_t, i_{t+1}) + \gamma^{m+1} V_{i_{m+1}} \mid i_0 = i\right] & \text{ if }\lambda < 1\\
    \mathbb{E}\left[\sum_{t=0}^\infty \gamma^t r(i_t, i_{t+1})\mid i_0 = i\right] & \text{ if }\lambda = 1.
    \end{cases}
\end{equation*}
If $V$ is an approximation of the value function $V^*$, $T^{(\lambda)}$ can be viewed as an improved approximation to $V^*.$ Notice that when $\lambda = 0$, $T^{(\lambda)}$ is identical with the Bellman operator.

Formally, \cite{TDwithFuncApprox} made four necessary assumptions for their main result (Theorem \ref{thm:asy-D-bound}). We omit the third assumption (\cite{TDwithFuncApprox}[Assumption 3]) in our summary because it must hold when the state space $\mathcal{N}$ is finite.

The first assumption (\cite{TDwithFuncApprox}[Assumption 1]) concerns the stationary distribution and the reward function of the Markov chain $\{i_t\}$. It must hold when Assumption \ref{assump:geometric-mixing} holds and every stage reward $r_t$ is upper bounded by $\bar{r}$, as assumed by Theorem \ref{thm:TD-indicator-feature-finite}.

\begin{assumption}\label{assp:TD-transtion-cost}
The transition probability and cost function satisfies the following two conditions:
\begin{enumerate}
    \item The Markov chain $\{i_t\}$ is irreducible and aperiodic. Furthermore, there is a unique distribution $d$ that satisfies $d^\top P = d^\top$ with $d_i > 0$ for all $i \in \mathcal{N}$. Let $\mathbb{E}_0$ stand for expectation with respect to this distribution.
    \item The reward function $r(i_t, i_{t+1})$ satisfies $\mathbb{E}_0\left[r^2(i_t, i_{t+1})\right] < \infty.$
\end{enumerate}
\end{assumption}

The second assumption (\cite{TDwithFuncApprox}[Assumption 2]) concerns the feature vectors and the feature matrix. It must hold when $\Phi$ is defined as \eqref{equ:def:Phi}.

\begin{assumption}\label{assp:TD-features}
The following two conditions hold for $\Phi$:
\begin{enumerate}
    \item The matrix $\Phi$ has full column rank; that is, the $m$ columns (named basis functions in \cite{TDwithFuncApprox}) $\{\phi_k \mid k = 1, \cdots, m\}$ are linearly independent.
    \item For every $k$, the basis function $\phi_k$ satisfies $\mathbb{E}_0\left[\phi_k^2(i_t)\right] < \infty.$
\end{enumerate}
\end{assumption}

The third assumption (\cite{TDwithFuncApprox}[Assumption 4]) concerns the learning step size. It must hold if the learning step sizes are as defined in Theorem \ref{thm:TD-indicator-feature-finite}.

\begin{assumption}\label{assp:TD-step-size}
The step sizes $\alpha_t$ are positive, nonincreasing, and chosen prior to execution of the algorithm. Furthermore, they satisfy $\sum_{t=0}^\infty \alpha_t = \infty$ and $\sum_{t=0}^\infty \alpha_t^2 < \infty.$
\end{assumption}

Now we are ready to present the main asymptotic convergence result given in \cite{TDwithFuncApprox}.

\begin{theorem}\label{thm:asy-D-bound} Under Assumptions \ref{assp:TD-transtion-cost}, \ref{assp:TD-features}, \ref{assp:TD-step-size}, the following hold.
\begin{enumerate}
    \item The value function $V$ is in $L_2(\mathcal{N}, D)$.
    \item For any $\lambda \in [0, 1]$, the TD$(\lambda)$ algorithm with linear function approximation converges with probability one.
    \item The limit of convergence $\theta^*$ is the unique solution of the equation
    \[\Pi T^{(\lambda)}\left(\Phi \theta^*\right) = \theta^*.\]
    \item Furthermore, $\theta^*$ satisfies
    \begin{equation}\label{thm:asy-D-bound:equ1}
        \norm{\Phi \theta^* - V^*}_D \leq \frac{1 - \lambda \gamma}{1 - \gamma}\norm{\Phi \Pi V^* - V^*}_D.
    \end{equation}
\end{enumerate}
\end{theorem}

Notice that \eqref{thm:asy-D-bound:equ1} is not exactly the result we want to obtain. Specifically, we want the both sides of \eqref{thm:asy-D-bound:equ1} to be in $\norm{\cdot}_\infty$ instead of $\norm{\cdot}_D$. Although this kind of result is not obtainable for general TD learning with linear function approximation, we can leverage the special assumptions for state aggregation, which are summarized below:

\begin{assumption}\label{assp:TD-abstraction}
$h: \mathcal{N} \to \mathcal{M}$ is a surjective function from set $\mathcal{N}$ to $\mathcal{M}$. The feature matrix $\Phi$ is as defined in \eqref{equ:def:Phi}, i.e. the feature vector associated with state $i \in \mathcal{N}$ is given by
\begin{equation*}
    \phi_k(i) = \begin{cases}
    1 & \text{ if }k = h(i)\\
    0 & \text{ otherwise}
    \end{cases}, \forall k \in \mathcal{M}.
\end{equation*}
Further, if $h(i) = h(i')$ for $i, i' \in \mathcal{N}$, we have $\abs{V^*(i) - V^*(i')} \leq \zeta$ for a fixed positive constant $\zeta$.
\end{assumption}

Under Assumption \ref{assp:TD-abstraction}, we can show the asymptotic error bound in the infinity norm as we desired:

\begin{theorem}\label{thm:asy-infty-bound}
Under Assumptions \ref{assp:TD-transtion-cost}, \ref{assp:TD-features}, \ref{assp:TD-step-size}, if Assumption \ref{assp:TD-abstraction} also holds, the limit of convergence $\theta^*$ of the $TD(\lambda)$ algorithm satisfies
\[\norm{\Phi \theta^* - V^*}_\infty \leq \frac{(1 - \lambda \gamma)}{1 - \gamma}\norm{\Phi \Pi V^* - V^*}_\infty \leq \frac{(1 - \lambda \gamma)}{1 - \gamma}\zeta.\]
\end{theorem}
To show Theorem \ref{thm:asy-infty-bound}, we need to prove several auxiliary lemmas first. 

\begin{lemma}\label{lemma:contraction-P}
Under Assumption \ref{assp:TD-transtion-cost}, for any $V \in L_2(\mathcal{N}, D)$, we have $\norm{PV}_\infty \leq \norm{V}_{\infty}$.
\end{lemma}
\begin{proof}[Proof of Lemma \ref{lemma:contraction-P}]
This lemma holds because the transition matrix $P$ is non-expansive in infinity norm.
\end{proof}

\begin{lemma}\label{lemma:contraction-T-lambda}
Under Assumption \ref{assp:TD-transtion-cost}, for any $V, \bar{V} \in L_2(\mathcal{N}, D)$, we have
\[\norm{T^{(\lambda)}V - T^{(\lambda)}\bar{V}}_\infty \leq \frac{\gamma (1 - \lambda)}{1 - \gamma \lambda}\norm{V - \bar{V}}_{\infty}.\]
\end{lemma}
\begin{proof}[Proof of Lemma \ref{lemma:contraction-T-lambda}]
By the definition of $T^{(\lambda)}$, we have that
\begin{subequations}\label{lemma:contraction-T-lambda:e1}
\begin{align}
    \norm{T^{(\lambda)}V - T^{(\lambda)}\bar{V}}_{\infty} &= \norm{(1 - \lambda)\sum_{m=0}^\infty \lambda^m (\gamma P)^{m+1}\left(V - \bar{V}\right)}_\infty\nonumber\\
    &\leq (1 - \lambda)\sum_{m=0}^\infty \lambda^m \gamma^{m+1}\norm{V - \bar{V}}_\infty \label{lemma:contraction-T-lambda:e1:s1}\\
    &\frac{\gamma (1 - \lambda)}{1 - \gamma \lambda}\norm{V - \bar{V}}_\infty,\nonumber
\end{align}
\end{subequations}
where inequality \eqref{lemma:contraction-T-lambda:e1:s1} holds because $\norm{V - \bar{V}}_\infty < \infty$ so we use Lemma \ref{lemma:contraction-P}.
\end{proof}

\begin{lemma}\label{lemma:contraction-Pi}
Under Assumption \ref{assp:TD-transtion-cost} and \ref{assp:TD-abstraction}, we have
\begin{equation}\label{lemma:contraction-Pi:e1}
    \norm{\Phi \Pi V^* - V^*}_\infty \leq \zeta
\end{equation}
and for any $V \in L_2(\mathcal{N}, D)$
\begin{equation}\label{lemma:contraction-Pi:e2}
    \norm{\Phi \Pi V}_\infty \leq \norm{V}_\infty.
\end{equation}
\end{lemma}
\begin{proof}[Proof of Lemma \ref{lemma:contraction-Pi}]
For $j \in \mathcal{M}$, we use $h^{-1}(j) \subseteq \mathcal{N}$ to denote all the elements in $\mathcal{N}$ whose feature is $e_{j}$, i.e. $h^{-1}(j) = \{i \mid i\in \mathcal{N}, h(i) = j\}.$ Since $h$ is surjection, $h^{-1}(j) \not = \emptyset, \forall j \in \mathcal{M}.$ Since $\Phi \Pi$ is the projection matrix that projects a vector in $\mathbb{R}^n$ to the set $\{\Phi \theta \mid \theta \in \mathbb{R}^m\}$, we have
\begin{equation*}
    \Pi V = \argmin_{\theta \in \mathbb{R}^m} \sum_{j \in \mathcal{M}} \sum_{i \in h^{-1}(j)} d_i \left(V_i - \theta_j\right).
\end{equation*}
Hence the optimal $\theta_j$ must be in the range $\left[\min_{i\in h^{-1}(j)}V_i, \max_{i\in h^{-1}(j)}V_i\right].$ Therefore, we see that
\begin{equation*}
    \abs{(\Phi \Pi V)_i} = \abs{(\Pi V)_{h(i)}} \leq \max_{i'\in h^{-1}(h(i))}\abs{V_{i'}},
\end{equation*}
which shows \eqref{lemma:contraction-Pi:e2}. Besides, we also have
\begin{equation}\label{lemma:contraction-Pi:e3}
    \begin{aligned}
    \abs{(\Phi \Pi V)_i - V_i} &\leq \max\left(\abs{\min_{i'\in h^{-1}(h(i))}V_{i'} - V_i}, \abs{\max_{i'\in h^{-1}(h(i))}V_{i'} - V_i}\right).
    \end{aligned}
\end{equation}
holds for all $z \in \mathcal{Z}.$ Let $V = V^*$ and use Assumption \ref{assp:TD-abstraction} in \eqref{lemma:contraction-Pi:e3} gives \eqref{lemma:contraction-Pi:e1}.
\end{proof}

Now we come back to the proof of Theorem \ref{thm:asy-infty-bound}.

Notice that
\begin{subequations}\label{thm:asy-infty-bound:e1}
\begin{align}
    \norm{\Phi \theta^* - V^*}_\infty &\leq \norm{\Phi \theta^* - \Phi \Pi V^*}_\infty + \norm{\Phi \Pi V^* - V^*}_\infty\label{thm:asy-infty-bound:e1:s1}\\
    &= \norm{\Phi \Pi T^{(\lambda)}\left(\Phi \theta^*\right) - \Phi \Pi V^*}_\infty + \norm{\Phi \Pi V^* - V^*}_\infty\label{thm:asy-infty-bound:e1:s2}\\
    &\leq \norm{T^{(\lambda)}\left(\Phi \theta^*\right) - V^*}_\infty + \norm{\Phi \Pi V^* - V^*}_\infty\label{thm:asy-infty-bound:e1:s3}\\
    &\leq \frac{\gamma(1 - \lambda)}{1 - \gamma \lambda}\norm{\Phi \theta^* - V^*}_\infty + \norm{\Phi \Pi V^* - V^*}_\infty, \label{thm:asy-infty-bound:e1:s4}
\end{align}
\end{subequations}
where we use the triangle inequality in \eqref{thm:asy-infty-bound:e1:s1}; Theorem \ref{thm:asy-D-bound} in \eqref{thm:asy-infty-bound:e1:s2}; Lemma \ref{lemma:contraction-Pi} in \eqref{thm:asy-infty-bound:e1:s3}; Lemma \ref{lemma:contraction-T-lambda} in \eqref{thm:asy-infty-bound:e1:s4}.

Therefore, we obtain that
\[\norm{\Phi \theta^* - V^*}_\infty \leq \frac{(1 - \lambda \gamma)}{1 - \gamma}\norm{\Pi V^* - V^*}_\infty \leq \frac{(1 - \lambda \gamma)}{1 - \gamma}\zeta,\]
where we use Lemma \ref{lemma:contraction-Pi} in the second inequality.

\subsection{Proof of Theorem \ref{thm:TD-indicator-feature-finite}}\label{appendix:thm:TD-indicator-feature-finite}
Before presenting the proof of Theorem \ref{thm:TD-indicator-feature-finite}, we first show two upper bounds that are needed in the assumptions of Theorem \ref{thm:Stochastic-Approx-Main}. We defer the proof of this result to Appendix \ref{appendix:proposition:upper-bound}.

\begin{proposition}\label{proposition:upper-bound}
Under the same assumptions as Theorem \ref{thm:TD-indicator-feature-finite}, we have $\norm{\theta(t)}_\infty \leq \bar{\theta} := \frac{\bar{r}}{1 - \gamma}$ holds for all $t$ almost surely and $\norm{\theta^*}_\infty \leq \bar{\theta}$. $\abs{w(t)} \leq \bar{w} := \frac{2\bar{r}}{1 - \gamma}$ also holds for all $t$ almost surely.
\end{proposition}

Now we come back to the proof of Theorem \ref{thm:TD-indicator-feature-finite}. Recall that we define $F$ as the Bellman Policy Operator and the noise sequence $w(t)$ as
\[w(t) = r_t + \gamma \theta_{h(i_{t+1})}(t) - \mathbb{E}_{i'\sim \mathbb{P}(\cdot \mid i_t)}\left[r(i_t, i') + \gamma \theta_{h(i')}(t)\right].\]
Let $\theta^*$ be the unique solution of the equation
\[\Pi F(\Phi \theta^*) = \theta^*.\]
By the triangle inequality, we have that
\begin{align}
    \norm{\Phi \cdot \theta(T) - V^*}_\infty &\leq \norm{\Phi \cdot \theta(T) - \Phi \cdot \theta^*}_\infty + \norm{\Phi \cdot \theta^* - V^*}_\infty\nonumber\\
    &\leq \norm{\theta(T) - \theta^*}_\infty + \norm{\Phi \cdot \theta^* - V^*}_\infty.\label{thm:TD-indicator-feature-finite:e1}
\end{align}
We first bound the first term of \eqref{thm:TD-indicator-feature-finite:e1} by Theorem \ref{thm:Stochastic-Approx-Main}. To do this, we first rewrite the update rule of TD learning with state aggregation \eqref{equ:TD0_update} in the form of the SA update rule \eqref{equ:Generalized-Asy-Stoc-Approx}:
\begin{equation*}
    \begin{aligned}
    \theta_{h(i_t)}(t+1) &= \theta_{h(i_t)}(t) + \alpha_t\left(F_{i_t}\left( \Phi \theta(t)\right) - \theta_{h(i_t)}(t) + w(t)\right),\\
    \theta_j(t+1) &= \theta_j(t) \text{ for }j \not = h(i_t), j \in \mathcal{M}.
    \end{aligned}
\end{equation*}
Now we verify all the assumptions of Theorem \ref{thm:Stochastic-Approx-Main}. Assumption \ref{assump:geometric-mixing} is assumed to be satisfied in the body of Theorem \ref{thm:TD-indicator-feature-finite}. As for Assumption \ref{assump:contraction}, $F$ is $\gamma$-contraction in the infinity norm because it is the Bellman operator, and we can set $C = \frac{2\bar{r}}{1 - \gamma}$ so that $C \geq (1 + \gamma)\norm{y^*}_\infty$ (see the discussion below Assumption \ref{assump:contraction}). As for Assumption \ref{assump:martingale}, by the definition of noise sequence $w(t)$, we see that
\begin{align}
    \mathbb{E}\left[w(t)\mid \mathcal{F}_t\right] &= \mathbb{E}\left[r_t + \gamma \theta_{h(i_{t+1})}(t) - \mathbb{E}_{i'\sim \mathbb{P}(\cdot \mid i_t)}\left[r(i_t, i') + \gamma \theta_{h(i')}(t)\right]\mid \mathcal{F}_t\right]\nonumber\\
    &= \mathbb{E}\left[r_t + \gamma \theta_{h(i_{t+1})}(t)\mid \mathcal{F}_t\right] - \mathbb{E}_{i'\sim \mathbb{P}(\cdot \mid i_t)}\left[r(i_t, i') + \gamma \theta_{h(i')}(t)\right]\nonumber\\
    &= 0.\nonumber
\end{align}
In addition, we can set $\bar{w} = \frac{2\bar{r}}{1 - \gamma}$ according to Proposition \ref{proposition:upper-bound}. Finally, we can set $\bar{\theta} = \frac{\bar{r}}{1 - \gamma}$ according to Proposition \ref{proposition:upper-bound}.

Therefore, by Theorem \ref{thm:Stochastic-Approx-Main}, we see that
\begin{equation}\label{thm:TD-indicator-feature-finite:e2}
    \norm{\theta(T) - \theta^*}_\infty \leq \frac{C_a}{\sqrt{T + t_0}} + \frac{C_a'}{T + t_0}, \text{ where}
\end{equation}
\begin{align*}
    C_a &= \frac{40H\bar{r}}{(1 - \gamma)^2}\sqrt{K_2 \log T}\cdot \sqrt{\log T + \log \log T + \log \left(\frac{4m K_2}{\delta}\right)},\\
    C_a'&= \frac{8\bar{r}}{(1 - \gamma)^2}\max\left(\frac{144 K_2 H \log T}{\sigma'} + 4K_1(1 + 2K_2 + 4H), 2K_2 \log T + t_0\right).
\end{align*}

As for the second term of \eqref{thm:TD-indicator-feature-finite:e1}, by Theorem \ref{thm:asy-infty-bound}, we have that
\begin{equation}\label{thm:TD-indicator-feature-finite:e3}
    \norm{\Phi \cdot \theta^* - V^*}_\infty \leq \frac{\zeta}{1 - \gamma}.
\end{equation}
Substituting \eqref{thm:TD-indicator-feature-finite:e2} and \eqref{thm:TD-indicator-feature-finite:e3} into \eqref{thm:TD-indicator-feature-finite:e1} finishes the proof.

\subsection{Proof of Proposition \ref{proposition:upper-bound}}\label{appendix:proposition:upper-bound}
We show $\norm{\theta(t)}_\infty \leq \frac{\bar{r}}{1 - \gamma}$ by induction on $t$. The statement holds for $t = 0$ because we initialize $\theta(0) = \mathbf{0}$. Suppose the statement holds for $t$. By the induction assumption, we see that
\begin{equation*}
    \begin{aligned}
    \theta_{h(i_t)}(t + 1) &= (1 - \alpha_t)\theta_{h(i_t)}(t) + \alpha_t \left[r_t + \gamma \theta_{h(i_{t+1})}(t)\right]\\
    &\leq (1 - \alpha_t)\norm{\theta(t)}_\infty + \alpha_t \left[r_t + \gamma \norm{\theta(t)}_\infty\right]\\
    &\leq (1 - \alpha_t)\frac{\bar{r}}{1 - \gamma} + \alpha_t \left[r_t + \gamma \cdot \frac{\bar{r}}{1 - \gamma}\right]\\
    &\leq \frac{\bar{r}}{1 - \gamma}.
    \end{aligned}
\end{equation*}
For $j \not = h(i_t), j \in \mathcal{M}$, we have that
\begin{equation*}
    \theta_j(t+1) = \theta_j(t) \leq \norm{\theta(t)}_\infty \leq \frac{\bar{r}}{1 - \gamma}.
\end{equation*}
Hence the statement also holds for $t+1$. Therefore, we have showed $\norm{\theta(t)}_\infty \leq \frac{\bar{r}}{1 - \gamma}$ by induction.

By Theorem \ref{thm:asy-D-bound}, we know $\theta^* = \lim_{t \to \infty}\theta(t)$. Since we have already shown that $\norm{\theta(t)}_\infty \leq \frac{\bar{r}}{1 - \gamma}$ holds for all $t$, we must have $\norm{\theta^*}_\infty \leq \frac{\bar{r}}{1 - \gamma}$.

Using $\norm{\theta(t)}_\infty \leq \frac{\bar{r}}{1 - \gamma}$, we see that
\begin{equation*}
    \begin{aligned}
    \abs{w(t)} &\leq \abs{r_t} + \gamma \abs{\theta_{h(i_{t+1})}(t)} - \abs{\mathbb{E}_{i' \sim \mathbb{P}(\cdot\mid i_t)}\left[r(i_t, i') + \gamma \theta_{h(i')}(t)\right]}\\
    &\leq 2\bar{r} + 2\gamma \bar{\theta}\\
    &= \frac{2\bar{r}}{1 - \gamma}.
    \end{aligned}
\end{equation*}

\subsection{Application of the SA Scheme to Q-learning with State and Action Aggregation}\label{appendix:asynchronous-Q}
We study $Q$-learning with state and action aggregation in a setting that is a generalization of the tabular setting studied in \cite{qu2020finite}. Specifically, we consider an MDP $M$ with a finite state space $\mathcal{S}$ and finite action space $\mathcal{A}$. Suppose the transition probability is given by $\mathbb{P}(s_{t+1} = s'\mid s_t = s, a_t = a) = \mathbb{P}(s'\mid s, a)$, and the stage reward at time step $t$ is a random variable $r_t$ with its expectation given by $R_{s_t, a_t}$. Under a stochastic policy $\pi$, the $Q$ function (vector) $Q^\pi \in \mathbb{R}^{\mathcal{S}\times \mathcal{A}}$ is defined as
\[Q_{s, a}^\pi = \mathbb{E}_\pi \left[\sum_{t=0}^\infty \gamma^t r_t \Big| (s_0, a_0) = (s, a)\right],\]
where $0 \leq \gamma < 1$ is the discounting factor. We use $Q^*$ to denote the $Q$ function corresponding to the optimal policy $\pi^*$.

Similar to \cite{qu2020finite}, we assume the trajectory $\{(s_t, a_t, r_t)\}_{t=0}^\infty$ is sampled by implementing a fixed behavioral stochastic policy $\pi$. In $Q$-learning with state and action aggregation, the state abstraction $\psi_1$ operates on the state space $\mathcal{S}$ and the action abstraction $\psi_2$ operates on action space $\mathcal{A}$. For simplicity of notation, we define the abstraction space as $\mathcal{M} = \psi_1(\mathcal{S})\times \psi_2(\mathcal{A})$ and the abstraction operator $h: \mathcal{S}\times \mathcal{A} \to \mathcal{M}$ as $h(s, a) = (\psi_1(s), \psi_2(a))$. The update rule for $Q$-learning with state and action aggregation is then given by
\begin{equation}\label{equ:Q-learning-update}
    \begin{aligned}
    \theta_{h(s_t, a_t)}(t + 1) &= (1 - \alpha_t)\theta_{h(s_t, a_t)}(t) + \alpha_t \left[r_t + \gamma \max_{a \in \mathcal{A}}\theta_{h(s_{t+1}, a)}(t)\right],\\
    \theta_j(t + 1) &= \theta_j(t) \text{ for }j \not = h(s_t, a_t).
    \end{aligned}
\end{equation}
As a remark, some previous work considers abstraction on the state space $\mathcal{S}$ but does not compress the action space (see \cite{Jiang2018NotesOS}). In contrast, our setting also compresses the action space, and when $\psi_2$ is the identity map, our setting reduces to the case with only state aggregation.

To apply the result in Section \ref{sec:stocApprox}, we define function $F$ as the \textit{Bellman Optimality Operator}, i.e. 
\[F_{s,a}(Q) = R_{s, a} + \gamma \mathbb{E}_{s' \sim \mathbb{P}(\cdot \mid s, a)}\max_{a'\in \mathcal{A}}Q_{s', a'}.\]
It is shown in \cite{bertsekas1996neuro} that $Q^*$ is the unique fixed point of function $F$. By viewing $\mathcal{S} \times \mathcal{A}$ as $\mathcal{N}$, we can define matrix $\Phi \in {\mathcal{N}\times \mathcal{M}}$ as in \eqref{equ:def:Phi}. We can rewrite the update rule \eqref{equ:Q-learning-update} as
\begin{equation*}
    \begin{aligned}
    \theta_{h(s_t, a_t)}(t + 1) &= \theta_{h(s_t, a_t)}(t) + \alpha_t \left[F_{s_t, a_t}\left(\Phi \theta(t)\right) - \theta_{h(s_t, a_t)}(t) + w(t)\right],\\
    \theta_j(t + 1) &= \theta_j(t) \text{ for }j \not = h(s_t, a_t),
    \end{aligned}
\end{equation*}
where
\begin{equation*}
    \begin{aligned}
    w(t) &= r_t + \gamma \max_{a \in \mathcal{A}}\theta_{h(s_{t+1}, a)}(t) - F_{s_t, a_t}\left(\Phi \theta(t)\right)\\
    &= (r_t - R_{s_t, a_t}) + \gamma \left[\max_{a \in \mathcal{A}}\theta_{h(s_{t+1}, a)}(t) - \mathbb{E}_{s'\sim \mathbb{P}(\cdot \mid s_t, a_t)}\max_{a' \in \mathcal{A}}\theta_{h(s', a')}(t)\right].
    \end{aligned}
\end{equation*}
Hence we have $\mathbb{E}[w(t)\mid \mathcal{F}_t] = 0.$ In order to apply Theorem \ref{thm:Stochastic-Approx-Main}, we need the following assumption on the induced Markov chain of stochastic policy $\pi$ which is standard, cf. \cite{qu2020finite}.

\begin{assumption}\label{assump:Q-learning}
The following conditions hold:
\begin{enumerate}
    \item For each time step $t$, the stage reward $r_t$ satisfies $\abs{r_t} \leq \bar{r}$ almost surely.
    \item Under the behavioral policy $\pi$, the induced Markov chain $(s_t, a_t)$ with state space $\mathcal{S}\times \mathcal{A}$ satisfies Assumption \ref{assump:geometric-mixing} with stationary distribution $d$ and parameters $\sigma', K_1, K_2$.
\end{enumerate}
\end{assumption}

The next assumption is approximate $Q^*$-irrelevant abstraction, which measures the quality of the abstraction map and is standard in the literature (see \cite{Jiang2018NotesOS}).

\begin{assumption}\label{assump:approximate-Q-irrelevant}
There exists an abstract $Q$ function $q: \mathcal{M} \to \mathbb{R}$ such that $\norm{\Phi q - Q^*}_\infty \leq \epsilon_{Q^*}$.
\end{assumption}

We can now state our theorem for $Q$-learning with state aggregation.

\begin{theorem}\label{thm:asy-Q-learning}
Under Assumption \ref{assump:Q-learning} and \ref{assump:approximate-Q-irrelevant}, suppose the step size of $Q$-learning with state aggregation is given by $\alpha_t = \frac{H}{t + t_0}$, where $t_0 = \max(4H, 2K_2 \log T)$ and $H \geq \frac{2}{\sigma' (1 - \gamma)}$. Then, with probability at least $1 - \delta$,
{\small\begin{equation*}
    \begin{aligned}
&\norm{\Phi \cdot \theta(T) - Q^*}_\infty \leq \frac{C_a}{\sqrt{T + t_0}} + \frac{C_a'}{T + t_0} + \frac{2\epsilon_{Q^*}}{1 - \gamma}, \text{ where} \\
    C_a &= \frac{40H\bar{r}}{(1 - \gamma)^2}\sqrt{K_2 \log T}\cdot \sqrt{\log T + \log \log T + \log \left(\frac{4m K_2}{\delta}\right)},\\
    C_a'&= \frac{8\bar{r}}{(1 - \gamma)^2}\max\left(\frac{144 K_2 H \log T}{\sigma'} + 4K_1(1 + 2K_2 + 4H), 2K_2 \log T + t_0\right).
\end{aligned}
\end{equation*}}
\end{theorem}
\begin{proof}[Proof of Theorem \ref{thm:asy-Q-learning}]
Define $\theta^*$ as the unique solution of equation $\theta = \Pi F(\Phi \theta)$, where the definition of $\Pi$ is given in \eqref{equ:def-Pi}. Under Assumption \ref{assump:Q-learning}, we see that $\norm{\theta^*}_\infty \leq \frac{\bar{r}}{1 - \gamma}$: otherwise, by assuming that $\abs{\theta_i^*} = \norm{\theta^*}_\infty > \frac{\bar{r}}{1 - \gamma}$, we can derive a contradiction that $\norm{\Pi F(\Phi \theta^*)}_\infty < \abs{\theta_i^*}$. To see this, recall that linear operators $\Pi$ and $\Phi$ are non-expansions in the infinity norm (see Appendix \ref{appendix:proposition:proj-contraction}), and $\norm{F(v)}_\infty < \norm{v}_\infty$ for a vector $v \in \mathbb{R}^{\mathcal{N}}$ if $\norm{v}_\infty > \frac{\bar{r}}{1 - \gamma}$.

Further, using a similar approach with the proof of Proposition \ref{proposition:upper-bound}, we also see that
\[\norm{\theta(t)}_\infty \leq \bar{\theta} := \frac{\bar{r}}{1 - \gamma}, \abs{w(t)} \leq \bar{w} := \frac{2\bar{r}}{1 - \gamma}\]
hold for all $t$ almost surely.

Therefore, by Theorem \ref{thm:Stochastic-Approx-Main}, we obtain that
\begin{equation}\label{thm:asy-Q-learning:e1}
    \norm{\theta(T) - \theta^*}_\infty \leq \frac{C_a}{\sqrt{T + t_0}} + \frac{C_a'}{T + t_0}.
\end{equation}
To finish the proof of Theorem \ref{thm:asy-Q-learning}, we only need to show that
\begin{equation}\label{thm:asy-Q-learning:e2}
    \norm{\Phi \theta^* - Q^*} \leq \frac{2\epsilon_{Q^*}}{1 - \gamma}.
\end{equation}

Given the behavioral policy $\pi$, we use $\{d_{s, a} \mid (s, a) \in \mathcal{S} \times \mathcal{A}\}$ to denote the stationary distribution under policy $\pi$. Recall that we define $\mathcal{M} = \psi_1(\mathcal{S}) \times \psi_2(\mathcal{A})$. For each abstract state-action pair $(x, y) \in \mathcal{M}$, we define a distribution $p_{(x, y)}$ over $h^{-1}(x, y)$ such that
\[p_{(x, y)}(s , a) = \frac{d_{s, a}}{\sum_{(\tilde{s}, \tilde{a})\in h^{-1}(x, y)}d_{\tilde{s}, \tilde{a}}}, \forall (s, a) \in h^{-1}(x, y).\]
Using the set of distributions $\{p_{(x, y)} \mid (x, y) \in \mathcal{M}\}$, we define two new MDPs:
\begin{equation}\label{equ:def-M-psi}
    M_\psi = \left(\psi_1(\mathcal{S}), \psi_2(\mathcal{A}), P_\psi, R_\psi, \gamma\right),
\end{equation}
where $(R_{\psi})_{x, y} = \mathbb{E}_{(s, a)\sim p_{(x, y)}}[R_{s, a}]$, and $P_\psi(x'\mid x, y) = \mathbb{E}_{(s, a)\sim p_{(x, y)}}[P(x'\mid s, a)]$; and
\begin{equation}\label{equ:def-M-psi-prime}
    M_\psi' = (\mathcal{S}, \mathcal{A}, P_\psi', R_\psi', \gamma),
\end{equation}
where $(R_\psi')_{s, a} = \mathbb{E}_{(\tilde{s}, \tilde{a})\sim p_{h(s, a)}}[R_{\tilde{s}, \tilde{a}}], P_\psi'(s'\mid s, a) = \mathbb{E}_{(\tilde{s}, \tilde{a})\sim p_{h(s, a)}}[P(s'\mid \tilde{s}, \tilde{a})].$

We use $\Gamma$ to denote the Bellman Optimality Operator. For simplicity, we use the subscript to distinguish the value functions ($V^*$), the state-action value functions ($Q^*$), and the Bellman Optimality Operators ($\Gamma$) of the three MDPs $M, M_\psi$ and $M_{\psi}'$. Notice that $\Gamma_M$ is identical with $F$. %

We can show that $\theta^*$ is identical with the state-action value function of $M_\psi$, i.e.,
\begin{equation}\label{thm:asy-Q-learning:e3}
    \theta^* = Q_{M_\psi}^*.
\end{equation}
To see this, we notice that $(\Phi \theta^*)_{s, a} = \theta^*_{h(s, a)}$. Hence we get that
\begin{align*}
    F(\Phi \theta^*)_{s, a} &= [\Gamma_M \Phi \theta^*]_{s, a}\\
    &= R_{s, a} + \mathbb{E}_{s'\sim P(s, a)}\left[\max_a (\Phi \theta^*)_{s', a}\right]\\
    &= R_{s, a} + \mathbb{E}_{s'\sim P(s, a)}\left[\max_a \theta^*_{h(s', a)}\right].
\end{align*}
Using this, we further obtain that
\begin{align*}
    \left(\Pi F(\Phi \theta^*)\right)_{x, y} &= \sum_{(s, a)\in h^{-1}(x, y)}\frac{d_{s, a}}{\sum_{(\tilde{s}, \tilde{a})\in h^{-1}(x, y)}d_{\tilde{s}, \tilde{a}}}\left(R_{s, a} + \mathbb{E}_{s'\sim P(s, a)}\left[\max_a \theta^*_{h(s', a)}\right]\right)\\
    &= \sum_{(s, a)\in h^{-1}(x, y)}p_{(x, y)}(s, a)\left(R_{s, a} + \mathbb{E}_{s'\sim P(s, a)}\left[\max_a \theta^*_{h(s', a)}\right]\right)\\
    &= (R_\psi)_{x, y} + \sum_{(s, a)\in h^{-1}(x, y)}p_{(x, y)}(s, a) \sum_{x'\in \psi_1(\mathcal{S})}P(x'\mid s, a) \max_a \theta^*_{x', \psi_2(a)}\\
    &= (R_\psi)_{x, y} + \sum_{x'\in \psi_1(\mathcal{S})}P_\psi(x'\mid x, y) \max_{y'} \theta^*_{x', y'}\\
    &= [\Gamma_{M_\psi}\theta^*]_{x, y}.
\end{align*}
Since we have $\Pi F(\Phi \theta^*) = \theta^*$ by definition, we see that
\[[\Gamma_{M_\psi}\theta^*]_{x, y} = \theta^*_{x, y}, \forall (x, y) \in \mathcal{M}.\]
Thus we have shown that $\theta^* = Q_{M_\psi}^*$.

Next, we observe that the state-value function of MDP $M_\psi'$ is given by
\begin{equation}\label{thm:asy-Q-learning:e4}
    Q_{M_\psi'}^* = \Phi Q_{M_\psi}^*.
\end{equation}
This is because
\begin{subequations}\label{thm:asy-Q-learning:e5}
\begin{align}
    \left(\Gamma_{M_\psi'}(\Phi Q_{M_\psi}^*)\right)_{s, a} ={}& (R_\psi')_{s, a} + \gamma \sum_{s' \in \mathcal{S}} P_\psi'(s' \mid s, a) \max_{a'} (\Phi Q_{M_\psi}^*)_{s', a'}\nonumber\\
    ={}& (R_\psi')_{s, a} + \gamma \langle P_\psi'(s, a), \Phi V_{M_\psi}^*\rangle \nonumber\\
    ={}& \sum_{(\tilde{s}, \tilde{a})\in h^{-1}(h(s, a))}p_{h(s, a)}(\tilde{s}, \tilde{a})\left(R_{\tilde{s}, \tilde{a}} + \gamma \langle P(\tilde{s}, \tilde{a}), \Phi V_{M_\psi}^*\rangle \right)\label{thm:asy-Q-learning:e5:s1}\\
    ={}& \sum_{(\tilde{s}, \tilde{a})\in h^{-1}(h(s, a))}p_{h(s, a)}(\tilde{s}, \tilde{a})R_{\tilde{s}, \tilde{a}}\nonumber\\
    &+ \sum_{(\tilde{s}, \tilde{a})\in h^{-1}(h(s, a))}p_{h(s, a)}(\tilde{s}, \tilde{a})\gamma \langle P(\tilde{s}, \tilde{a}), \Phi V_{M_\psi}^*\rangle\nonumber\\
    ={}& (R_\psi)_{h(s, a)} + \gamma \langle P_\psi(h(s, a)), V_{M_\psi}^* \rangle\label{thm:asy-Q-learning:e5:s2}\\
    ={}& (Q_{M_\psi}^*)_{h(s, a)}\nonumber\\
    ={}& (\Phi Q_{M_\psi}^*)_{s, a}, \nonumber
\end{align}
\end{subequations}
where we use the definition of $M_\psi'$ (see \eqref{equ:def-M-psi-prime}) in \eqref{thm:asy-Q-learning:e5:s1}; we use the definition of $M_\psi$ (see \eqref{equ:def-M-psi}) in \eqref{thm:asy-Q-learning:e5:s2}.

By \eqref{thm:asy-Q-learning:e4}, we see that
\begin{equation}\label{thm:asy-Q-learning:e6}
    \norm{\Phi Q_{M_\psi}^* - Q_M^*}_\infty = \norm{Q_{M_\psi'}^* - Q_M^*}_\infty \leq \frac{1}{1 - \gamma}\norm{\Gamma_{M_\psi'}Q_M^* - Q_M^*}_\infty.
\end{equation}
We further notice that
\begin{subequations}\label{thm:asy-Q-learning:e7}
\begin{align}
    &\abs{(\Gamma_{M_\psi}^*Q_M^*)_{s, a} - (Q_M^*)_{s, a}}\nonumber\\
    ={}&\abs{(R_\psi')_{s, a} + \gamma \langle P_\psi(s, a), V_M^*\rangle - (Q_M^*)_{s, a}}\nonumber\\
    ={}&\abs{\left(\sum_{(\tilde{s}, \tilde{a})\in h^{-1}(h(s, a))}p_{h(s, a)}(\tilde{s}, \tilde{a})(R_{\tilde{s}, \tilde{a}} + \gamma\langle P(\tilde{s}, \tilde{a}), V_M^*\rangle)\right) - (Q_M^*)_{s, a}}\label{thm:asy-Q-learning:e7:s1}\\
    ={}& \abs{\sum_{(\tilde{s}, \tilde{a})\in h^{-1}(h(s, a))}p_{h(s, a)}(\tilde{s}, \tilde{a})\left((Q_M^*)_{\tilde{s}, \tilde{a}} - (Q_M^*)_{s, a}\right)}\nonumber\\
    \leq{}& \sum_{(\tilde{s}, \tilde{a})\in h^{-1}(h(s, a))}p_{h(s, a)}(\tilde{s}, \tilde{a})\abs{(Q_M^*)_{\tilde{s}, \tilde{a}} - (Q_M^*)_{s, a}}\nonumber\\
    \leq{}& \sum_{(\tilde{s}, \tilde{a})\in h^{-1}(h(s, a))}p_{h(s, a)}(\tilde{s}, \tilde{a})(2\epsilon_{Q^*})\label{thm:asy-Q-learning:e7:s2}\\
    ={}& 2\epsilon_{Q^*},\nonumber
\end{align}
\end{subequations}
where we use the definition of $M_\psi$ in \eqref{thm:asy-Q-learning:e7:s1}; we use Assumption \ref{assump:approximate-Q-irrelevant} in \eqref{thm:asy-Q-learning:e7:s2}.

Substituting \eqref{thm:asy-Q-learning:e7} into \eqref{thm:asy-Q-learning:e6} gives that
\begin{equation}\label{thm:asy-Q-learning:e8}
    \norm{\Phi Q_{M_\psi}^* - Q_M^*}_\infty \leq \frac{2\epsilon_{Q^*}}{1 - \gamma}.
\end{equation}
Combining \eqref{thm:asy-Q-learning:e3} and \eqref{thm:asy-Q-learning:e8} finishes the proof.
\end{proof}

\end{document}